\documentclass[runningheads]{llncs}

\usepackage[mobile]{Styles/eccv}

\usepackage{Styles/eccvabbrv}

\usepackage{graphicx}
\usepackage{booktabs}
\usepackage{wrapfig}
\usepackage{enumitem}
\usepackage{multirow}
\usepackage{hyperref}

\newcommand{\minus}{\scalebox{0.75}[1.0]{$-$}}

\usepackage[accsupp]{axessibility}  

\usepackage{hyperref}

\usepackage{orcidlink}

\begin{document}

\title{Exploring the Impact of Synthetic Data for Aerial-view Human Detection} 

\titlerunning{Exploring the Impact of Synthetic Data}

\author{Hyungtae Lee\inst{1,*}~~~~~~~~~~~~~~~
Yan Zhang\inst{2,*}~~~~~~~~~~~~~~~
Yi-Ting Shen\inst{2}\\
Heesung Kwon\inst{3}~~~~~~~~~~~~~~~
Shuvra S. Bhattacharyya\inst{2}}
\authorrunning{H. Lee et al.}
%
\institute{BlueHalo \and 
University of Maryland, College Park \and
DEVCOM Army Research Laboratory\\\vspace{0.25cm}
* equal contribution}

\maketitle

\begin{abstract}

Aerial-view human detection has a large demand for large-scale data to capture more diverse human appearances compared to ground-view human detection. Therefore, synthetic data can be a good resource to expand data, but the domain gap with real-world data is the biggest obstacle to its use in training. As a common solution to deal with the domain gap, the \emph{sim2real transformation} is used, and its quality is affected by three factors: i) the real data serving as a reference when calculating the domain gap, ii) the synthetic data chosen to avoid the transformation quality degradation, and iii) the synthetic data pool from which the synthetic data is selected. In this paper, we investigate the impact of these factors on maximizing the effectiveness of synthetic data in training in terms of improving learning performance and acquiring domain generalization ability--two main benefits expected of using synthetic data. As an evaluation metric for the second benefit, we introduce a method for measuring the distribution gap between two datasets, which is derived as the normalized sum of the Mahalanobis distances of all test data. As a result, we have discovered several important findings that have never been investigated or have been used previously without accurate understanding. We expect that these findings can break the current trend of either naively using or being hesitant to use synthetic data in machine learning due to the lack of understanding, leading to more appropriate use in future research.

\keywords{Synthetic data \and sim2real transformation \and distribution gap \and aerial-view human detection}

\end{abstract}    
\section{Introduction}
\label{sec:intro}

Advances in aerial-view human detection have lagged significantly behind the conventional human detection, which typically deals with ground-view images. This is because there is a lack of real-world data that sufficiently includes the human appearance, which becomes very diverse when viewed from the air compared to when viewed from the ground. Therefore, the demand for synthetic data grows as additional data that can expand the human appearance from the aerial-view in the data. Many attempts\cite{YShenCVPR2023,YShenAccess2023} to use synthetic data for training data augmentation have been made in aerial-view human detection, but properly creating synthetic data relevant to given learning tasks remains a challenge. That is mainly because fully exploiting the inherent strengths of synthetic data requires an appropriate understanding of various properties inducing the domain gap compared to real data.

\begin{wrapfigure}{r}{0.5\textwidth}
\vspace{-0.7cm}
\centering
\includegraphics[trim=5mm 5mm 5mm 5mm,clip,width=\linewidth]{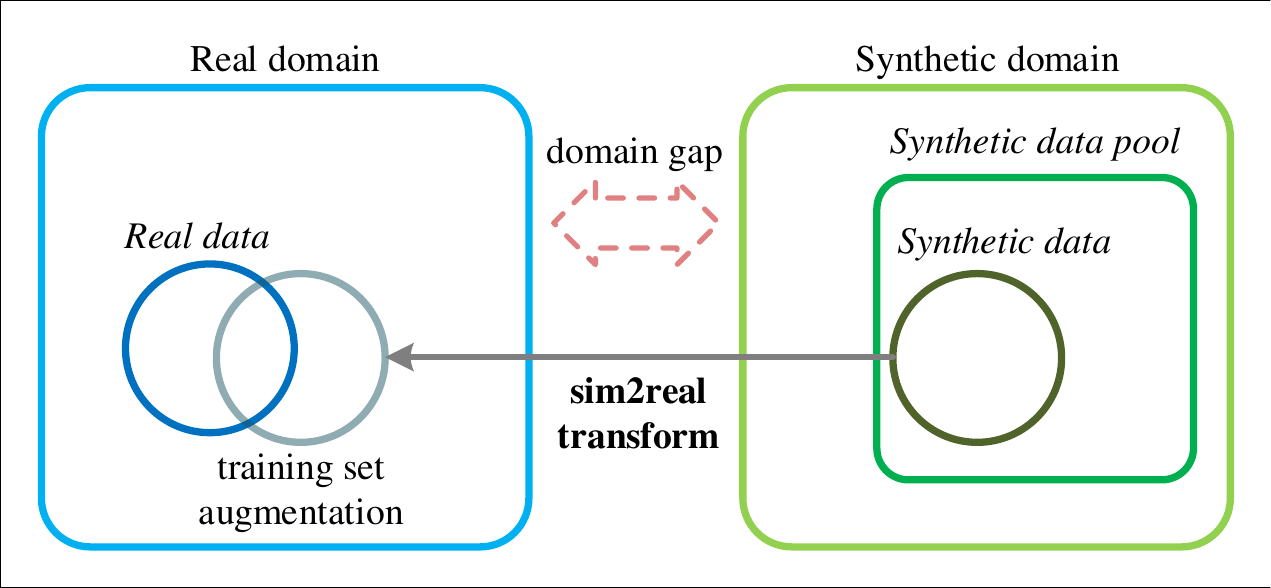}
\vspace{-0.6cm}
\caption{{\bf Sim2real transformation mechanism.} Three datasets (real data, synthetic data, and the synthetic data pool) can influence the impact of synthetic data used in training.}
\label{fig:concept}
\vspace{-0.6cm}
\end{wrapfigure}

A common solution for bridging the domain gap is to transform the properties of synthetic data to enhance realism. In general, the \emph{sim2real} transformers are trained on a source-to-target adaptation framework (\eg, conditional GAN~\cite{JZhuICCV2017,JHoffmanICML2018,YShenCVPR2023}), treating synthetic and real data as the source and target domains, respectively. However, the satisfactory quality of the sim2real transformation cannot be expected if the domain gap between the two sets is not manageable to overcome. One effective way to deal with this dilemma, where the sim2real transformation to handle the large domain gap is negatively affected by the large domain gap, is to use only a portion of synthetic data with a small domain gap with the real data. To take advantage of the diversity of synthetic data in training, data with relatively large domain gaps can also be selected at a lower rate.~\cite{YShenCVPR2023} In summary, there are three factors that affect sim2real transformation quality (Fig.~\ref{fig:concept}): i) the real data serving as a reference when measuring the domain gap, ii) the selected synthetic data used in training, and iii) the synthetic data pool from which a subset of synthetic data is selected. To maximize the impact of synthetic data in training while adequately addressing the domain gap, a thorough investigation into how the three factors play into optimally integrating synthetic data into learning is critical in aerial-view human detection.

In this study, we evaluated the impact of synthetic data in a same-domain task where the training and test sets were built on the same domain, and a cross-domain task where the two sets were from different domains. With the two tasks, we aim to ensure the following two benefits of using synthetic data in training are realized: i) improving learning performance by training with hybrid sets of real and synthetic data, and ii) leading the model to acquire domain generalization ability to achieve satisfactory accuracy regardless of the dataset in a cross-domain task. To do that, we firstly evaluated the model performance. We also measured the distribution gap between the training and test sets in a cross-domain task with and without using synthetic data in training to specifically confirm the second benefit. As a measure of the distribution gap, we use the cross-entropy while representing the distributions of the two sets as a multivariate Gaussian distribution and a mixture of Delta distribution, respectively. We have shown that, theoretically, the distribution for a certain category of a training set used in detector training can be represented as a multivariate Gaussian distribution in the detector's representation space. In the end, the distribution gap can be derived as the normalized sum of the Mahalanobis distances~\cite{PMahalanobisNISI1936} from the training set for each test data.

After carrying out a comprehensive study based on extensive experiments with the two measurements (detection accuracy and distribution gaps), we uncovered the following useful findings:
\begin{enumerate}[label=\arabic*)]
    \item  In a cross-domain task, using synthetic data helps to significantly reduce the distribution gaps of most of the test data but also unexpectedly produces considerably large distribution gaps for some outlier data.
    \item To enhance the impact of synthetic data in training, it is important to increase the amount of not only synthetic data but also real data in both the same-domain and cross-domain tasks.
    \item In a cross-domain task, improving the sim2real transformation quality of the synthetic data is more effective in leading the model to acquiring domain generalization ability than reducing the distribution gap between the training and test sets when achieving the two objectives together is not feasible.
    \item The properties of the synthetic data pool (\ie, the density and diversity of the synthetic data distribution in the feature space, and the domain gap with respect to the real data) also influence the impact from using synthetic data in training.
\end{enumerate}
In recent works, the above findings have not been carefully considered or have been used without accurate understanding. We provide empirical evidence verifying the findings through extensive experiments. We hope that our study can provide a clue for a breakthrough that can address the community's hesitant or improper use of synthetic data even in tasks other than aerial-view human detection.
\section{Related Works}
\label{sec:rel_works}

\noindent{\bf Aerial-view human detection.} There exists very little literature on aerial-view human detection compared to ground-view human detection. This is because off-the-shelf models developed for ground-view detection also perform well in aerial-view human detection on general benchmarks~\cite{ZWuICCV2019,PZhuTPAMI2021}. However, these detectors do not play a role at all in data-scarce regimes where the training set does not represent the test set, \eg, few-shot learning or cross-domain detection. To cope with this data-scarce situation, which is difficult to avoid in aerial-view human detection, several attempts have been made to expand the training data, \eg, generating novel-view images from existing images via NeRF~\cite{CMaxeyICRA2024} or using synthetic data developed for aerial-view human detection~\cite{YShenAccess2023,YShenCVPR2023}. Our method was also developed to expand the training data with a greater focus on eliciting a variety of human poses.\smallskip

\noindent{\bf Measuring distribution gap between two datasets.} Measuring the differences in the properties of distinctive datasets is crucial for analyzing performance in cross-dataset tasks (\eg, domain adaptation/generalization, sim2real transformation). Measurements depend on which property is focused on in the analysis. To measure the differences in \emph{class conditional distributions} of two datasets, scatter~\cite{MGhifaryTPAMI2017}, maximum mean discrepancy (MMD)~\cite{HYanCVPR2017,HLiCVPR2018}, high-order MMD (HoMM)~\cite{CChenAAAI2020}, {\it etc}. are used. Statistical measures over the distances between samples of different datasets in the \emph{feature space} are also considered to estimate the distribution gap of the datasets. Here, the feature space can be learned in a direction of preserving the properties of the synthetic data in the sim2real transformation~\cite{JZhuICCV2017} or minimizing the feature distribution of two datasets through contrastive learning~\cite{SMotiianICCV2017,XYaoCVPR2022} or knowledge distillation~\cite{MDoblerCVPR2023}. All the methods above are involved in training as a loss function for learning the dataset-invariant representation. On the other hand, we use the distribution gap measure to investigate its relationship with \emph{post-training} performance.\smallskip

\noindent{\bf Exploring proper uses of synthetic data.} It is challenging to expect effectiveness in training with synthetic data without adequately addressing the domain gaps with real-world test sets. One category of leveraging synthetic data in training employs special processing to reduce domain gaps when generating synthetic data, \eg, incorporating some real-world components (texture, background)~\cite{XPengICCV2015,ZWangCVPR2020,ZWangECCV2022,SBokaniaECCV2022,LLiCVPR2023} and cloning real sets~\cite{GRosCVPR2016,XLiuCVPR2023,LZengCVPR2023}. Synthetic data created by simply injecting noise~\cite{GLiCVPR2023}, messiness~\cite{QWeiCVPR2023}, or simple-shape objects such as rain~\cite{YBaECCV2022}, is relatively free of the domain gap. There also exist other methods~\cite{CWuECCV2022,MNiICLR2023} to bridge the domain gap relying on recently emerging high-performance image generators (\eg, CLIP~\cite{ARadfordICML2021}, VQ-GAN~\cite{PEsserCVPR2021}). 

Unfortunately, the aforementioned methods do not provide a comprehensive solution for reducing the domain gap. Among more general solutions recently developed, some methods mitigate the domain gap rather than completely reducing it via creating \emph{easily generalizable} feature embeddings instead of raw data~\cite{JSuECCV2022,NKumarCVPR2023}, or adjusting the ratio with the real data during training~\cite{GRosCVPR2016,SRichterECCV2016,HLeeJSTARS2021}. The recently introduced PTL~\cite{YShenCVPR2023} is a method that iteratively selects subsets of synthetic data while accounting for domain gaps, resulting in significant performance gains in general detection tasks.\smallskip

\noindent{\bf Analyzing properties of synthetic datasets.} There are many studies that have analyzed synthetic data in various aspects, such as safety/reliability~\cite{OZendelIJCV2017,RHeICLR2023}, diversity~\cite{JGaoICLR2023,RHeICLR2023}, density/coverage~\cite{MNaeemICML2020,JHanICLR2023}, \etc. The impact of using synthetic datasets has been analyzed according to the scalability~\cite{YLiuECCV2022,MSariyildizCVPR2023} or variation factors used to build the dataset~\cite{HTangCVPR2023}. \cite{ZLiCVPR2023} observes accuracy in same-domain and cross-domain tasks in the Visual Question Answering (VQA) task to figure out the transfer capability of synthetic data. While the aforementioned work performed these analyzes on specific synthetic datasets, we have carried out more general and comprehensive analyses on various aspects.
\section{Methodology}
\label{sec:method}

Our primary goal is to conduct a comprehensive study to find the environment that maximizes the two expected benefits of using synthetic data: i) improving performance and ii) leading the model to acquire domain generalization ability. To fulfill this goal, in particular, to ensure that the second benefit is realized, we first introduce how to theoretically measure the distribution gap between train and test sets in a cross-domain task. Then, we introduce a recently introduced method that provides a simple yet effective way to leverage synthetic images in training, \ie, PTL~\cite{YShenCVPR2023}. PTL was remarkably better at providing detection accuracy and acquiring domain generalizability compared to other counterparts that also leverage synthetic images (\eg, naive merge and pretrain-finetune) in aerial-view human detection.\footnote{One might consider domain adaptation methods in leveraging synthetic data in training. Generally, domain adaptation deals with cases where labels of the target dataset are not available. Therefore, it is not suitable for our task that uses labels from real datasets.} We found a strategy to reduce PTL's training time crucial to completing large volumes of comprehensive experiments.

\subsection{Measuring Distribution Gap}

\noindent{\bf Modeling the dataset with multivariate Gaussian distribution.} The distribution of a dataset for a specific category can be modeled as a multivariate Gaussian distribution in the representation space of a detector trained on the dataset if the following two conditions are satisfied: i) the detector takes the form of sigmoid-based outputs and ii) the representation space is built with the output of the penultimate layer of the detector.\footnote{This modeling is proven in the supplementary material.} Specifically, let ${\bf x}\in\mathcal{X}$ and $y=\{y_c\}_{c=1,\cdots,C}\in\mathcal{Y},~y_c\in\{0,1\}$ be an input and its categorical label, respectively. Then, the representation for the category $c$ can be expressed as follows:
\begin{equation}
    P(f({\bf x})|y_c=1)=\mathcal{N}(f({\bf x})|\mu_c,\Sigma_c),\label{eq:gaussian}
\end{equation}
where $f(\cdot)$ denotes the output of the penultimate layer of the detector. $\mu_c$ and $\Sigma_c$ are the mean and the covariance (\ie, two parameters defining the multivariate Gaussian distribution) of the representation for the category $c$, respectively.\footnote{Hereafter, since the target object is human only in this paper, we use terms without subscript $c$, meaning a specific category, throughout the paper.} These parameters can be computed empirically with the dataset.\smallskip

\noindent{\bf Distribution gap to the new dataset.} To measure the distribution gap between two datasets (a reference dataset $\mathcal{D}_r$ and a new dataset $\mathcal{D}_\text{new}$), we used the cross-entropy, which statistically measures how a given distribution is different from the reference distribution. (i.e., $\mathcal{H}(P,Q)=-\int_\mathcal{X}{p({\bf x})\ln q({\bf x})d{\bf x}}$, where $p$ and $q$ denote the probability densities of two distributions $P$ and $Q$, respectively. Here, $Q$ is the reference distribution.) We regard the dataset where the representation space is built as the reference dataset and calculate the distribution gap from the reference dataset to the new dataset in the representation space.

As demonstrated in the previous section, the probability density of $\mathcal{D}_r$ can be expressed as a multivariate Gaussian distribution, as in eq~\ref{eq:gaussian}. Since $\mathcal{D}_n$ is not involved in detector training, we regard the probability density of the dataset as a mixture model where each component indicating a single element of the dataset takes the form of a Dirac delta function, as follows:
\begin{equation}
    p({\bf x})\!=\!\frac{1}{|\mathcal{D}_n|}\!\!\sum_{{\bf x}'\in \mathcal{D}_n}\!\!\!\delta({\bf x}-{\bf x}'),\label{eq:p}
\end{equation}
where $\delta ({\bf x})$ is a Dirac delta function whose value is zero everywhere except at ${\bf x}=0$ and whose integral over $\mathcal{X}$ which is the entire space of ${\bf x}$ is one (\ie, $\int_\mathcal{X}{{\bf x}d{\bf x}}=1$).

Using the two probability densities of $P$ and $Q$ defined in eq~\ref{eq:gaussian} and \ref{eq:p}, cross-entropy can be derived\footnote{This derivation can be found in the supplementary material.}, as:
\begin{equation}
    \mathcal{H}(P,Q)\!=\!\frac{1}{2|\mathcal{D}_n|}\!\!\sum_{{\bf x}\in\mathcal{D}_n}{\!\!\!(f({\bf x})\!-\!\mu)^\top\Sigma^{-1}(f({\bf x})\!-\!\mu)}\!+\!C,\label{eq:cross_entropy}
\end{equation}
where $C$ is a constant that is not affected by $\mathcal{D}_n$. Accordingly, to quantitatively compare distribution gaps of two new datasets with respect to the reference dataset, we define a distribution gap for the new dataset by removing $C$ from the cross-entropy in~\ref{eq:cross_entropy}, as:
\begin{equation}
    d(\mathcal{D}_n|\mu,\Sigma)\!=\!\frac{1}{2|\mathcal{D}_n|}\!\!\sum_{{\bf x}\in\mathcal{D}_n}{\!\!\!(f({\bf x})\!-\!\mu)^\top\Sigma^{-1}(f({\bf x})\!-\!\mu)}.\label{eq:distribution gap}
\end{equation}
As a result, the distribution gap measure takes the form of a normalized sum of the Mahalanobis distances~\cite{PMahalanobisNISI1936} over all data in $\mathcal{D}_n$.

\subsection{Leveraging Synthetic Images in Training}

\noindent{\bf Progressive Transformation Learning (PTL).} PTL gradually expands training data by repeating two steps: i) selecting a subset of synthetic data and ii) transforming the selected synthetic images to look more realistic. This progressive strategy is used to address quality degradation of the sim2real transformation that can occur due to the large domain gap between the real and the synthetic domains.

The subset of the synthetic set is constructed by selecting more synthetic images with a closer domain gap to the training set. The \emph{sim2real transformer} is trained via a conditional GAN (specifically, CycleGAN~\cite{JZhuICCV2017}) to transform selected synthetic images to have the visual properties of the current training set. Note that two training processes for the detector and the sim2real transformer are involved in the PTL process for each iteration.\smallskip

\noindent{\bf PTL training time curtailment.} The biggest bottleneck when conducting a comprehensive study with PTL is the lengthy training time ({\it e.g.}, 10 and 16 hours for PTL training under the Vis-20/Vis-50\footnote{We refer to the setting using $N$ images of the VisDrone dataset as a real training set as `Vis-N' throughout all experiments, e.g., Vis-20.} setups, respectively). Sim2real transformer training takes up the largest portion of PTL training time, followed by detector training. Originally, these two training processes start from scratch for every PTL iteration because the training set changes with every PTL iteration. Instead of this time-consuming training approach, we consider the \emph{tuning-from-previous-iteration} strategy, where the model to be trained is initialized from the model learned in the previous PTL iteration, with fewer training iterations.

\begin{table}[t]
\caption{{\bf Training time curtailment via `tuning-from-previous-iteration' strategy.} $f_t$ and $f_d$ represent the sim2real transformer and the detector, respectively. The training time and the accuracy are measured with wall-clock time in hours and AP@[.5:.95], respectively. The number in parentheses in `time' indicates the relative time compared to the original PTL training.}
\label{tab:time_reduction}
\vspace{-0.3cm}
\centering
\resizebox{\linewidth}{!}{%
\setlength{\tabcolsep}{10.0pt}
\renewcommand{\arraystretch}{1.2}
\begin{tabular}{l|l|ccc|l|ccc}
& \multicolumn{4}{c|}{Vis-20} & \multicolumn{4}{c}{Vis-50} \\
from-prev-iter & \multicolumn{1}{c|}{time} & Vis & Oku & ICG & \multicolumn{1}{c|}{time} & Vis & Oku & ICG \\\hline
\textcolor{gray}{Original} & ~\textcolor{gray}{10} & \textcolor{gray}{1.94} & \textcolor{gray}{7.45} & \textcolor{gray}{7.22} & ~\textcolor{gray}{16} & \textcolor{gray}{2.85} & \textcolor{gray}{11.46} & \textcolor{gray}{7.27} \\
$f_{t}$ & 6.5 \textcolor{teal}{($\times$0.65)} & 1.95 & 7.01 & 8.93 & ~~~9 \textcolor{teal}{($\times$0.56)} & 2.78 & 11.52 & 9.90 \\
$f_t$ \& $f_d$ & 5.5 \textcolor{teal}{($\times$0.55)} & 1.61 & 6.03 & 4.71 & 7.5 \textcolor{teal}{($\times$0.47)} & 2.43 & ~~9.81 & 6.78 \\
\end{tabular}
}
\end{table}

Table~\ref{tab:time_reduction} shows the change in training time and accuracy with this time-curtailing strategy on the Vis-20/50 setups. When using this strategy for sim2real transformer training, it was effective as the time was significantly reduced (at least $\times$0.65) without loss of accuracy. On the other hand, applying this strategy in training the detector (with the sim2real transformer training) has a negative impact as accuracy is significantly reduced but time curtailment is not as great as that achieved with the sim2real transformer solely. Based on this comparison, we used the \emph{tune-from-previous-iteration} strategy in training the sim2real transformer only throughout the following experiments.
\section{Experimental Settings}
\label{sec:setup}

\noindent{\bf Task and dataset.} For our task, we use \textit{N}-shot detection tasks, where a limited number of \textit{N} images, are used for training, and cross-domain detection tasks, where the same domain images are not available in training. \textit{N}-shot detection, and cross-domain detection are tasks required in a problem space where real data is extremely scarce; thus synthetic data is in high demand.

We use five datasets built for aerial-view human detection for real datasets: VisDrone~\cite{PZhuTPAMI2021}, Okutama-Action~\cite{MBarekatainCVPRW2017}, ICG~\cite{ICGlink}, HERIDAL~\cite{DBozicStulicIJCV2019}, and SARD~\cite{SSambolekAccess2021}. VisDrone is used as a training set, and all five datasets are used as test sets. For a synthetic data pool, we use the Archangel-Synthetic dataset~\cite{YShenAccess2023}.\smallskip

\noindent{\bf Evaluation metrics.} We use MS COCO style AP@.5 and AP@[.5:.95] as evaluation metrics in our study. Due to space limitation, only AP@[.5:.95] is reported in the main manuscript while AP@.5 values are additionally reported in the supplementary material. We perform three runs and report the average value to address potential random effects in the \textit{N}-shot detection task.
\section{Results and Analysis}
\label{sec:analysis}

\subsection{A Study on the Impact of Real Data}

For the first study, we explore the scalability behavior of real data regarding the two impacts of using synthetic data in training: i) increasing detection accuracy and ii) reducing the distribution gaps. Specifically, these two aspects are compared among four cases using real data with different quantities (\ie, 20, 50, 100, and 200).\smallskip

\begin{figure}[t]
\centering
\begin{subfigure}{0.24\linewidth}
\includegraphics[trim=10mm 0mm 10mm 0mm,clip,width=\linewidth]{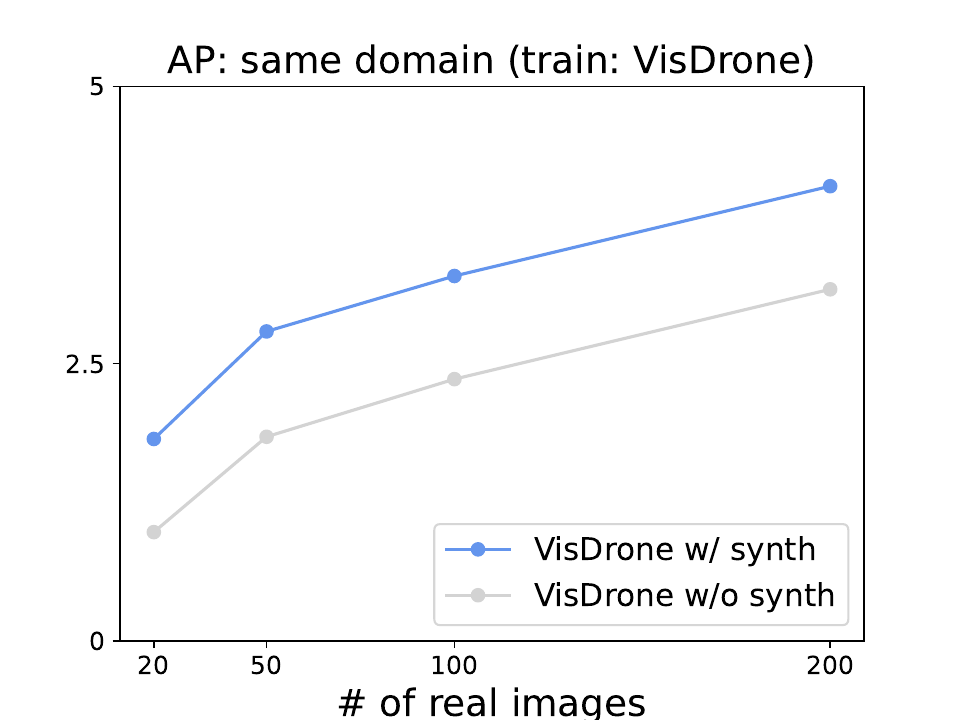}
\vspace{-0.3cm}
\caption{}
\label{fig:num_real_same_domain}
\end{subfigure}
\begin{subfigure}{0.24\linewidth}
\includegraphics[trim=10mm 0mm 10mm 0mm,clip,width=\linewidth]{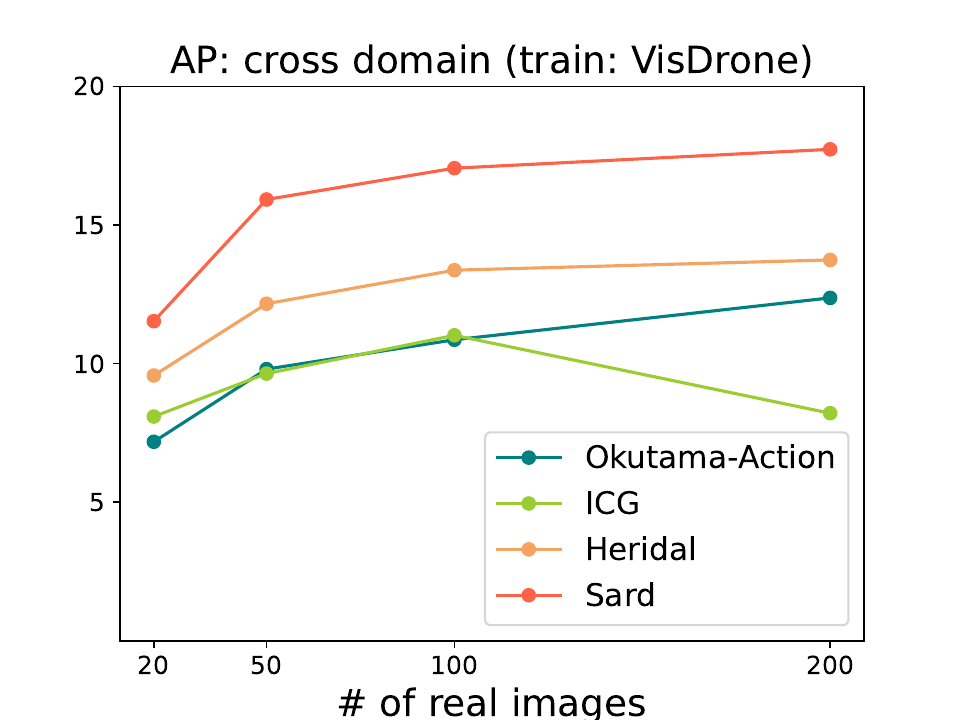}
\vspace{-0.3cm}
\caption{}
\label{fig:num_real_cross_domain}
\end{subfigure}
\begin{subfigure}{0.24\linewidth}
\includegraphics[trim=10mm 0mm 10mm 0mm,clip,width=\linewidth]{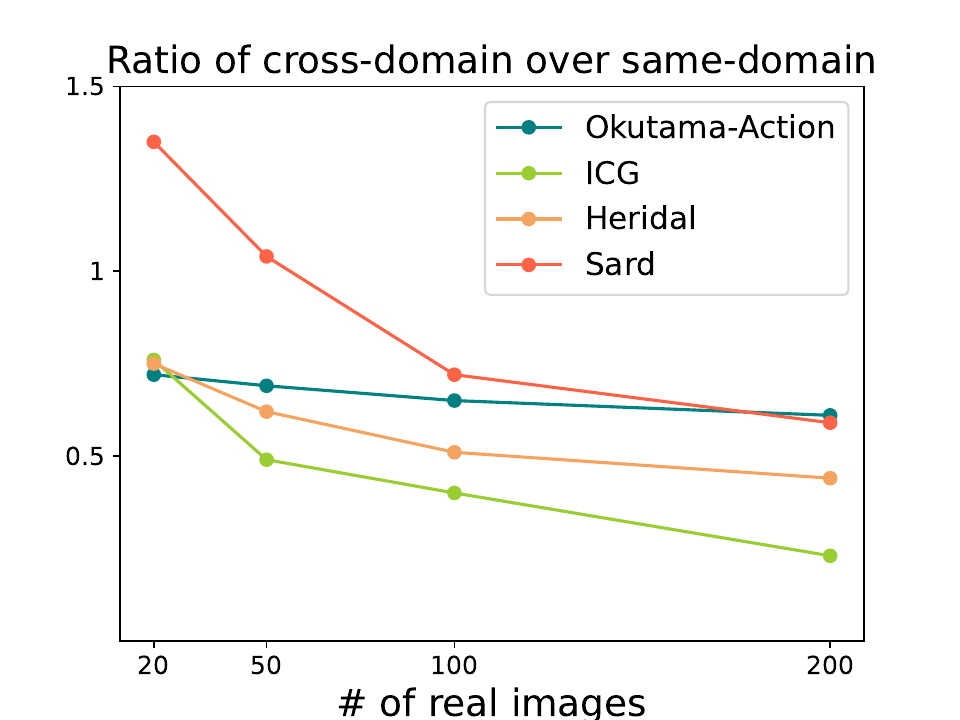}
\vspace{-0.3cm}
\caption{}
\label{fig:num_real_ratio}
\end{subfigure}
\vspace{-0.3cm}
\caption{{\bf Accuracy with the size of real dataset.} (b) and (c) show the accuracy when synthetic images are used in training.}
\label{fig:scaling_behavior_of_realimg}
\end{figure}

\noindent{\bf Analysis in terms of accuracy.} In same-domain tasks (Fig.~\ref{fig:num_real_same_domain}), detection accuracy unsurprisingly increased proportionally with the size of the real dataset, regardless of whether or not synthetic data is used. The use of synthetic data consistently increases accuracy irrespective of the size of the real data set. Interestingly, adopting a larger real dataset yields better accuracy even in most cross-domain tasks (Fig.~\ref{fig:num_real_cross_domain}). These trends indicate that it is essential to use synthetic data with real data, including cross-domain tasks.

Using a large number of real images (\ie, 200), on the other hand, results in little increase or adversely affects accuracy compared to using fewer images. The rationale behind this notable observation can be accounted for with Fig.~\ref{fig:num_real_ratio}, which illustrates the accuracy ratio of a cross-domain task to a same-domain task. Here, the two tasks use different training sets (for the cross-domain task, we use VisDrone as the training set.) but are evaluated on the same test set. When the number of real images is small, the cross-domain presents similar or better accuracy than the same-domain. However, the effectiveness of the cross-domain continues to decrease as the number of real images increases. 

These analyses strongly indicate that \emph{synthetic data is effective in both same-domain and cross-domain tasks, particularly the impact of synthetic images is more significant when the amount of real data is small and then it continues to diminish as the amount of real data increases}. These findings fully confirm that synthetic data can greatly enhance learning in the data scarcity realm, where real data are hard to obtain if adequately integrated into learning.\smallskip

\begin{table}[t]
\caption{{\bf Distribution gaps of various datasets} from VisDrone. The better accuracy between using and not using synthetic images is shown in bold.}
\label{tab:distribution_gap}
\vspace{-0.3cm}
\centering
\resizebox{\linewidth}{!}{%
\setlength{\tabcolsep}{12.0pt}
\renewcommand{\arraystretch}{1.2}
\begin{tabular}{l|c|rr|rr|rr|rr}
& & \multicolumn{2}{c|}{Okutama} & \multicolumn{2}{c|}{ICG} & \multicolumn{2}{c|}{HERIDAL} & \multicolumn{2}{c}{SARD} \\
setup & w/ synth & \multicolumn{1}{c}{50\%} & \multicolumn{1}{c|}{all} & \multicolumn{1}{c}{50\%} & \multicolumn{1}{c|}{all} & \multicolumn{1}{c}{50\%} & \multicolumn{1}{c|}{all} & \multicolumn{1}{c}{50\%} & \multicolumn{1}{c}{all} \\\hline
\multirow{2}{*}{Vis-20} & & 90.6 & 4,627.1 & 63.0 & 2,372.9 & 151.0 & 6,524.2 & 132.2 & 5,083.5 \\
& \textcolor{red}{\checkmark} & {\bf 35.1} & {\bf 487.9} & {\bf 39.4} & {\bf 518.2} & {\bf 40.8} & {\bf 406.9} & {\bf 36.8} & {\bf 420.3} \\\hline
\multirow{2}{*}{Vis-50} & & 40.3 & {\bf 274.0} & 32.8 & {\bf 285.6} & 62.7 & {\bf 340.5} & 40.7 & {\bf 323.0} \\
& \textcolor{red}{\checkmark} & {\bf 31.7} & 431.1 & {\bf 31.9} & 368.7 & {\bf 35.3} & 639.1 & {\bf 33.6} & 958.4 \\\hline
\multirow{2}{*}{Vis-100} & & 36.3 & 119.4 & 32.3 & {\bf 103.1} & 58.5 & 190.0 & 63.2 & {\bf 222.0} \\
& \textcolor{red}{\checkmark} & {\bf 23.2} & {\bf 82.7} & {\bf 28.0} & 182.4 & {\bf 27.7} & {\bf 177.9} & {\bf 27.6} & 300.7 \\\hline
\multirow{2}{*}{Vis-200} & & 27.4 & 167.1 & 29.3 & {\bf 103.5} & 40.6 & {\bf 130.2} & 38.8 & {\bf 122.8} \\
& \textcolor{red}{\checkmark} & {\bf 19.8} & {\bf 159.5} & {\bf 22.5} & 207.7 & {\bf 27.2} & 305.4 & {\bf 25.8} & 424.5 \\
\end{tabular}
}
\end{table}

\noindent{\bf Analysis in terms of distribution gaps.} Table~\ref{tab:distribution_gap} provides distribution gaps for the various test sets with and without synthetic data in a cross-domain setup. It is observed that for some cases, the use of synthetic data (`all' in the Table) unexpectedly increases the distribution gap compared to the cases without synthetic data. On the other hand, the distribution gap over half of the test images located closer to the reference dataset in terms of the Mahalanobis distance (`50\%' in the Table) decreased as expected when synthetic data is added.

\begin{wrapfigure}{r}{0.5\textwidth}
\vspace{-0.7cm}
\centering
\begin{subfigure}{.49\linewidth}
\includegraphics[width=\linewidth,trim=5mm 0mm 10mm 0mm,clip]{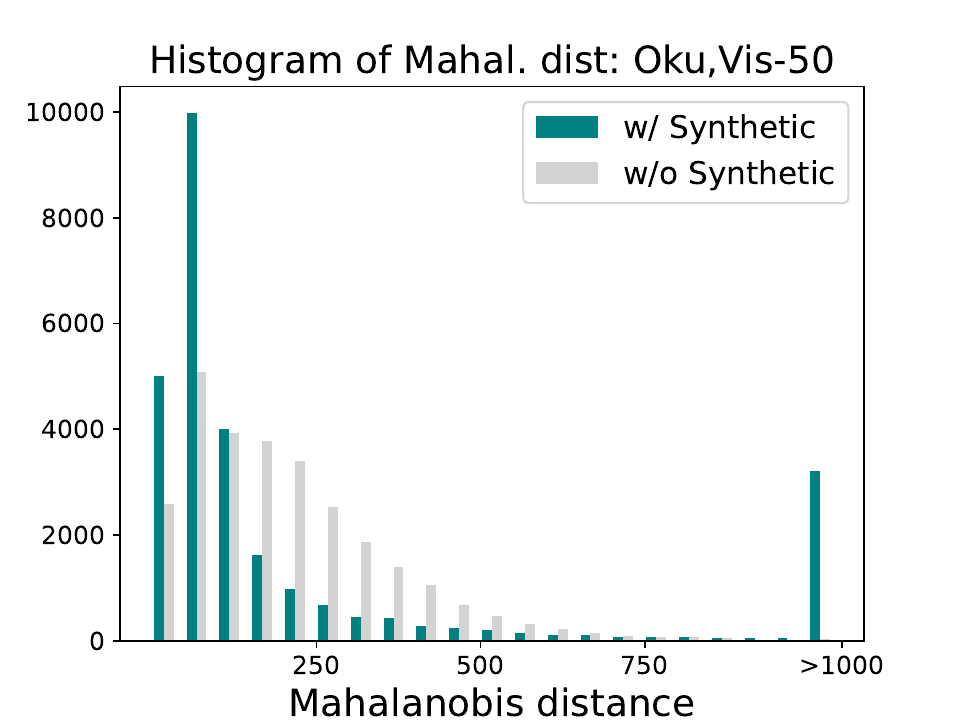}
\caption{}
\label{fig:histogram}
\end{subfigure}
\begin{subfigure}{.49\linewidth}
\includegraphics[width=\linewidth,trim=5mm 0mm 10mm 0mm,clip]{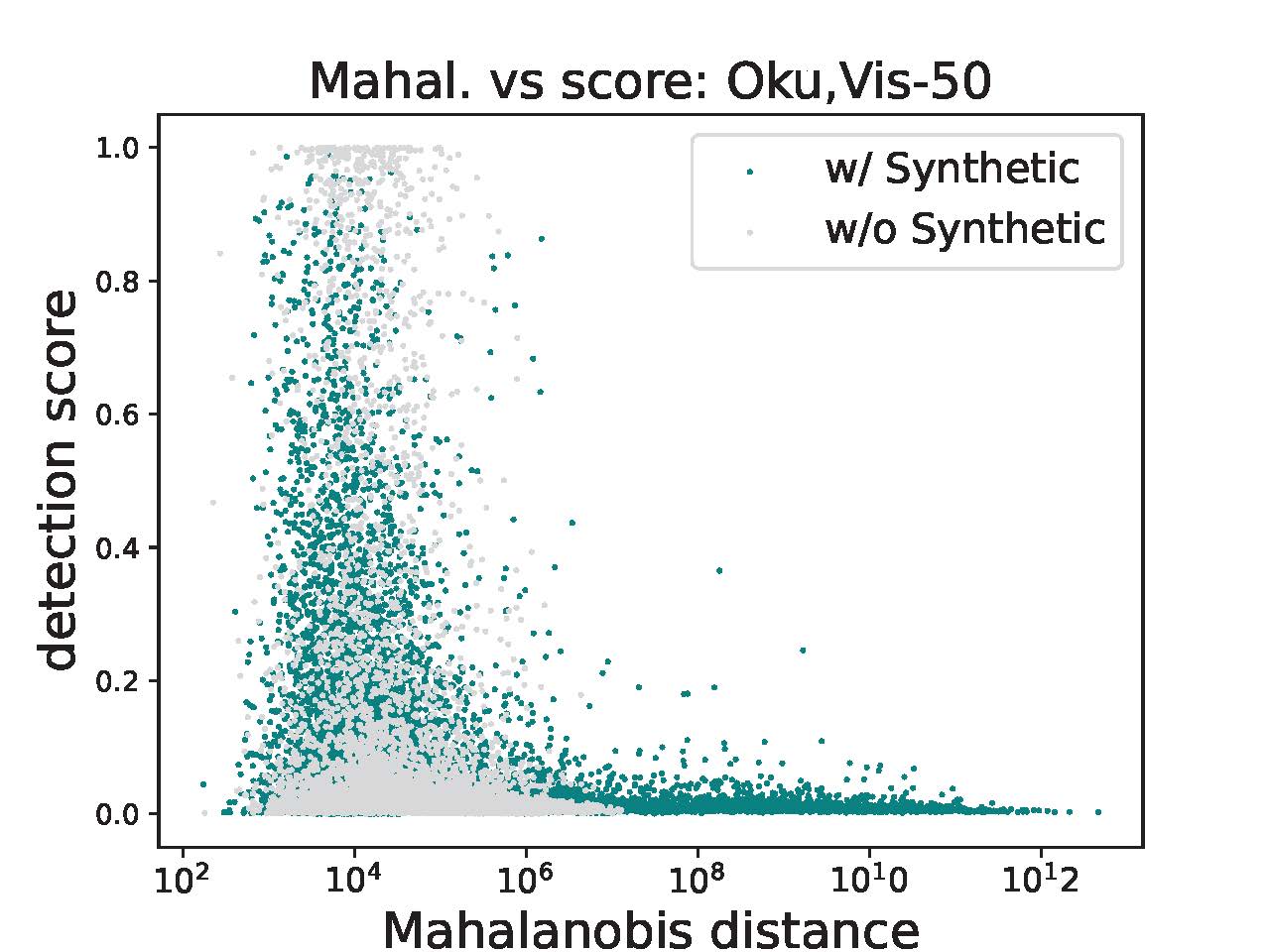}
\caption{}
\label{fig:scatter}
\end{subfigure}
\vspace{-0.7cm}
\caption{{\bf Change in distribution gap when synthetic images are used} for each image of the Okutama-Action dataset under the Vis-50 setup.}
\end{wrapfigure}

To investigate the change in the distribution gap in detail, we compare histograms representing the number of test images with respect to the Mahalanobis distance with and without synthetic data (Fig.~\ref{fig:histogram}). Including synthetic data effectively reduces the Mahalanobis distance for most of the test images, yet the number of outliers with extremely large Mahalanobis distances also increases. We also compare how the distribution of test images with respect to detection scores \vs.~Mahalanobis distances differ depending on whether synthetic data is included or not (Fig.~\ref{fig:scatter}). When using synthetic data in training, a majority of test images come with high detection confidence and small Mahalanobis distances, contributing to a detection accuracy increase. However, the test images with large Mahalanobis distances and low detection confidence also appear more frequently. The analysis indicates that \emph{a majority of synthetic data serves to improve the detector's ability for most test images in general with some exceptions of outlier images in a cross-domain setup}.

\subsection{A Study on the Impact of Synthetic Data}

For the second study, we explored the scalability behavior of synthetic data on the two impacts of using the synthetic data in training, mentioned in the first study. Specifically, we compare five cases with no synthetic images, 100, 500, 1000, and 2000 synthetic images in training in terms of the accuracy and distribution gap.\smallskip

\begin{figure}[t]
\begin{subfigure}{0.24\linewidth}
\includegraphics[trim=10mm 0mm 10mm 0mm,clip,width=\linewidth]{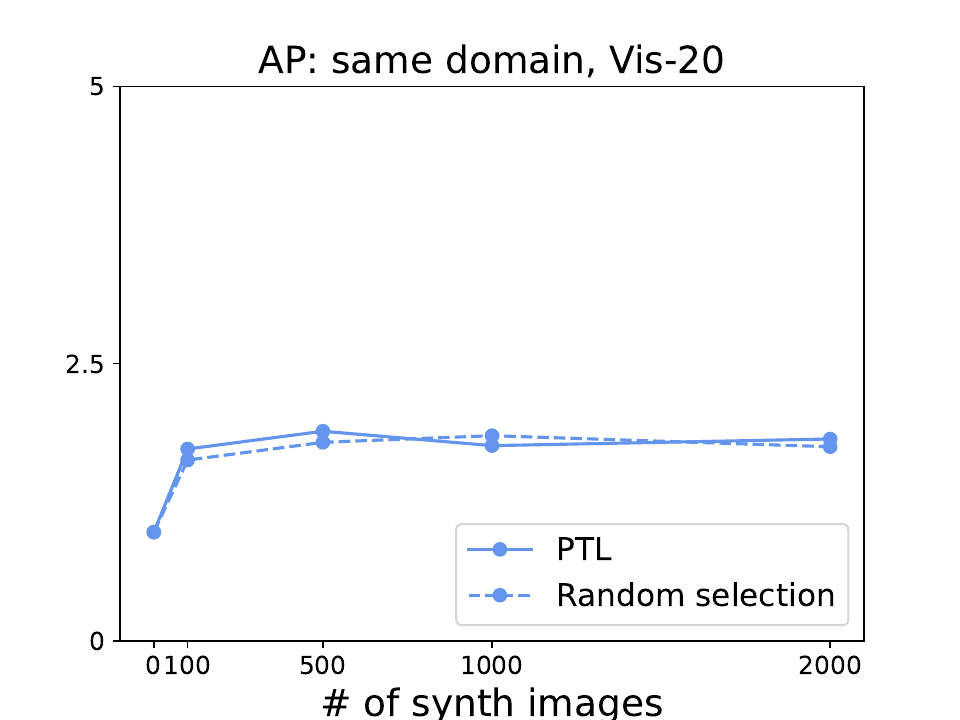}
\label{fig:num_synth_same_domain_vis_20}
\end{subfigure}
\hfill
\begin{subfigure}{0.24\linewidth}
\includegraphics[trim=10mm 0mm 10mm 0mm,clip,width=\linewidth]{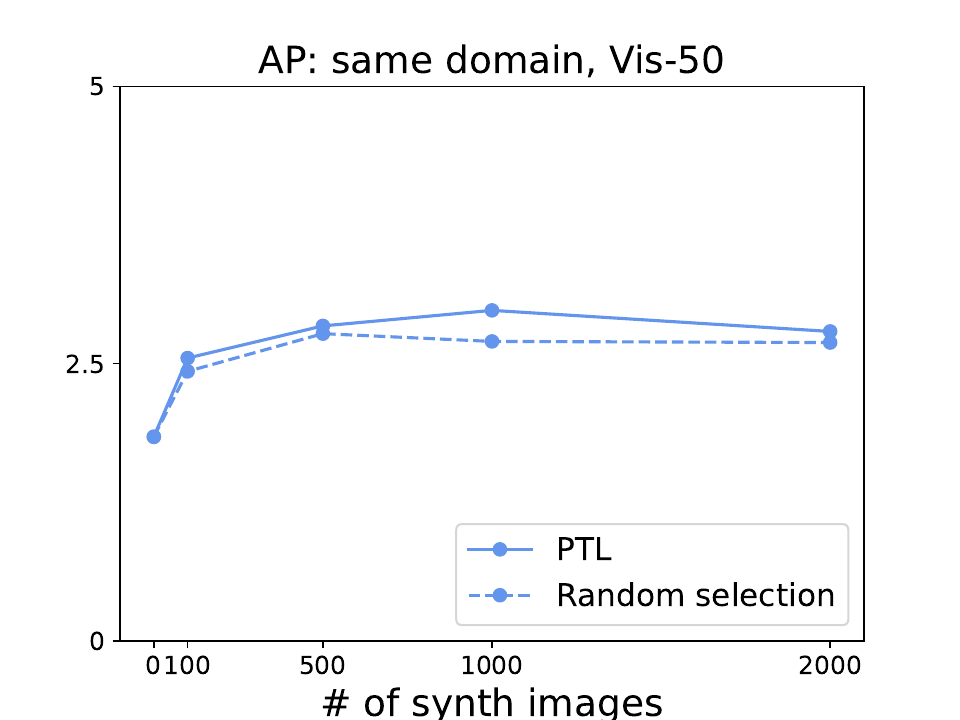}
\label{fig:num_synth_same_domain_vis_50}
\end{subfigure}
\hfill
\begin{subfigure}{0.24\linewidth}
\includegraphics[trim=10mm 0mm 10mm 0mm,clip,width=\linewidth]{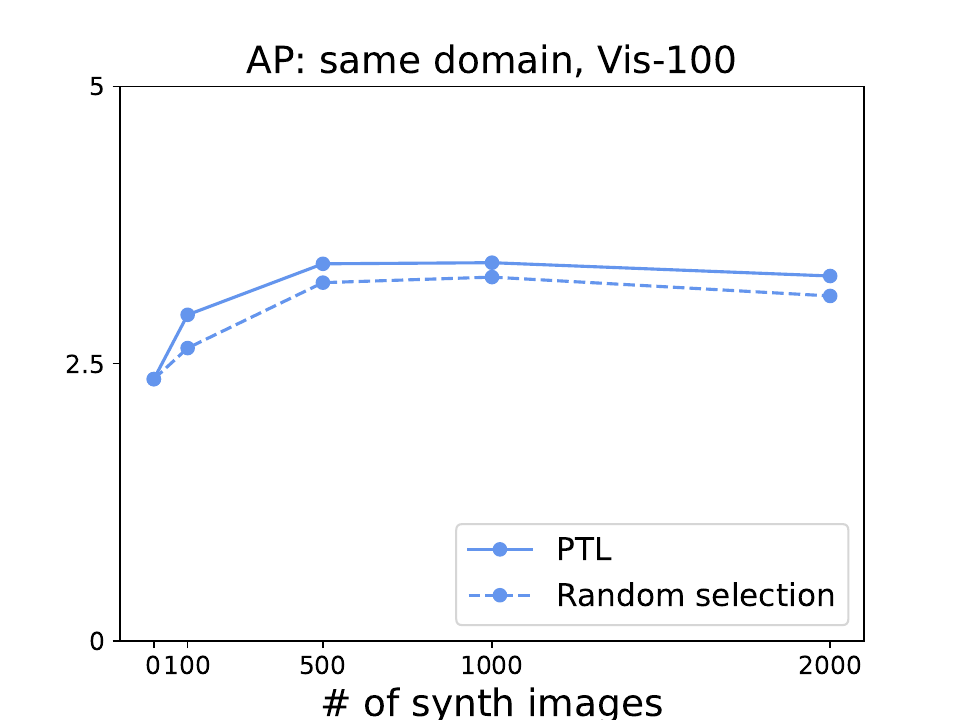}
\label{fig:num_synth_same_domain_vis_100}
\end{subfigure}
\hfill
\begin{subfigure}{0.24\linewidth}
\includegraphics[trim=10mm 0mm 10mm 0mm,clip,width=\linewidth]{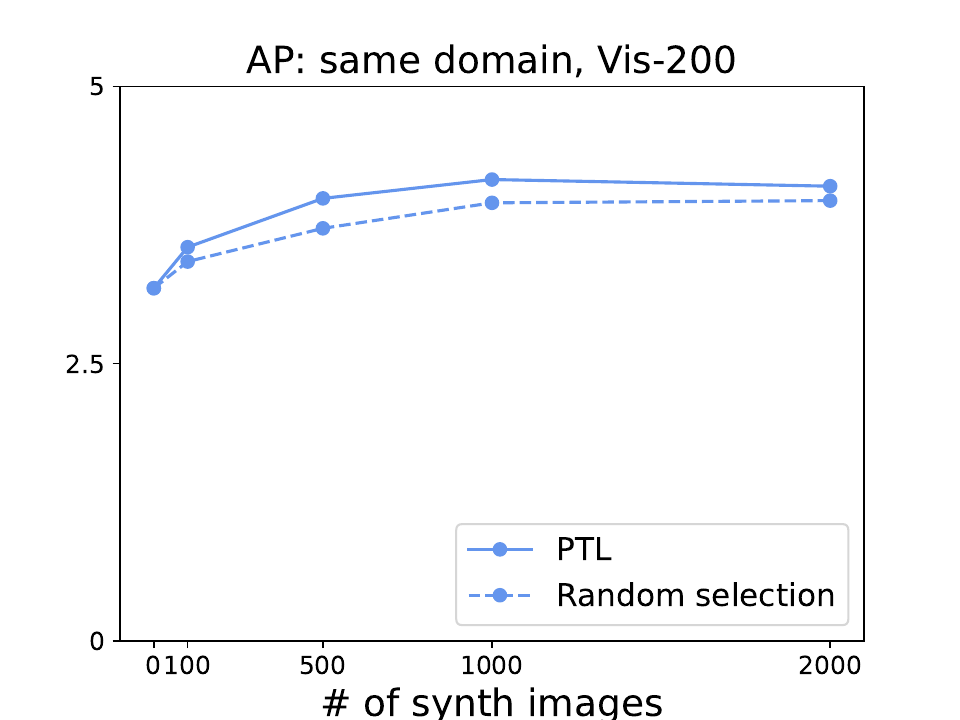}
\label{fig:num_synth_same_domain_vis_200}
\end{subfigure}
\\
\begin{subfigure}{0.24\linewidth}
\vspace{-0.4cm}
\includegraphics[trim=10mm 0mm 10mm 0mm,clip,width=\linewidth]{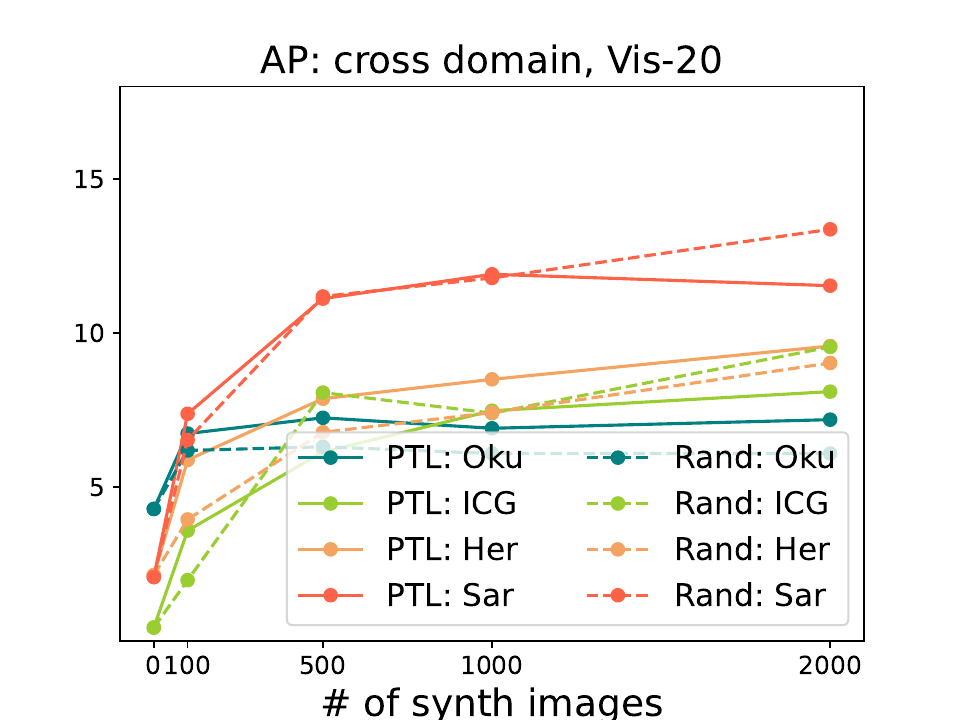}
\label{fig:num_synth_cross_domain_vis_20}
\end{subfigure}
\hfill
\begin{subfigure}{0.24\linewidth}
\vspace{-0.4cm}
\includegraphics[trim=10mm 0mm 10mm 0mm,clip,width=\linewidth]{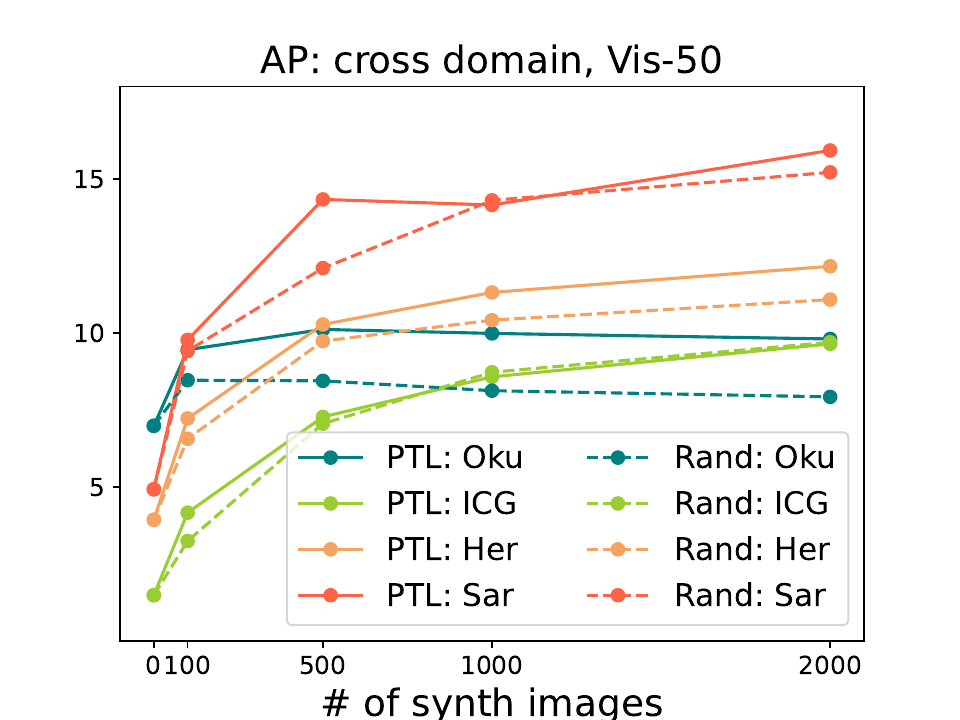}
\label{fig:num_synth_cross_domain_vis_50}
\end{subfigure}
\hfill
\begin{subfigure}{0.24\linewidth}
\vspace{-0.4cm}
\includegraphics[trim=10mm 0mm 10mm 0mm,clip,width=\linewidth]{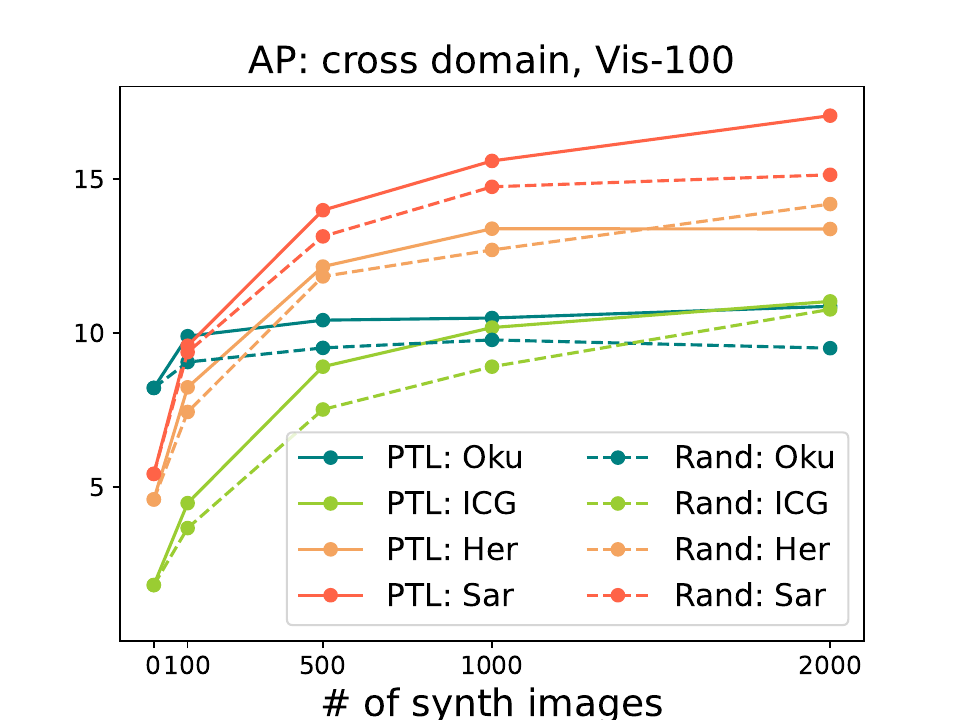}
\label{fig:num_synth_cross_domain_vis_100}
\end{subfigure}
\hfill
\begin{subfigure}{0.24\linewidth}
\vspace{-0.4cm}
\includegraphics[trim=10mm 0mm 10mm 0mm,clip,width=\linewidth]{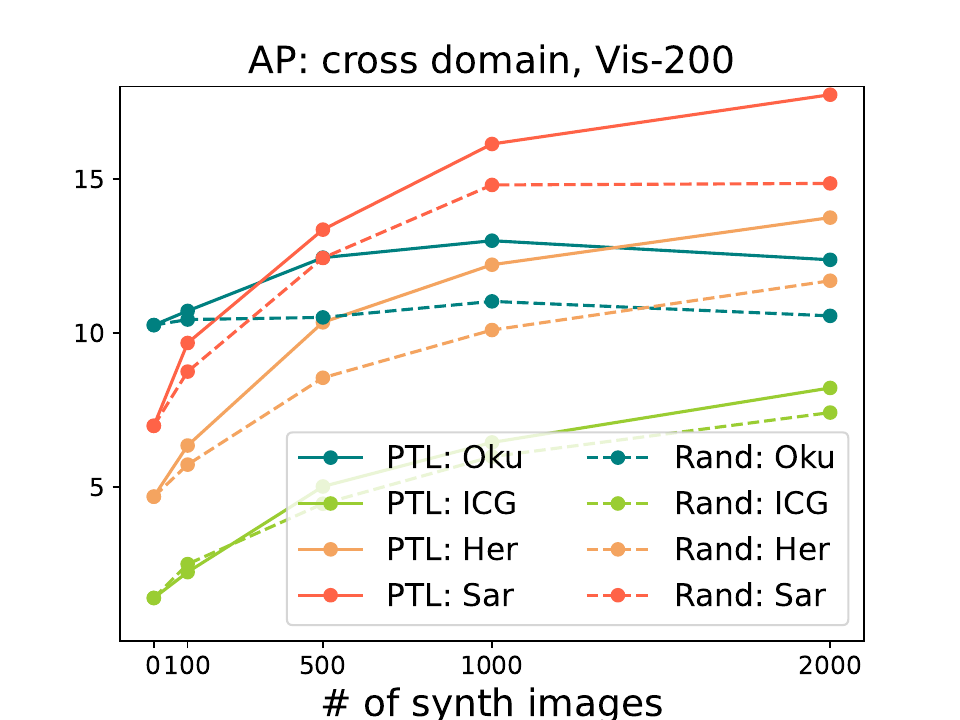}
\label{fig:num_synth_cross_domain_vis_200}
\end{subfigure}
\vspace{-0.6cm}
\caption{{\bf Accuracy with the size of the synthetic dataset.} Plots in the top and bottom rows show APs in same-domain and cross-domain tasks, respectively.}
\label{fig:scaling_behavior_of_synthimg}
\end{figure}

\noindent{\bf Analysis in terms of accuracy.} In this study, as PTL, our standard method of using synthetic images in training, gradually increases the number of synthetic images as training progresses, we investigate the scaling behavior of synthetic data by comparing models at different training checkpoints. To exclude the potential methodological influence of PTL in this general investigation, we also consider a random selection method, which randomly selects the same number of synthetic images as those used in PTL for training after applying the sim2real transformation.

In Fig.~\ref{fig:scaling_behavior_of_synthimg}, two notable observations can be found regarding accuracy: i) in all setups, including same-domain and cross-domain, regardless of the method for synthetic data integration, the accuracy continues to increase while the rate of accuracy increase decreases as more synthetic images are used in training, and ii) as more real images are used in training, the checkpoint where the increase in accuracy rapidly diminishes usually occurs when a relatively large number of synthetic images are used. These observations indicate that \emph{the impact of synthetic data continues to decrease as more synthetic images are included, but the capacity to use more synthetic data without sacrificing accuracy is expanded as more real data is used}.\smallskip

\noindent{\bf Analysis in terms of distribution gaps.} In Table~\ref{tab:distribution_gap_with_various_synthetic_images} that shows the scalability behavior of synthetic data with respect to the distribution gap, it is observed that the distribution gap mostly continues to decrease while the rate of change also decreases as more synthetic images are used in training. This is aligned to that of the previous analysis regarding accuracy.

\begin{wrapfigure}{r}{0.5\textwidth}
\vspace{0.2cm}
\begin{subfigure}{0.47\linewidth}
\includegraphics[trim=10mm 0mm 10mm 0mm,clip,width=\linewidth]{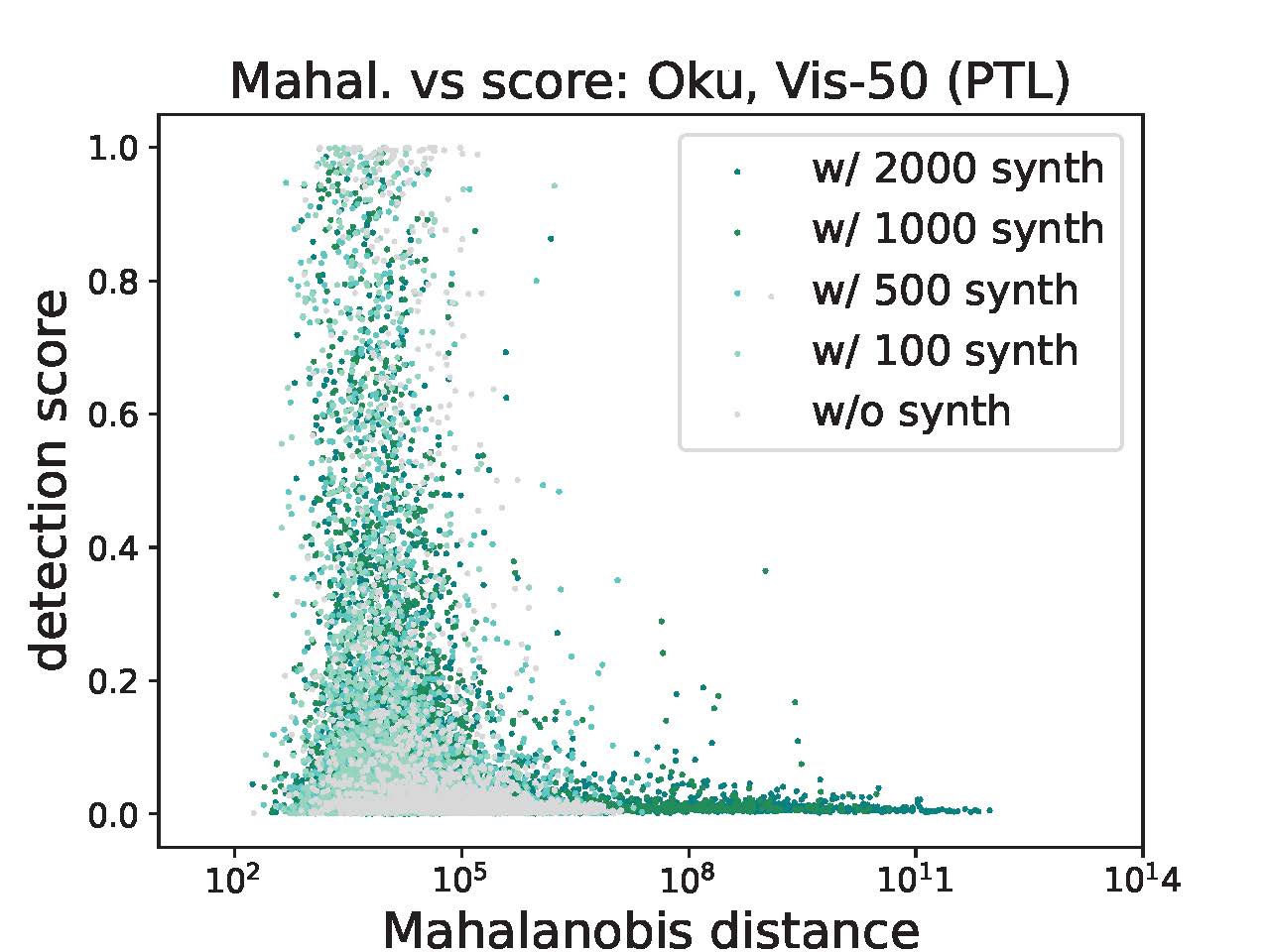}
\label{fig:scatter_PTL}
\end{subfigure}
\hfill
\begin{subfigure}{0.47\linewidth}
\includegraphics[trim=10mm 0mm 10mm 0mm,clip,width=\linewidth]{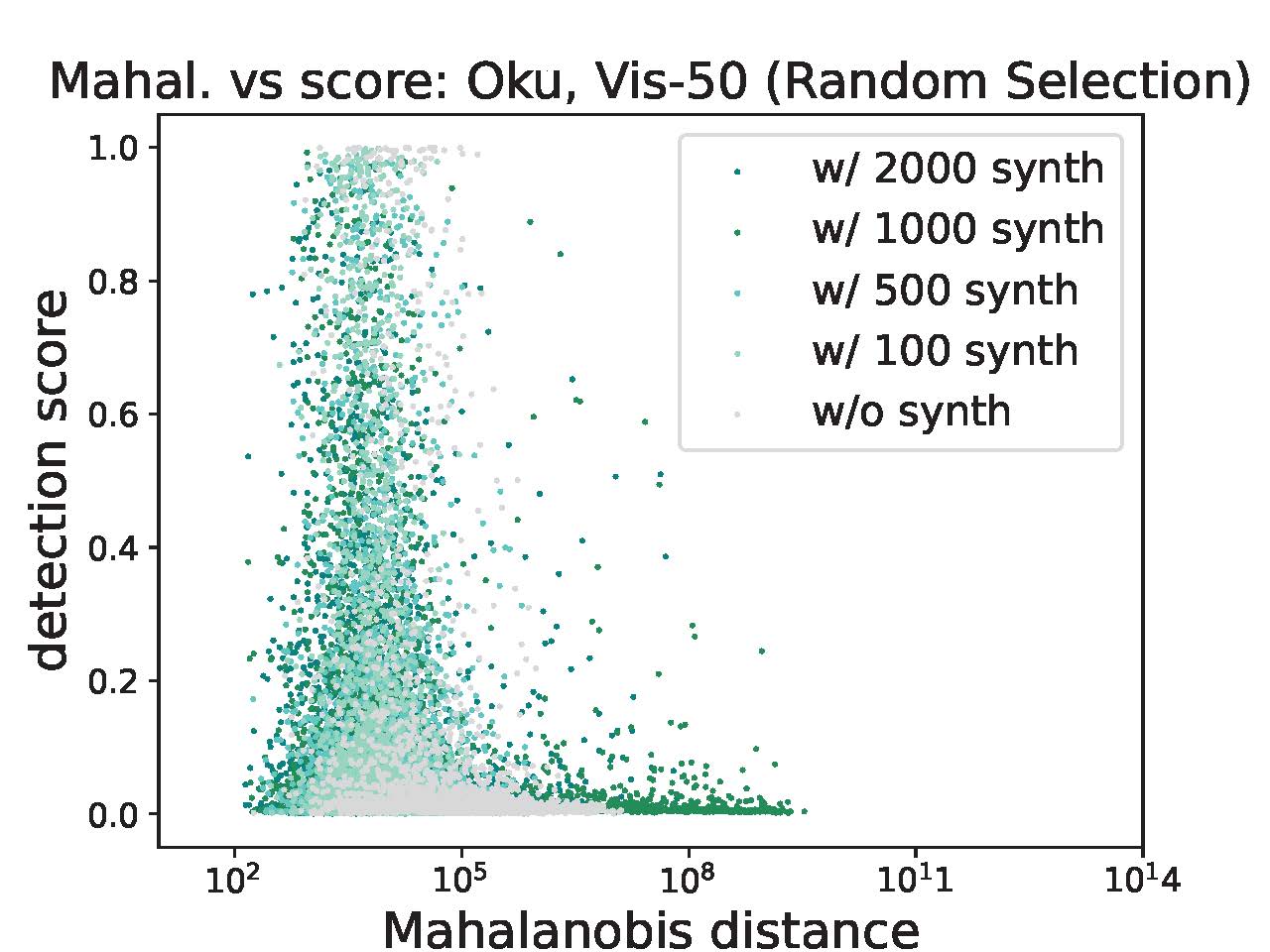}
\label{fig:scatter_random}
\end{subfigure}
\vspace{-0.5cm}
\caption{{\bf Detection accuracy-distribution gap scatter plot with various numbers of synthetic images.} The left and right plots are made with PTL and a random selection, respectively, using the Okutama-Action dataset under the Vis-50 setup. Darker dots represent test data when using more synthetic images for training.}
\label{fig:scatter_of_synthimg}
\vspace{-0.6cm}
\end{wrapfigure}

In Fig~\ref{fig:scatter_of_synthimg}, we can figure out how the distribution of test images for the detection score and Mahalanobis distance changed with the number of synthetic images used in training. Two notable observations are presented in the scatter plots: i) samples with high detection score ($>$0.2) appear more often as more synthetic images are used, and ii) samples with large Mahalanobis distance also appear more frequently when using a very large number of synthetic images (\ie, 2000).\smallskip

\noindent

\noindent{\bf PTL \textit{vs}. random selection.} In the previous analysis, two conflicting observations were found regarding the comparison between PTL and random selection. Firstly, PTL consistently provides better accuracy than random selection, regardless of training settings, for both the same domain and cross-domain tasks (Fig.~\ref{fig:scaling_behavior_of_synthimg}). On the other hand, random selection is generally more effective in reducing the distribution gap using synthetic data than PTL (Tab.~\ref{tab:distribution_gap_with_various_synthetic_images}). When selecting synthetic images from the synthetic pool, PTL focuses more on synthetic images with similar characteristics to the reference dataset more frequently than simply increasing the generalization ability of the training set. As this selection strategy is proposed to prevent degradation of the sim2real transformation quality, using higher-quality transformed synthetic images in training has a positive impact on increasing detection accuracy.

\begin{table*}[t]
\caption{{\bf Distribution gaps with various numbers of synthetic images.} 50\% of the test images with the smallest Mahalanobis distance from the reference dataset are used for calculation.}
\vspace{-0.3cm}
\label{tab:distribution_gap_with_various_synthetic_images}
\begin{subtable}[b]{\linewidth}
\caption{PTL}
\vspace{-0.3cm}
\centering
\resizebox{\linewidth}{!}{%
\setlength{\tabcolsep}{3.0pt}
\renewcommand{\arraystretch}{1.2}
\begin{tabular}{l|rrrrr|rrrrr|rrrrr|rrrrr}
& \multicolumn{5}{c|}{Vis-20} & \multicolumn{5}{c|}{Vis-50} & \multicolumn{5}{c|}{Vis-100} & \multicolumn{5}{c}{Vis-200} \\
dataset & \multicolumn{1}{c}{0} & \multicolumn{1}{c}{100} & \multicolumn{1}{c}{500} & \multicolumn{1}{c}{1000} & \multicolumn{1}{c|}{2000} & \multicolumn{1}{c}{0} & \multicolumn{1}{c}{100} & \multicolumn{1}{c}{500} & \multicolumn{1}{c}{1000} & \multicolumn{1}{c|}{2000} & \multicolumn{1}{c}{0} & \multicolumn{1}{c}{100} & \multicolumn{1}{c}{500} & \multicolumn{1}{c}{1000} & \multicolumn{1}{c|}{2000} & \multicolumn{1}{c}{0} & \multicolumn{1}{c}{100} & \multicolumn{1}{c}{500} & \multicolumn{1}{c}{1000} & \multicolumn{1}{c}{2000} \\\hline
Okutama & 90.6 & 68.9 & 34.5 & 30.9 & 35.1 & 40.3 & 27.9 & 26.9 & 26.0 & 31.7 & 36.3 & 32.6 & 24.2 & 26.5 & 23.2 & 27.4 & 27.8 & 20.8 & 19.7 & 19.8 \\
ICG & 63.0 & 33.8 & 33.5 & 34.9 & 39.4 & 32.8 & 24.6 & 24.4 & 28.4 & 32.8 & 32.3 & 28.2 & 26.5 & 28.5 & 28.0 & 29.3 & 30.1 & 22.9 & 23.2 & 22.5 \\
HERIDAL & 151.0 & 85.7 & 40.4 & 38.0 & 40.8 & 62.7 & 82.2 & 32.5 & 31.9 & 35.3 & 58.5 & 43.7 & 34.5 & 39.1 & 27.7 & 40.6 & 46.9 & 26.8 & 31.9 & 27.2 \\
SARD & 132.2 & 101.9 & 36.5 & 37.0 & 36.8 & 40.7 & 32.8 & 31.4 & 30.0 & 33.6 & 63.2 & 41.0 & 32.6 & 46.4 & 27.6 & 38.8 & 40.4 & 27.8 & 32.5 & 25.8 \\
\multicolumn{21}{c}{}\\
\end{tabular}
}\vfill
\end{subtable}
\begin{subtable}[b]{\linewidth}
\caption{Random selection}
\vspace{-0.3cm}
\centering
\resizebox{\linewidth}{!}{%
\setlength{\tabcolsep}{3.0pt}
\renewcommand{\arraystretch}{1.2}
\begin{tabular}{l|rrrrr|rrrrr|rrrrr|rrrrr}
& \multicolumn{5}{c|}{Vis-20} & \multicolumn{5}{c|}{Vis-50} & \multicolumn{5}{c|}{Vis-100} & \multicolumn{5}{c}{Vis-200} \\
dataset & \multicolumn{1}{c}{0} & \multicolumn{1}{c}{100} & \multicolumn{1}{c}{500} & \multicolumn{1}{c}{1000} & \multicolumn{1}{c|}{2000} & \multicolumn{1}{c}{0} & \multicolumn{1}{c}{100} & \multicolumn{1}{c}{500} & \multicolumn{1}{c}{1000} & \multicolumn{1}{c|}{2000} & \multicolumn{1}{c}{0} & \multicolumn{1}{c}{100} & \multicolumn{1}{c}{500} & \multicolumn{1}{c}{1000} & \multicolumn{1}{c|}{2000} & \multicolumn{1}{c}{0} & \multicolumn{1}{c}{100} & \multicolumn{1}{c}{500} & \multicolumn{1}{c}{1000} & \multicolumn{1}{c}{2000} \\\hline
Okutama & 90.6 & 31.4 & 26.1 & 29.1 & 27.2 & 40.3 & 31.1 & 21.9 & 19.7 & 22.0 & 36.3 & 24.3 & 20.4 & 20.0 & 18.4 & 27.4 & 25.8 & 19.4 & 16.0 & 15.5 \\
ICG & 63.0 & 106.0 & 25.8 & 29.8 & 28.8 & 32.8 & 53.2 & 23.0 & 21.4 & 23.1 & 32.3 & 24.4 & 23.9 & 24.7 & 22.4 & 29.3 & 26.7 & 22.2 & 20.0 & 20.3 \\
HERIDAL & 151.0 & 136.2 & 31.2 & 34.6 & 27.9 & 62.7 & 45.2 & 24.2 & 22.9 & 22.6 & 58.5 & 34.4 & 25.5 & 31.5 & 21.5 & 40.6 & 35.7 & 26.8 & 22.1 & 20.1 \\
SARD & 132.2 & 436.4 & 33.0 & 37.3 & 33.5 & 40.7 & 69.2 & 26.6 & 23.0 & 24.0 & 63.2 & 35.0 & 29.3 & 27.4 & 22.8 & 38.8 & 34.1 & 28.1 & 20.9 & 20.2 \\
\end{tabular}
}
\end{subtable}
\end{table*}

\subsection{A Study on the Impact of the Synthetic Data Pool} 

For the third study, we explored the inherent properties of the synthetic data pool that in turn influence the use of synthetic data.\smallskip

\noindent{\bf Accuracy comparison \wrt rendering parameters.} Our synthetic data pool, the Archangel-Synthetic dataset, was built to show various human appearances captured with a virtual UAV by controlling several rendering parameters in a simulation space (altitudes and radii of camera location, camera viewing angles, and human characters and poses). To examine the effect of each parameter on using synthetic data, we construct five subsets of the synthetic data pool, where each is built more sparsely for one parameter while fixing the values of other parameters. Each sub-pool includes the synthetic data with sparsely and regularly sampled altitudes (SAlt), radii (SRad), viewing angles (SAng), human characters (SCha), or human poses (SPos). In addition to the regularly sampled sub-pools, we also construct extremely biased sampled sub-pools along the three parameters: altitudes, radii, and viewing angles. For altitude and radii, subsets (BSAlt and BSRad) were constructed by collecting images assuming that the camera location is far away, and for viewing angle, a subset (BSAng) was constructed with only images with the front view of a person. These additional sub-pools can be used to investigate the impact of low-diversity synthetic pools when leveraging synthetic data in training.\footnote{Details on how to build subsets of the synthetic data pool are provided in the supplementary material.}

In Table~\ref{tab:comparison_synthetic_pool}, we compare the detection accuracy of the original pool and its five subsets. `SPos' exhibits significantly lower accuracy than the original, while the other four subsets show similar or even higher accuracy. In sampling the synthetic pool, reducing the variety of human poses significantly decreases detection accuracy as it leads to the inability to cover a wide range of human poses in test data. However, the decrease in accuracy is not observed when using subsets of synthetic data linked to the sparse sampling of other parameters.\smallskip

\noindent{\bf Properties of the synthetic pool.} We introduce several metrics to understand the variation in the ability of the synthetic pool to cover a variety of human appearances, depending on the rendering parameters, when used sparsely. We firstly consider \emph{how densely data is located in the feature space} (density) and \emph{how diverse the data distribution is in the feature space} (diversity). Specifically, the density and diversity of the pool $\mathcal{P}$ can be defined as below:

\begin{itemize}
\item density:
\begin{equation}
\frac{1}{|\text{adj}(\mathcal{P})|}\!\!\!\!\!\!\!\!\!\sum_{~~~~~~{\bf p},{\bf q}\in\text{adj}(\mathcal{P})}{\!\!\!\!\!\!f({\bf p})^\top f({\bf q})}.
\label{eq:density}
\end{equation}

\item diversity:
\begin{equation}
\frac{1}{|\mathcal{P}|}\sum_{{\bf x}\in\mathcal{P}}{||f({\bf x})-\mu||_2^k}.
\label{eq:diversity}
\end{equation}
\end{itemize}

\noindent $f(\cdot)$ is the embedding in the feature space of the detector. Here, we use a detector trained without the synthetic data to avoid the influence of them used in training when measuring the properties of the pool. $\text{adj}(\mathcal{P})$ includes all data pairs associated with different neighboring values from each rendering parameter while others are fixed. Intuitively, a high $f({\bf p})^\top f({\bf q})$ in eq.~\ref{eq:density} indicates that ${\bf p}$ and ${\bf q}$ lie close to each other in the feature space. $\mu$ is the mean feature over all data points in $\mathcal{P}$ (\ie, $\mu=\sum_{{\bf x}\in \mathcal{P}}{f({\bf x})}$). $k$ is a hyper-parameter that controls how each data point deviates from $\mu$ when calculating diversity. Higher $k$ leads to more weights on the data points away from $\mu$. (We use $k=10$). We also consider the domain gap between the synthetic pool and the reference dataset, which can be calculated in a similar way to measuring the distribution gap (eq.~\ref{eq:distribution gap}).

In Table~\ref{tab:comparison_synthetic_pool}, three sub-pools (`SPos', `BSAlt', and `BSRad'), which showed significantly lower accuracy than the original pool, have the following properties: higher density, less diversity, and closer domain gap to the reference dataset than the original and other pools. This strongly indicates that \emph{sampling synthetic data from a denser but less diverse distribution adversely affects using synthetic data in training, leading to low accuracy}. Moreover, `BSAlt', which presents very low accuracy, has a very large domain gap to the reference data. Using synthetic pools with large domain gaps to the reference data adversely impacts accuracy in both same-domain and cross-domain tasks. Unexpectedly, when `BSAng' was used, detection accuracy did not decrease, and the properties of the synthetic pool were normal. This is probably because the reference (real) data and other test real datasets also contain images of people looking straight ahead.

\smallskip

\noindent{\bf Analysis in terms of distribution gaps.} In Table~\ref{tab:comparison_synthetic_pool}, using `SPos', `BSAlt', or `BSAng' results in a larger distribution gap for cross-domain datasets than using other pools. This is aligned well with our previous analyses.

\begin{table*}[t]
\caption{{\bf Comparison of various synthetic pools in terms of various aspects.} This comparison is performed with the Vis-20 setting.}
\label{tab:comparison_synthetic_pool}
\vspace{-0.3cm}
\centering
\resizebox{\linewidth}{!}{%
\setlength{\tabcolsep}{3.0pt}
\renewcommand{\arraystretch}{1.2}
\begin{tabular}{lc|ccccc||ccc|cccc}
& & \multicolumn{5}{c||}{accuracy} & \multicolumn{3}{c|}{property} & \multicolumn{4}{c}{distribution gap}\\
pool & \# img & VisDrone & Okutama & ICG & HERIDAL & SARD & density & diversity & domain gap & Okutama & ICG & HERIDAL & SARD \\\hline
\textcolor{gray}{original} & \textcolor{gray}{17,280} & \textcolor{gray}{1.82} & \textcolor{gray}{7.18} & \textcolor{gray}{8.09} & ~~\textcolor{gray}{9.57} & \textcolor{gray}{11.53} & \textcolor{gray}{473.1} & \textcolor{gray}{1.7e+15} & \textcolor{gray}{135.8} & \textcolor{gray}{35.1} & \textcolor{gray}{39.4} & \textcolor{gray}{40.8} & \textcolor{gray}{36.8}\\
SAlt & ~~8,640 & 1.76 & 6.15 & 7.48 & ~~9.79 & 13.67 & 482.8 & 1.1e+15 & 149.3 & 32.9 & 37.9 & 38.9 & 35.0\\
SRad & ~~8,640 & 1.84 & 7.56 & 6.58 & ~~9.21 & 12.97 & 481.5 & 1.6e+15 & 137.6 & 31.9 & 36.3 & 35.1 & 32.3\\
SAng & ~~8,640 & 1.96 & 6.95 & 9.36 & ~~9.55 & 13.67 & 468.5 & 1.7e+15 & 141.3 & 33.0 & 37.4 & 36.2 & 35.7\\
SCha & ~~8,640 & 1.84 & 7.89 & 6.85 & 10.02 & 13.11 & 481.9 & 1.5e+15 & 142.0 & 30.2 & 37.3 & 36.4 & 33.1\\
SPos & ~~5,760 & {\bf 1.88} & {\bf 6.62} & {\bf 2.13} & ~~{\bf 3.41} & ~~{\bf 6.45} & {\bf 542.9} & {\bf 5.7e+14} & {\bf 124.0} & {\bf 35.5} & {\bf 40.3} & {\bf 51.1} & {\bf 49.1}\\\hline
BSAlt & ~~8,640 & {\bf 1.51} & {\bf 4.92} & {\bf 1.29} & ~~{\bf 1.72} & ~~{\bf 3.26} & {\bf 548.8} & {\bf 8.7e+12} & {\bf 265.7} & {\bf 34.8} & {\bf 44.3} & {\bf 67.5} & {\bf 93.2} \\
BSRad & ~~5,760 & {\bf 1.71} & {\bf 6.41} & {\bf 1.51} & ~~{\bf 3.84} & ~~{\bf 5.79} & {\bf 527.9} & {\bf 2.7e+14} & {\bf 151.5} & {\bf 35.8} & {\bf 42.5} & {\bf 51.2} & {\bf 45.6} \\
BSAng & ~~6,912 & 1.90 & 7.48 & 8.92 & ~~9.17 & 11.83 & 474.2 & 1.6e+15 & 137.0 & 35.3 & 38.2 & 42.4 & 36.7 \\
\end{tabular}
}
\end{table*}
\section{Discussions}
\label{sec:discussion}

Through our comprehensive analysis based on the extensive experiments, we have brought to light valuable findings that have not been previously identified or have been used without an accurate understanding. Our findings are described as follows:\smallskip

\noindent{\bf 1) General mechanism for acquiring domain generalization ability.} Our experiments show that in cross-domain tasks, synthetic data has a major impact on reducing the distribution gaps from reference datasets for most data, resulting in remarkable increases in cross-domain accuracy. On the other hand, a considerable number of outlier data points unexpectedly had very large distribution gaps. Outliers may arise due to i) insufficient diversity of reference data that serves as a standard for collecting synthetic data and training the sim2real transformer, or ii) the inherent limitation of the synthetic data pool, which does not fully represent the entire cross-domain data. We further discuss the behavior of these factors (\ie reference (real) data, sim2real transformation, synthetic data pool) affecting the acquisition of domain generalization ability in the following findings.\smallskip

\noindent{\bf 2) Relationship in scalability between synthetic data and real data.} Our experiments show that the more real images are used, the more positive is the impact of synthetic data on detection performance, not only in same-domain tasks but also in cross-domain tasks. Our experiments also indicate that as the amount of synthetic data used for training gradually increases, the accuracy continuously improves and then plateaus at some point. Impressively, the maximum number of synthetic images that can be used without accuracy plateauing increases as more real images are used. Therefore, to maximize the impact of synthetic data in training, it is important to increase the amount of not only synthetic data but also real data. Our findings on scalability may have some connections to the previous works~\cite{SRichterECCV2016,GRosCVPR2016,HLeeJSTARS2021} searching for an optimal ratio between real and synthetic data in a training batch.\smallskip

\noindent{\bf 3) Sim2real transformation quality \vs.~distribution gap.} Which is more important: improving the sim2real transformation quality or reducing the distribution gap between datasets to acquire domain generalization ability? The answer to this question is that improving the sim2real transformation quality is more important. In our experiments comparing PTL and random selection, PTL designed to prevent the sim2real transformation quality degradation was less effective than random selection in reducing the distribution gaps. However, PTL consistently yields better accuracy than random selection in most experimental settings.\smallskip

\noindent{\bf 4) Effect of the synthetic data pool.} In our experiments, we analyzed how the properties of the synthetic data pool were related to the effectiveness of using synthetic data in training. Examining different properties of different synthetic data pools, we found that the density and diversity of pools are correlated with cross-domain detection accuracy. Therefore, we can select the optimal synthetic data pool to maximize the benefit of synthetic data by investigating the properties of the pool in advance.\smallskip

In closing, we expect that our illumination of these findings can help to break the current trend of either naively using or being hesitant to use synthetic data in computer vision applications, leading to more appropriate use of synthetic data in future research and practice.\smallskip

\noindent{\bf Broader impacts.} The comprehensive study in this paper was conducted on tasks that place a high emphasis on synthetic data, \ie, aerial-view human detection, few-shot learning, and cross-domain detection. While we can clearly see how various factors affecting the use of synthetic data in training considerably influence the results in these tasks, one can doubt if the findings here are generally applicable to other tasks.

Numerical results regarding detection accuracy and distribution gaps in our work are based on the particular large-scale synthetic dataset and various real-world datasets publically available in the aerial-based perception domain. Therefore, when applying the findings discovered in this study to a task from relatively different applications where data attributes or characteristics are considerably different, an initial investigation into data and domain gap distributions needs to be considered to maximize the benefits of leveraging the findings in this study.\smallskip

\medskip

\noindent{\bf Acknowledgements.} This research was sponsored by the Defense Threat Reduction Agency (DTRA) and the DEVCOM Army Research Laboratory (ARL) under Grant No. W911NF2120076. This research was also sponsored in part by the Army Research Office and Army Research Laboratory (ARL) under Grant Number W911NF-21-1-0258. The views and conclusions contained in this document are those of the authors and should not be interpreted as representing the official policies, either expressed or implied, of the Army Research Office, Army Research Laboratory (ARL) or the U.S. Government. The U.S. Government is authorized to reproduce and distribute reprints for Government purposes notwithstanding any copyright notation herein.

\appendix

\section{Preliminaries}
\label{sec:preliminary}

\subsection{Modeling Representation Space of Sigmoid-based Detector} 

In this section, we describe modeling the representation space of a sigmoid-based object detector by fitting a multivariate Gaussian distribution. We denote the random variable of the input and its label of a linear classifier as ${\bf x} \in \mathcal{X}$ and $y = \{y_c\}_{c=1,\cdots,C} \in \mathcal{Y}, y_c=\{0, 1\}$, respectively. Then, the posterior distribution defined by the linear classifier whose output is formed by the sigmoid function can be expressed as follows: 
\begin{equation}
    P(y_c=1|{\bf x}) = \frac{1}{1+\exp{\left(\minus w_c{\bf x}\minus b_c\right)}} = \frac{\exp{\left(w_c{\bf x}+b_c\right)}}{\exp{\left(w_c{\bf x}+b_c\right)}+1},\label{eq:post_sigmoid}
\end{equation}
where $w_c$ and $b_c$ are the weights and bias of the linear classifier for a category $c$, respectively.

Gaussian Discriminant Analysis (GDA) models the posterior distribution of the classifier by assuming that the class conditional distribution ($P({\bf x}|y)$) and the class prior distribution ($P(y)$) follow the multivariate Gaussian and the Bernoulli distributions, respectively, as follows:
\begin{eqnarray}
P({\bf x}|y_c=0) = \mathcal{N}(\mu_0, \Sigma_0), ~~~~~~~~ P({\bf x}|y_c=1) = \mathcal{N}(\mu_1, \Sigma_1),~~~\nonumber\\
P(y_c=0) = \beta_0/\left(\beta_0+\beta_1\right), ~~~~~~~ P(y_c=1) = \beta_1/\left(\beta_0+\beta_1\right),~
\end{eqnarray}

where $\mu_{\{0, 1\}}$ and $\Sigma_{\{0, 1\}}$ are the mean and covariance of the multivariate Gaussian distribution, and $\beta_{\{0, 1\}}$ is the unnormalized prior for the category $c$ and the background.

For the special case of GDA where all categories share the same covariance matrix ({\it i.e.}, $\Sigma_0 = \Sigma_1 = \Sigma_c$), known as Linear Discriminant Analysis (LDA), the posterior distribution ($P(y_c|{\bf x})$) can be expressed with $P({\bf x}|y_c)$ and $P(y_c)$ as follows:
\begin{eqnarray}
P(y_c=1|{\bf x})&& = \frac{P(y_c=1)P({\bf x}|y_c=1)}{P(y_c=0)P({\bf x}|y_c=0)+P(y_c=1)P({\bf x}|y_c=1)}\nonumber\\
&& = \frac{
\begin{aligned}\exp\Bigl(\left(\mu_1\minus\mu_0\right)^\top\Sigma_c^{\minus 1}{\bf x}\ldots~~~~~~~~~~~~~~~~~~~~~~~~~~~~~~~~~~~\\
-\frac{1}{2}\mu_1^\top\Sigma_c^{\minus 1}\mu_1+\frac{1}{2}\mu_0^\top\Sigma_c^{\minus 1}\mu_0+\log{\beta_1/\beta_0} \Bigr)\end{aligned}
}{
\begin{aligned}\exp\Bigl(\left(\mu_1\minus\mu_0\right)^\top\Sigma_c^{\minus 1}{\bf x}\ldots~~~~~~~~~~~~~~~~~~~~~~~~~~~~~~~~~~~\\
-\frac{1}{2}\mu_1^\top\Sigma_c^{\minus 1}\mu_1+\frac{1}{2}\mu_0^\top\Sigma_c^{\minus 1}\mu_0+\log{\beta_1/\beta_0} \Bigr)+1\end{aligned}
}.
\label{eq:post_gda}
\end{eqnarray}
Note that the quadratic term is canceled out since the shared covariance matrix is used. The posterior distribution derived by GDA in eq.~\ref{eq:post_gda} then becomes equivalent to the posterior distribution of the linear classifier with the sigmoid function in eq.~\ref{eq:post_sigmoid} when $w_c = \left(\mu_1\minus\mu_0\right)^\top\Sigma_c^{\minus 1}$ and $b_c = -\frac{1}{2}\mu_1^\top\Sigma_c^{\minus 1}\mu_1+\frac{1}{2}\mu_0^\top\Sigma_c^{\minus 1}\mu_0+\ln{\beta_1/\beta_0}$. This implies that the representation space formed by {\bf x} can be modeled by a multivariate Gaussian distribution. 

Based on the above derivation, if ${\bf x}$ is the output of the penultimate layer of an object detector for a region proposal, and a linear classifier defined by $w_c$ and $b_c$ is the last layer of the object detector, it can be said that the representation space of the object detector for a category $c$ can be modeled with a multivariate Gaussian distribution. In other words, the representation space for a category $c$ can be represented by two parameters $\mu_1$ ({\it i.e.}, $\mu_c$) and $\Sigma_c$ of the multivariate Gaussian distribution.\smallskip

\noindent{\bf Discussion.} The sigmoid function can be viewed as a special case of the softmax function defined for a single category as both functions take the form of an exponential term for the category-of-interest normalized by the sum of exponential terms for all considered categories. Therefore, it is straightforward to derive the modeling for the sigmoid-based detector from the previous work~\cite{KLeeNeurIPS2018}, who shows that the softmax-based classifier can be modeled with a multivariate Gaussian distribution in the representation space. However, our derivation is still meaningful in that it extends the applicability of an existing modeling limited to a certain type of classifier ({\it i.e.}, based on softmax) to general object detectors ({\it i.e.}, based on sigmoid). Most object detectors, especially one-stage detectors, generally use the sigmoid function, which does not consider other categories when calculating the model output for a certain category, since more than one category can be active on a single output.

\subsection{Cross-entropy with Mixture of Delta Distributions and Multivariate Gaussian Distribution} 

In this section, we derive the cross-entropy of two distributions that are modeled by a mixture of delta distributions and a multivariate Gaussian distribution as the normalized sum of Mahalanobis distances. Assume that the data distributions $P$ and $Q$ in two sets $\mathcal{D}_P$ and $\mathcal{D}_Q$ can be modeled by density functions ($p$ and $q$) that take the form of a mixture of delta distributions and a multivariate Gaussian distribution, respectively, as follows:
\begin{eqnarray}
p({\bf x}) &=& \frac{1}{|\mathcal{D}_P|}\sum_{{\bf x'}\in\mathcal{D}_P}{\delta({\bf x}-{\bf x}')},\\
q({\bf x}) &=& \frac{1}{\sqrt{2\pi}^k\det(\Sigma)^{1/2}}\exp\left(-\frac{1}{2}({\bf x}-\mu)^\top\Sigma^{-1}({\bf x}-\mu)\right),
\end{eqnarray}
where $\delta({\bf x})$ is a Dirac delta function whose value is zero everywhere except at ${\bf x}={\bf 0}$ and whose integral over $\mathcal{X}$, which is the entire space of {\bf x}, is one. $\mu$ and $\Sigma$ are two parameters of the multivariate Gaussian distribution, which can be calculated empirically over all ${\bf x}\in\mathcal{D}_Q$.

Then, the cross entropy, which statistically measures the difference from $Q$ to $P$ where $Q$ is treated as the reference distribution, can be expressed as:
\begin{eqnarray}
\mathcal{H}(P,Q)&&= \minus\int_{\mathcal{X}}{p({\bf x})\ln{q({\bf x})}d{\bf x}}\nonumber\\
&&\begin{aligned}= \minus\int_{\mathcal{X}}{}\frac{1}{|\mathcal{D}_P|}\sum_{{\bf x}'\in\mathcal{D}_P}{\delta{({\bf x}-{\bf x}')}}\ln{}\Biggl(\frac{1}{\sqrt{2\pi}^k\det{(\Sigma)}^{\frac{1}{2}}}\ldots~~~~~~~\\
\exp{\left(\minus\frac{1}{2}({\bf x}\minus\mu)^\top\Sigma^{\minus 1}({\bf x}\minus\mu)\right)}\Biggr)d{\bf x}.\end{aligned}
\label{eq:cross_entropy}
\end{eqnarray}
Using two basic rules of i) $\int_{\mathcal{X}}{\left(\sum_nf_n(x)\right)dx} = \sum_n\left(\int_{\mathcal{X}}{f_n(x)dx}\right)$ if the summation is performed on a finite set, and ii) $\int_{\mathcal{X}}{\delta(x-a)f(x)dx}=f(a)$ if $f(x)$ is continuous on $\mathcal{X}$, the cross entropy in eq.~\ref{eq:cross_entropy} can be derived as:
\begin{eqnarray}
\mathcal{H}(P,Q)&=& -\frac{1}{|\mathcal{D}_p|}\sum_{{\bf x}\in\mathcal{D}_p}{\ln{\left(\frac{1}{\sqrt{2\pi}^k\det{(\Sigma)}^{1/2}}\exp{\left(-\frac{1}{2}({\bf x}-\mu)^\top\Sigma^{-1}({\bf x}-\mu)\right)}\right)}}\nonumber\\
&=& \frac{1}{2|\mathcal{D}_p|}\sum_{{\bf x}\in\mathcal{D}_p}{({\bf x}-\mu)^\top\Sigma^{-1}({\bf x}-\mu)}+\ln{(\sqrt{2\pi}^k\det{(\Sigma)}^{1/2})}.
\label{eq:sum_of_Mahalanobis}
\end{eqnarray}
Note that conditions for realizing the two basic rules are satisfied in our scenario, as i) the summation is computed on a finite set $\mathcal{D}_P$, and ii) a log of the multivariate Gaussian distribution is continuous on $\mathcal{X}$.

In our scenario where the cross-entropy is used to compare the distribution gaps of different test datasets (here, $\mathcal{D}_P$s) while the reference dataset ($\mathcal{D}_Q$) is fixed, and it is computed on the representation space of the detector, the cross-entropy can be expressed as:
\begin{equation}
    \mathcal{H}(P,Q)=\frac{1}{2|\mathcal{D}_P|}\sum_{{\bf x}\in\mathcal{D}_P}{\!\!(f({\bf x})\minus\mu)^\top\Sigma^{\minus 1}(f({\bf x})\minus\mu)} + C,
\end{equation}
where $f(\cdot)$ is the output of the detector in the representation space. The second term of eq.~\ref{eq:sum_of_Mahalanobis} can be regarded as a constant since $k$ (the dimension of the representation space) and $\Sigma$ (parameter of the reference dataset $\mathcal{D}_Q$) are not affected by the test dataset $\mathcal{D}_P$.\smallskip

\noindent{\bf Discussion (\vs FID).} Our metric has some similarities to FID (Fr{\` e}chet Inception Distance)~\cite{MHeuselNeruIPS2017}, a widely used metric in evaluating image quality generated by generative models, in that it analyzes a dataset by measuring the distance to a reference dataset. There are two notable differences: i) The two datasets being compared are represented as the multi-variate Gaussian distribution in the FID calculation, while only the reference dataset is represented as the multi-variate Gaussian distribution in our metric, ii) FID calculation is performed using features of the inception model~\cite{CSzegedyCVPR2015} trained on the ImageNet dataset~\cite{JDengCVPR2009}, while our metric uses features from the detector trained on the reference dataset. These differences occurred because FID determined the above characteristics for practical reasons, whereas our method derived them mathematically. In other words, there may be cases in which the evaluation with FID is not valid (\eg, the case where the dataset being compared has very different characteristics from the ImageNet dataset), but our metric is always valid if the prerequisite conditions are satisfied.
\section{Implementation Details}
\label{sec:implementation}

\noindent{\bf PTL.} We followed the original PTL paper~\cite{YShenCVPR2023} for all architectural details and training specifications of PTL except for the numbers of training epochs and iterations. The numbers of training epochs (used in sim2real transformer training) and training iterations (used in detector training) are modified to adopt a training time curtailment strategy.  Specifically, in the original PTL, the sim2real transformer and detector are trained for 100 epochs and 6.0k iterations, respectively, but when adopting the strategy, they are trained for 20 epochs and 1.2k iterations, respectively, after the 0th iteration.\smallskip

\noindent{\bf Random selection.} For random selection, we used the PTL implementation after modifying the synthetic data selection. In particular, while PTL is designed to select synthetic images by weighting images closer in domain gap to the training set, this selection is modified to randomly select synthetic data. All other parts except this selection process of the PTL training pipeline were used unchanged.\smallskip

\noindent{\bf Subsets of synthetic data pool} The Archangel-synthetic dataset~\cite{YShenAccess2023} was originally created by varying the five rendering parameters as follows: 10 altitudes (from 5$m$ to 50$m$ at 5$m$ interval), 6 radii (from 5$m$ to 30$m$ at 5$m$ interval), 12 angles (from 0$^\circ$ to 330$^\circ$ at 30$^\circ$ interval) 8 human characters (Juliet, Kelly, Lucy, Mary, Romeo, Scott, Troy, and Victor), and 3 human poses (stand, prone, squat). Each subset of the synthetic data pool is built using a sparse set of each rendering parameter, as follows:
\begin{itemize}
    \item SAlt: using sparser 5 altitudes from 10$m$ to 50$m$ at 10$m$ interval.
    \item SRad: using sparser 3 radii from 10$m$ to 30$m$ at 10$m$ interval.
    \item SAng: using sparser 6 angles from 0$^\circ$ to 300$^\circ$ at 60$^\circ$ interval.
    \item SCha: using sparser 4 human characters of Juliet, Kelly, Romeo, and Scott.
    \item SPos: using sparser 1 human pose of standing.
    \item BSAlt: using sparser 5 altitudes from 30$m$ to 50$m$ at 5$m$ interval.
    \item BSRad: using sparser 3 radii from 20$m$ to 30$m$ at 5$m$ interval.
    \item BSAng: using narrow and sparser 5 angles of 300$^\circ$, 330$^\circ$, 0$^\circ$, 30$^\circ$, and 60$^\circ$.
\end{itemize}
For each subset, all other parameters were the same as those of the original pool, except for the rendering parameter indicated to be used sparsely.

\section{Numerical Results}
\label{sec:numerical_result}

In this section, we present the numerical results of the graphs used for analysis in the main manuscript and additional results not presented in the main manuscript.

\begin{table}[t]
\caption{{\bf \textit{Wall-clock} Training time breakdown} for sim2real transformer training and detector training. Training time is shown in \textit{mins}. The numbers in the parentheses indicate training epochs and iterations for the corresponding PTL iteration for sim2real transformer training and detector training, respectively.}
\vspace{-0.3cm}
\label{tab:time_breakdown}
\begin{subtable}[b]{.46\linewidth}
\caption{sim2real transformer, Vis-20}
\vspace{-0.3cm}
\centering
\resizebox{\linewidth}{!}{%
\setlength{\tabcolsep}{2.0pt}
\renewcommand{\arraystretch}{1.2}
\begin{tabular}{c|rrrrr}
from-prev-iter & \multicolumn{1}{c}{0} & \multicolumn{1}{c}{1} & \multicolumn{1}{c}{2} & \multicolumn{1}{c}{3} & \multicolumn{1}{c}{4} \\\hline
& 28 (100) & ~~41 (100) & ~~56 (100) & ~~69 (100) & ~~83 (100) \\
\textcolor{red}{\checkmark} & 28 (100) & 8~~~(20) & 12~~~(20) & 14~~~(20) & 17~~~(20) \\
\end{tabular}
}
\end{subtable}
\hfill
\begin{subtable}[b]{.54\linewidth}
\caption{detector, Vis-20}
\vspace{-0.3cm}
\centering
\resizebox{\linewidth}{!}{%
\setlength{\tabcolsep}{2.0pt}
\renewcommand{\arraystretch}{1.2}
\begin{tabular}{c|cccccc}
from-prev-iter & 0 & 1 & 2 & 3 & 4 & 5 \\\hline
& 40 (6.0k) & 36 (6.0k) & 32 (6.0k) & 28 (6.0k) & 25 (6.0k) & 22 (6.0k) \\
\textcolor{red}{\checkmark} & 40 (6.0k) & 24 (1.2k) & 21 (1.2k) & 19 (1.2k) & 17 (1.2k) & 15 (1.2k) \\
\end{tabular}
}
\end{subtable}
\vspace{0.3cm}
\\
\begin{subtable}[b]{.46\linewidth}
\caption{sim2real transformer, Vis-50}
\vspace{-0.3cm}
\centering
\resizebox{\linewidth}{!}{%
\setlength{\tabcolsep}{2.0pt}
\renewcommand{\arraystretch}{1.2}
\begin{tabular}{c|rrrrr}
from-prev-iter & \multicolumn{1}{c}{0} & \multicolumn{1}{c}{1} & \multicolumn{1}{c}{2} & \multicolumn{1}{c}{3} & \multicolumn{1}{c}{4} \\\hline
& 87 (100) & 101 (100) & 114 (100) & 128 (100) & 142 (100) \\
\textcolor{red}{\checkmark} & 87 (100) & 20~~~(20) & 23~~~(20) & 26~~~(20) & 29~~~(20) \\
\end{tabular}
}
\end{subtable}
\hfill
\begin{subtable}[b]{.54\linewidth}
\caption{detector, Vis-50}
\vspace{-0.3cm}
\centering
\resizebox{\linewidth}{!}{%
\setlength{\tabcolsep}{2.0pt}
\renewcommand{\arraystretch}{1.2}
\begin{tabular}{c|cccccc}
from-prev-iter & 0 & 1 & 2 & 3 & 4 & 5 \\\hline
& 40 (6.0k) & 38 (6.0k) & 36 (6.0k) & 35 (6.0k) & 33 (6.0k) & 31 (6.0k) \\
\textcolor{red}{\checkmark} & 40 (6.0k) & 25 (1.2k) & 24 (1.2k) & 23 (1.2k) & 22 (1.2k) & 21 (1.2k) \\
\end{tabular}
}
\end{subtable}
\end{table}

\subsection{Curtailment of PTL Training Time}

The reduced training time and altered accuracy by adopting the \emph{tune-from-previous-iteration} strategy is reported in the main manuscript. Here, we also present the reduced time for two separate components of the PTL training pipeline that are affected by the strategy: detector training and sim2real transformer training (Table~\ref{tab:time_breakdown}). The corresponding training times (in \textit{mins}) with and without the strategy for every PTL iteration in Vis-20/50 settings are shown in the table. It is noteworthy that training time per PTL iteration is longer in Vis-50 than Vis-20 when using the same numbers of training iterations and epochs. This is because in our experimental setting, the real images are larger in size than the synthetic images (the image sizes of VisDrone images used as real data and the Archangel-Synthetic images used as synthetic data are 2000$\times$1500 and 512$\times$512, respectively) and thus require more computation time, and account for a larger portion of the training set in Vis-50.

\subsection{Scalability Behavior of Real Data}

In Table~\ref{tab:scalability_realimg}, we present numerical results used to generate Fig.~2 of the main manuscript. Specifically, the numbers in Table~\ref{tab:accuracy_scalability_of_realimg} correspond to Fig.~2(a) and (b) of the main manuscript while the numbers in Table~\ref{tab:ratio_scalability_of_realimg} are matched to Fig.~2(c). The results using AP@.5 follow a similar trend to those using AP@[.5:.95], which have been analyzed in the main manuscript.

\begin{table}[t]
\caption{{\bf Numerical results with the size of real dataset}. In each bin presenting accuracy, the mean and standard deviation of AP@.5 and AP@[.5:.95] calculated over 3 runs are reported.}
\label{tab:scalability_realimg}
\vspace{-0.3cm}
\begin{subtable}[b]{\linewidth}
\caption{Same-domain and cross-domain accuracy in the Vis-$N$}
\label{tab:accuracy_scalability_of_realimg}
\vspace{-0.3cm}
\centering
\resizebox{\linewidth}{!}{%
\setlength{\tabcolsep}{6.0pt}
\renewcommand{\arraystretch}{1.2}
\begin{tabular}{l|c|rrrrr}
\multicolumn{1}{c|}{setup} & w/ synth & \multicolumn{1}{c}{VisDrone} & \multicolumn{1}{c}{Okutama} & \multicolumn{1}{c}{ICG} & \multicolumn{1}{c}{HERIDAL} & \multicolumn{1}{c}{SARD} \\\hline
\multirow{2}{*}{Vis-20} & & 3.43\textcolor{gray}{{\scriptsize $\pm$0.57}} / 0.98\textcolor{gray}{{\scriptsize $\pm$0.12}} & 18.38\textcolor{gray}{{\scriptsize $\pm$8.74}} / ~~4.73\textcolor{gray}{{\scriptsize $\pm$2.61}} & 2.14\textcolor{gray}{{\scriptsize $\pm$0.70}} / ~~0.43\textcolor{gray}{{\scriptsize $\pm$0.15}} & 7.11\textcolor{gray}{{\scriptsize $\pm$3.45}} / ~~2.13\textcolor{gray}{{\scriptsize $\pm$1.08}} & 7.24\textcolor{gray}{{\scriptsize $\pm$3.32}} / ~~2.07\textcolor{gray}{{\scriptsize $\pm$1.14}} \\
& \textcolor{red}{\checkmark} & 6.18\textcolor{gray}{{\scriptsize $\pm$0.47}} / 1.82\textcolor{gray}{{\scriptsize $\pm$0.33}} & 29.93\textcolor{gray}{{\scriptsize $\pm$3.01}} / ~~7.18\textcolor{gray}{{\scriptsize $\pm$0.90}} & 29.30\textcolor{gray}{{\scriptsize $\pm$2.70}} / ~~8.09\textcolor{gray}{{\scriptsize $\pm$0.33}} & 28.61\textcolor{gray}{{\scriptsize $\pm$2.91}} / ~~9.57\textcolor{gray}{{\scriptsize $\pm$1.13}} & 34.96\textcolor{gray}{{\scriptsize $\pm$3.26}} / 11.53\textcolor{gray}{{\scriptsize $\pm$1.06}}\\\hline
\multirow{2}{*}{Vis-50} & & 6.13\textcolor{gray}{{\scriptsize $\pm$0.28}} / 1.84\textcolor{gray}{{\scriptsize $\pm$0.21}} & 25.65\textcolor{gray}{{\scriptsize $\pm$4.62}} / ~~6.98\textcolor{gray}{{\scriptsize $\pm$1.56}} & 6.57\textcolor{gray}{{\scriptsize $\pm$2.41}} / ~~1.48\textcolor{gray}{{\scriptsize $\pm$0.89}} & 12.12\textcolor{gray}{{\scriptsize $\pm$3.35}} / ~~3.93\textcolor{gray}{{\scriptsize $\pm$1.17}} & 17.73\textcolor{gray}{{\scriptsize $\pm$1.58}} / ~~4.92\textcolor{gray}{{\scriptsize $\pm$0.61}} \\
& \textcolor{red}{\checkmark} & 8.91\textcolor{gray}{{\scriptsize $\pm$0.20}} / 2.79\textcolor{gray}{{\scriptsize $\pm$0.08}} & 37.67\textcolor{gray}{{\scriptsize $\pm$0.59}} / ~~9.80\textcolor{gray}{{\scriptsize $\pm$0.25}} & 32.86\textcolor{gray}{{\scriptsize $\pm$5.36}} / ~~9.64\textcolor{gray}{{\scriptsize $\pm$2.03}} & 36.88\textcolor{gray}{{\scriptsize $\pm$3.79}} / 12.16\textcolor{gray}{{\scriptsize $\pm$1.93}} & 45.75\textcolor{gray}{{\scriptsize $\pm$2.16}} / 15.92\textcolor{gray}{{\scriptsize $\pm$1.73}}\\\hline
\multirow{2}{*}{Vis-100} & & 7.91\textcolor{gray}{{\scriptsize $\pm$0.13}} / 2.36\textcolor{gray}{{\scriptsize $\pm$0.07}} & 31.37\textcolor{gray}{{\scriptsize $\pm$1.49}} / ~~8.21\textcolor{gray}{{\scriptsize $\pm$0.29}} & 7.60\textcolor{gray}{{\scriptsize $\pm$1.51}} / ~~1.81\textcolor{gray}{{\scriptsize $\pm$0.32}} & 14.64\textcolor{gray}{{\scriptsize $\pm$6.04}} / ~~4.59\textcolor{gray}{{\scriptsize $\pm$1.47}} & 18.27\textcolor{gray}{{\scriptsize $\pm$1.25}} / ~~5.42\textcolor{gray}{{\scriptsize $\pm$0.50}} \\
& \textcolor{red}{\checkmark} & 10.56\textcolor{gray}{{\scriptsize $\pm$0.49}} / 3.29\textcolor{gray}{{\scriptsize $\pm$0.23}} & 41.18\textcolor{gray}{{\scriptsize $\pm$3.35}} / 10.86\textcolor{gray}{{\scriptsize $\pm$1.07}} & 35.69\textcolor{gray}{{\scriptsize $\pm$1.65}} / 11.02\textcolor{gray}{{\scriptsize $\pm$1.35}} & 38.54\textcolor{gray}{{\scriptsize $\pm$7.75}} / 13.37\textcolor{gray}{{\scriptsize $\pm$2.36}} & 48.15\textcolor{gray}{{\scriptsize $\pm$1.18}} / 17.05\textcolor{gray}{{\scriptsize $\pm$0.87}}\\\hline
\multirow{2}{*}{Vis-200} & & 10.55\textcolor{gray}{{\scriptsize $\pm$1.41}} / 3.18\textcolor{gray}{{\scriptsize $\pm$0.55}} & 38.58\textcolor{gray}{{\scriptsize $\pm$4.81}} / 10.25\textcolor{gray}{{\scriptsize $\pm$1.66}} & 6.50\textcolor{gray}{{\scriptsize $\pm$3.17}} / ~~1.39\textcolor{gray}{{\scriptsize $\pm$0.63}} & 14.76\textcolor{gray}{{\scriptsize $\pm$9.76}} / ~~4.68\textcolor{gray}{{\scriptsize $\pm$2.72}} & 21.87\textcolor{gray}{{\scriptsize $\pm$7.48}} / ~~6.98\textcolor{gray}{{\scriptsize $\pm$1.97}} \\
& \textcolor{red}{\checkmark} & 12.78\textcolor{gray}{{\scriptsize $\pm$0.48}} / 4.10\textcolor{gray}{{\scriptsize $\pm$0.28}} & 46.62\textcolor{gray}{{\scriptsize $\pm$0.93}} / 12.37\textcolor{gray}{{\scriptsize $\pm$0.21}} & 30.48\textcolor{gray}{{\scriptsize $\pm$0.34}} / ~~8.21\textcolor{gray}{{\scriptsize $\pm$0.43}} & 37.60\textcolor{gray}{{\scriptsize $\pm$2.12}} / 13.74\textcolor{gray}{{\scriptsize $\pm$1.46}} & 49.60\textcolor{gray}{{\scriptsize $\pm$1.78}} / 17.73\textcolor{gray}{{\scriptsize $\pm$0.49}}\\
\end{tabular}
}
\end{subtable}
\vspace{0.3cm}
\\
\begin{subtable}[b]{\linewidth}
\caption{Same-domain accuracy w/o synthetic data and its ratio to the Vis-$N$ (w/ synthetic data) when using the same number of real images}
\label{tab:ratio_scalability_of_realimg}
\vspace{-0.3cm}
\centering
\resizebox{\linewidth}{!}{%
\setlength{\tabcolsep}{8.0pt}
\renewcommand{\arraystretch}{1.2}
\begin{tabular}{l|c|rrrr}
& & \multicolumn{4}{c}{\# of real image} \\
& testset & \multicolumn{1}{c}{20} & \multicolumn{1}{c}{50} & \multicolumn{1}{c}{100} & \multicolumn{1}{c}{200} \\\hline
accuracy & \multirow{2}{*}{Okutama} & 37.93\textcolor{gray}{{\scriptsize $\pm$1.75}} / ~~9.93\textcolor{gray}{{\scriptsize $\pm$0.16}} & 51.31\textcolor{gray}{{\scriptsize $\pm$2.05}} / 14.13\textcolor{gray}{{\scriptsize $\pm$0.90}} & 55.98\textcolor{gray}{{\scriptsize $\pm$0.75}} / 16.68\textcolor{gray}{{\scriptsize $\pm$0.45}} & 64.76\textcolor{gray}{{\scriptsize $\pm$0.97}} / 20.28\textcolor{gray}{{\scriptsize $\pm$0.17}}\\
ratio to Vis-$N$ (w/ synth) & & \multicolumn{1}{c}{0.79 / 0.72} & \multicolumn{1}{c}{0.73 / 0.69} & \multicolumn{1}{c}{0.74 / 0.65} & \multicolumn{1}{c}{0.72 / 0.61} \\\hline
accuracy & \multirow{2}{*}{ICG} & 40.36\textcolor{gray}{{\scriptsize $\pm$0.96}} / 10.63\textcolor{gray}{{\scriptsize $\pm$0.38}} & 60.95\textcolor{gray}{{\scriptsize $\pm$1.76}} / 19.57\textcolor{gray}{{\scriptsize $\pm$1.19}} & 73.23\textcolor{gray}{{\scriptsize $\pm$0.76}} / 27.53\textcolor{gray}{{\scriptsize $\pm$0.95}} & 84.21\textcolor{gray}{{\scriptsize $\pm$1.50}} / 35.98\textcolor{gray}{{\scriptsize $\pm$1.20}}\\
ratio to Vis-$N$ (w/ synth) & & \multicolumn{1}{c}{0.73 / 0.76} & \multicolumn{1}{c}{0.54 / 0.49} & \multicolumn{1}{c}{0.49 / 0.40} & \multicolumn{1}{c}{0.36 / 0.23} \\\hline
accuracy & \multirow{2}{*}{HERIDAL} & 41.39\textcolor{gray}{{\scriptsize $\pm$2.86}} / 12.75\textcolor{gray}{{\scriptsize $\pm$1.82}} & 58.97\textcolor{gray}{{\scriptsize $\pm$2.86}} / 19.76\textcolor{gray}{{\scriptsize $\pm$0.96}} & 65.78\textcolor{gray}{{\scriptsize $\pm$0.70}} / 26.27\textcolor{gray}{{\scriptsize $\pm$1.25}} & 71.53\textcolor{gray}{{\scriptsize $\pm$0.49}} / 31.18\textcolor{gray}{{\scriptsize $\pm$1.70}}\\
ratio to Vis-$N$ (w/ synth) & & \multicolumn{1}{c}{0.69 / 0.75} & \multicolumn{1}{c}{0.63 / 0.62} & \multicolumn{1}{c}{0.59 / 0.51} & \multicolumn{1}{c}{0.53 / 0.44} \\\hline
accuracy & \multirow{2}{*}{SARD} & 33.44\textcolor{gray}{{\scriptsize $\pm$6.36}} / ~~8.52\textcolor{gray}{{\scriptsize $\pm$2.20}} & 51.35\textcolor{gray}{{\scriptsize $\pm$1.68}} / 15.28\textcolor{gray}{{\scriptsize $\pm$0.78}} & 66.81\textcolor{gray}{{\scriptsize $\pm$3.15}} / 23.75\textcolor{gray}{{\scriptsize $\pm$1.79}} & 75.76\textcolor{gray}{{\scriptsize $\pm$1.62}} / 30.17\textcolor{gray}{{\scriptsize $\pm$0.98}}\\
ratio to Vis-$N$ (w/ synth) & & \multicolumn{1}{c}{1.05 / 1.35} & \multicolumn{1}{c}{0.89 / 1.04} & \multicolumn{1}{c}{0.72 / 0.72} & \multicolumn{1}{c}{0.65 / 0.59} \\
\end{tabular}
}
\end{subtable}

\end{table}

\subsection{Scalability Behavior of Synthetic Data}

In Table~\ref{tab:scalability_synthimg}, we present numerical results used to generate Fig.~4 of the main manuscript. The results using AP@.5 follows a similar trend to those using AP@[.5:.95], which have been analyzed in the main manuscript.

\begin{table}[t]
\caption{{\bf Numerical results with the size of synthetic dataset.} `Random' denotes random selection.}
\label{tab:scalability_synthimg}
\vspace{-0.3cm}
\begin{subtable}[b]{\linewidth}
\caption{Vis-20}
\vspace{-0.3cm}
\centering
\resizebox{\linewidth}{!}{%
\setlength{\tabcolsep}{5.0pt}
\renewcommand{\arraystretch}{1.2}
\begin{tabular}{l|c|rrrrr}
& & \multicolumn{5}{c}{\# of synthetic image} \\
\multicolumn{1}{c|}{method} & test set & \multicolumn{1}{c}{0} & \multicolumn{1}{c}{100} & \multicolumn{1}{c}{500} & \multicolumn{1}{c}{1000} & \multicolumn{1}{c}{2000} \\\hline
PTL & \multirow{2}{*}{VisDrone} & \multirow{2}{*}{3.43\textcolor{gray}{{\scriptsize $\pm$0.57}} /~~0.98\textcolor{gray}{{\scriptsize $\pm$0.12}}} & 5.88\textcolor{gray}{{\scriptsize $\pm$0.58}} /~~1.73\textcolor{gray}{{\scriptsize $\pm$0.21}} & 6.48\textcolor{gray}{{\scriptsize $\pm$0.56}} /~~1.89\textcolor{gray}{{\scriptsize $\pm$0.13}} & 6.28\textcolor{gray}{{\scriptsize $\pm$0.86}} /~~1.76\textcolor{gray}{{\scriptsize $\pm$0.28}} & 6.18\textcolor{gray}{{\scriptsize $\pm$0.47}} /~~1.82\textcolor{gray}{{\scriptsize $\pm$0.33}} \\
Random & & & 5.90\textcolor{gray}{{\scriptsize $\pm$0.29}} /~~1.63\textcolor{gray}{{\scriptsize $\pm$0.07}} & 6.40\textcolor{gray}{{\scriptsize $\pm$0.85}} /~~1.79\textcolor{gray}{{\scriptsize $\pm$0.28}} & 6.41\textcolor{gray}{{\scriptsize $\pm$0.20}} /~~1.85\textcolor{gray}{{\scriptsize $\pm$0.09}} & 6.09\textcolor{gray}{{\scriptsize $\pm$0.69}} /~~1.75\textcolor{gray}{{\scriptsize $\pm$0.28}} \\\hline
PTL & \multirow{2}{*}{Okutama} & \multirow{2}{*}{18.38\textcolor{gray}{{\scriptsize $\pm$8.74}} /~~4.73\textcolor{gray}{{\scriptsize $\pm$2.61}}} & 28.01\textcolor{gray}{{\scriptsize $\pm$2.89}} /~~6.73\textcolor{gray}{{\scriptsize $\pm$0.80}} & 30.54\textcolor{gray}{{\scriptsize $\pm$1.06}} /~~7.24\textcolor{gray}{{\scriptsize $\pm$0.36}} & 29.63\textcolor{gray}{{\scriptsize $\pm$1.21}} /~~6.90\textcolor{gray}{{\scriptsize $\pm$0.41}} & 29.93\textcolor{gray}{{\scriptsize $\pm$3.01}} /~~7.18\textcolor{gray}{{\scriptsize $\pm$0.90}} \\
Random & & & 27.54\textcolor{gray}{{\scriptsize $\pm$1.40}} /~~6.18\textcolor{gray}{{\scriptsize $\pm$0.62}} & 28.20\textcolor{gray}{{\scriptsize $\pm$1.74}} /~~6.30\textcolor{gray}{{\scriptsize $\pm$0.34}} & 27.20\textcolor{gray}{{\scriptsize $\pm$0.90}} /~~6.08\textcolor{gray}{{\scriptsize $\pm$0.24}} & 27.80\textcolor{gray}{{\scriptsize $\pm$3.03}} /~~6.08\textcolor{gray}{{\scriptsize $\pm$0.99}} \\\hline
PTL & \multirow{2}{*}{ICG} & \multirow{2}{*}{2.14\textcolor{gray}{{\scriptsize $\pm$0.70}} /~~0.43\textcolor{gray}{{\scriptsize $\pm$0.15}}} & 15.16\textcolor{gray}{{\scriptsize $\pm$3.85}} /~~3.58\textcolor{gray}{{\scriptsize $\pm$1.24}} & 23.03\textcolor{gray}{{\scriptsize $\pm$4.68}} /~~6.13\textcolor{gray}{{\scriptsize $\pm$1.79}} & 26.70\textcolor{gray}{{\scriptsize $\pm$1.86}} /~~7.47\textcolor{gray}{{\scriptsize $\pm$1.37}} & 29.30\textcolor{gray}{{\scriptsize $\pm$2.70}} /~~8.09\textcolor{gray}{{\scriptsize $\pm$0.33}} \\
Random & & & 8.97\textcolor{gray}{{\scriptsize $\pm$0.67}} /~~1.97\textcolor{gray}{{\scriptsize $\pm$0.20}} & 29.62\textcolor{gray}{{\scriptsize $\pm$2.03}} /~~8.06\textcolor{gray}{{\scriptsize $\pm$0.30}} & 26.69\textcolor{gray}{{\scriptsize $\pm$6.21}} /~~7.39\textcolor{gray}{{\scriptsize $\pm$1.14}} & 33.70\textcolor{gray}{{\scriptsize $\pm$0.85}} /~~9.54\textcolor{gray}{{\scriptsize $\pm$0.87}} \\\hline
PTL & \multirow{2}{*}{HERIDAL} & \multirow{2}{*}{7.11\textcolor{gray}{{\scriptsize $\pm$3.45}} /~~2.13\textcolor{gray}{{\scriptsize $\pm$1.08}}} & 20.24\textcolor{gray}{{\scriptsize $\pm$3.85}} /~~5.87\textcolor{gray}{{\scriptsize $\pm$1.83}} & 26.50\textcolor{gray}{{\scriptsize $\pm$3.02}} /~~7.86\textcolor{gray}{{\scriptsize $\pm$1.10}} & 26.98\textcolor{gray}{{\scriptsize $\pm$5.38}} /~~8.49\textcolor{gray}{{\scriptsize $\pm$1.14}} & 28.61\textcolor{gray}{{\scriptsize $\pm$2.91}} /~~9.57\textcolor{gray}{{\scriptsize $\pm$1.13}} \\
Random & & & 14.82\textcolor{gray}{{\scriptsize $\pm$2.57}} /~~3.94\textcolor{gray}{{\scriptsize $\pm$1.00}} & 23.37\textcolor{gray}{{\scriptsize $\pm$2.19}} /~~6.77\textcolor{gray}{{\scriptsize $\pm$0.92}} & 25.98\textcolor{gray}{{\scriptsize $\pm$2.92}} /~~7.41\textcolor{gray}{{\scriptsize $\pm$0.73}} & 29.22\textcolor{gray}{{\scriptsize $\pm$4.55}} /~~9.02\textcolor{gray}{{\scriptsize $\pm$1.95}} \\\hline
PTL & \multirow{2}{*}{SARD} & \multirow{2}{*}{7.24\textcolor{gray}{{\scriptsize $\pm$3.32}} /~~2.07\textcolor{gray}{{\scriptsize $\pm$1.14}}} & 24.51\textcolor{gray}{{\scriptsize $\pm$3.39}} /~~7.37\textcolor{gray}{{\scriptsize $\pm$1.33}} & 34.74\textcolor{gray}{{\scriptsize $\pm$1.93}} /11.11\textcolor{gray}{{\scriptsize $\pm$1.14}} & 35.65\textcolor{gray}{{\scriptsize $\pm$3.07}} /11.90\textcolor{gray}{{\scriptsize $\pm$0.78}} & 34.96\textcolor{gray}{{\scriptsize $\pm$3.26}} /11.53\textcolor{gray}{{\scriptsize $\pm$1.06}} \\
Random & & & 22.15\textcolor{gray}{{\scriptsize $\pm$3.13}} /~~6.52\textcolor{gray}{{\scriptsize $\pm$1.44}} & 34.98\textcolor{gray}{{\scriptsize $\pm$5.04}} /11.18\textcolor{gray}{{\scriptsize $\pm$1.77}} & 37.35\textcolor{gray}{{\scriptsize $\pm$3.35}} /11.78\textcolor{gray}{{\scriptsize $\pm$0.67}} & 40.01\textcolor{gray}{{\scriptsize $\pm$2.07}} /13.36\textcolor{gray}{{\scriptsize $\pm$0.85}} \\
\end{tabular}
}
\end{subtable}
\vspace{0.3cm}
\\
\begin{subtable}[b]{\linewidth}
\caption{Vis-50}
\vspace{-0.3cm}
\centering
\resizebox{\linewidth}{!}{%
\setlength{\tabcolsep}{5.0pt}
\renewcommand{\arraystretch}{1.2}
\begin{tabular}{l|c|rrrrr}
& & \multicolumn{5}{c}{\# of synthetic image} \\
\multicolumn{1}{c|}{method} & test set & \multicolumn{1}{c}{0} & \multicolumn{1}{c}{100} & \multicolumn{1}{c}{500} & \multicolumn{1}{c}{1000} & \multicolumn{1}{c}{2000} \\\hline
PTL & \multirow{2}{*}{VisDrone} & \multirow{2}{*}{6.13\textcolor{gray}{{\scriptsize $\pm$0.28}} /~~1.84\textcolor{gray}{{\scriptsize $\pm$0.21}}} & 8.48\textcolor{gray}{{\scriptsize $\pm$0.26}} /~~2.55\textcolor{gray}{{\scriptsize $\pm$0.10}} & 9.27\textcolor{gray}{{\scriptsize $\pm$0.29}} /~~2.84\textcolor{gray}{{\scriptsize $\pm$0.12}} & 9.39\textcolor{gray}{{\scriptsize $\pm$0.12}} /~~2.98\textcolor{gray}{{\scriptsize $\pm$0.07}} & 8.91\textcolor{gray}{{\scriptsize $\pm$0.20}} /~~2.79\textcolor{gray}{{\scriptsize $\pm$0.08}} \\
Random & & & 8.17\textcolor{gray}{{\scriptsize $\pm$0.28}} /~~2.43\textcolor{gray}{{\scriptsize $\pm$0.22}} & 9.23\textcolor{gray}{{\scriptsize $\pm$0.29}} /~~2.77\textcolor{gray}{{\scriptsize $\pm$0.15}} & 8.97\textcolor{gray}{{\scriptsize $\pm$0.08}} /~~2.70\textcolor{gray}{{\scriptsize $\pm$0.11}} & 9.01\textcolor{gray}{{\scriptsize $\pm$0.56}} /~~2.69\textcolor{gray}{{\scriptsize $\pm$0.04}} \\\hline
PTL & \multirow{2}{*}{Okutama} & \multirow{2}{*}{25.65\textcolor{gray}{{\scriptsize $\pm$4.62}} /~~6.98\textcolor{gray}{{\scriptsize $\pm$1.56}}} & 35.21\textcolor{gray}{{\scriptsize $\pm$4.67}} /~~9.45\textcolor{gray}{{\scriptsize $\pm$1.70}} & 37.94\textcolor{gray}{{\scriptsize $\pm$1.84}} /~~9.88\textcolor{gray}{{\scriptsize $\pm$0.76}} & 37.17\textcolor{gray}{{\scriptsize $\pm$2.10}} /~~9.63\textcolor{gray}{{\scriptsize $\pm$0.95}} & 38.85\textcolor{gray}{{\scriptsize $\pm$3.34}} /10.04\textcolor{gray}{{\scriptsize $\pm$1.10}} \\
Random & & & 32.66\textcolor{gray}{{\scriptsize $\pm$5.86}} /~~8.46\textcolor{gray}{{\scriptsize $\pm$1.92}} & 34.48\textcolor{gray}{{\scriptsize $\pm$5.68}} /~~8.44\textcolor{gray}{{\scriptsize $\pm$2.03}} & 33.68\textcolor{gray}{{\scriptsize $\pm$6.44}} /~~8.12\textcolor{gray}{{\scriptsize $\pm$2.04}} & 33.29\textcolor{gray}{{\scriptsize $\pm$4.53}} /~~7.92\textcolor{gray}{{\scriptsize $\pm$1.20}} \\\hline
PTL & \multirow{2}{*}{ICG} & \multirow{2}{*}{6.57\textcolor{gray}{{\scriptsize $\pm$2.41}} /~~1.48\textcolor{gray}{{\scriptsize $\pm$0.89}}}& 16.87\textcolor{gray}{{\scriptsize $\pm$2.23}} /~~4.16\textcolor{gray}{{\scriptsize $\pm$0.71}} & 29.70\textcolor{gray}{{\scriptsize $\pm$3.13}} /~~7.27\textcolor{gray}{{\scriptsize $\pm$1.27}} & 30.94\textcolor{gray}{{\scriptsize $\pm$6.70}} /~~8.57\textcolor{gray}{{\scriptsize $\pm$2.54}} & 32.86\textcolor{gray}{{\scriptsize $\pm$5.36}} /~~9.64\textcolor{gray}{{\scriptsize $\pm$2.03}} \\
Random & & & 14.43\textcolor{gray}{{\scriptsize $\pm$2.72}} /~~3.24\textcolor{gray}{{\scriptsize $\pm$0.90}} & 28.78\textcolor{gray}{{\scriptsize $\pm$4.90}} /~~7.05\textcolor{gray}{{\scriptsize $\pm$1.65}} & 31.45\textcolor{gray}{{\scriptsize $\pm$1.49}} /~~8.72\textcolor{gray}{{\scriptsize $\pm$1.44}} & 35.51\textcolor{gray}{{\scriptsize $\pm$2.12}} /~~9.69\textcolor{gray}{{\scriptsize $\pm$0.33}} \\\hline
PTL & \multirow{2}{*}{HERIDAL} & \multirow{2}{*}{12.12\textcolor{gray}{{\scriptsize $\pm$3.35}} /~~3.93\textcolor{gray}{{\scriptsize $\pm$1.17}}} & 22.81\textcolor{gray}{{\scriptsize $\pm$1.43}} /~~7.22\textcolor{gray}{{\scriptsize $\pm$0.59}} & 31.62\textcolor{gray}{{\scriptsize $\pm$1.32}} /10.27\textcolor{gray}{{\scriptsize $\pm$0.74}} & 33.24\textcolor{gray}{{\scriptsize $\pm$1.66}} /11.31\textcolor{gray}{{\scriptsize $\pm$1.30}} & 36.88\textcolor{gray}{{\scriptsize $\pm$3.79}} /12.16\textcolor{gray}{{\scriptsize $\pm$1.93}} \\
Random & & & 21.71\textcolor{gray}{{\scriptsize $\pm$2.14}} /~~6.56\textcolor{gray}{{\scriptsize $\pm$0.72}} & 29.87\textcolor{gray}{{\scriptsize $\pm$3.93}} /~~9.73\textcolor{gray}{{\scriptsize $\pm$1.39}} & 32.11\textcolor{gray}{{\scriptsize $\pm$5.48}} /10.41\textcolor{gray}{{\scriptsize $\pm$2.96}} & 34.06\textcolor{gray}{{\scriptsize $\pm$5.59}} /11.08\textcolor{gray}{{\scriptsize $\pm$2.31}} \\\hline
PTL & \multirow{2}{*}{SARD} & \multirow{2}{*}{17.73\textcolor{gray}{{\scriptsize $\pm$1.58}} /~~4.92\textcolor{gray}{{\scriptsize $\pm$0.61}}} & 32.18\textcolor{gray}{{\scriptsize $\pm$3.75}} /~~9.77\textcolor{gray}{{\scriptsize $\pm$1.06}} & 43.67\textcolor{gray}{{\scriptsize $\pm$4.01}} /14.33\textcolor{gray}{{\scriptsize $\pm$1.66}} & 43.59\textcolor{gray}{{\scriptsize $\pm$2.31}} /14.15\textcolor{gray}{{\scriptsize $\pm$2.17}} & 45.75\textcolor{gray}{{\scriptsize $\pm$2.16}} /15.92\textcolor{gray}{{\scriptsize $\pm$1.73}} \\
Random & & & 30.74\textcolor{gray}{{\scriptsize $\pm$1.61}} /~~9.41\textcolor{gray}{{\scriptsize $\pm$0.65}} & 38.52\textcolor{gray}{{\scriptsize $\pm$0.88}} /12.10\textcolor{gray}{{\scriptsize $\pm$0.96}} & 44.28\textcolor{gray}{{\scriptsize $\pm$2.83}} /14.30\textcolor{gray}{{\scriptsize $\pm$1.68}} & 45.56\textcolor{gray}{{\scriptsize $\pm$3.26}} /15.21\textcolor{gray}{{\scriptsize $\pm$1.64}} \\
\end{tabular}
}
\end{subtable}
\vspace{0.3cm}
\\
\begin{subtable}[b]{\linewidth}
\caption{Vis-100}
\vspace{-0.3cm}
\centering
\resizebox{\linewidth}{!}{%
\setlength{\tabcolsep}{5.0pt}
\renewcommand{\arraystretch}{1.2}
\begin{tabular}{l|c|rrrrr}
& & \multicolumn{5}{c}{\# of synthetic image} \\
\multicolumn{1}{c|}{method} & test set & \multicolumn{1}{c}{0} & \multicolumn{1}{c}{100} & \multicolumn{1}{c}{500} & \multicolumn{1}{c}{1000} & \multicolumn{1}{c}{2000} \\\hline
PTL & \multirow{2}{*}{VisDrone} & \multirow{2}{*}{7.91\textcolor{gray}{{\scriptsize $\pm$0.13}} /~~2.36\textcolor{gray}{{\scriptsize $\pm$0.07}}} & 9.58\textcolor{gray}{{\scriptsize $\pm$0.57}} /~~2.94\textcolor{gray}{{\scriptsize $\pm$0.21}} & 10.79\textcolor{gray}{{\scriptsize $\pm$0.28}} /~~3.40\textcolor{gray}{{\scriptsize $\pm$0.09}} & 10.82\textcolor{gray}{{\scriptsize $\pm$0.45}} /~~3.41\textcolor{gray}{{\scriptsize $\pm$0.07}} & 10.56\textcolor{gray}{{\scriptsize $\pm$0.49}} /~~3.29\textcolor{gray}{{\scriptsize $\pm$0.23}} \\
Random & & & 9.13\textcolor{gray}{{\scriptsize $\pm$0.60}} /~~2.64\textcolor{gray}{{\scriptsize $\pm$0.16}} & 10.66\textcolor{gray}{{\scriptsize $\pm$0.10}} /~~3.23\textcolor{gray}{{\scriptsize $\pm$0.09}} & 10.67\textcolor{gray}{{\scriptsize $\pm$0.43}} /~~3.28\textcolor{gray}{{\scriptsize $\pm$0.14}} & 10.13\textcolor{gray}{{\scriptsize $\pm$0.27}} /~~3.11\textcolor{gray}{{\scriptsize $\pm$0.03}} \\\hline
PTL & \multirow{2}{*}{Okutama} & \multirow{2}{*}{31.37\textcolor{gray}{{\scriptsize $\pm$1.49}} /~~8.21\textcolor{gray}{{\scriptsize $\pm$0.29}}} & 36.61\textcolor{gray}{{\scriptsize $\pm$4.46}} /~~9.89\textcolor{gray}{{\scriptsize $\pm$1.28}} & 40.37\textcolor{gray}{{\scriptsize $\pm$4.12}} /10.41\textcolor{gray}{{\scriptsize $\pm$0.91}} & 40.76\textcolor{gray}{{\scriptsize $\pm$4.69}} /10.48\textcolor{gray}{{\scriptsize $\pm$1.12}} & 41.18\textcolor{gray}{{\scriptsize $\pm$3.35}} /10.86\textcolor{gray}{{\scriptsize $\pm$1.07}} \\
Random & & & 34.92\textcolor{gray}{{\scriptsize $\pm$2.86}} /~~9.05\textcolor{gray}{{\scriptsize $\pm$0.85}} & 38.12\textcolor{gray}{{\scriptsize $\pm$2.66}} /~~9.51\textcolor{gray}{{\scriptsize $\pm$0.42}} & 38.80\textcolor{gray}{{\scriptsize $\pm$2.42}} /~~9.77\textcolor{gray}{{\scriptsize $\pm$0.26}} & 38.18\textcolor{gray}{{\scriptsize $\pm$1.79}} /~~9.50\textcolor{gray}{{\scriptsize $\pm$0.40}} \\\hline
PTL & \multirow{2}{*}{ICG} & \multirow{2}{*}{7.60\textcolor{gray}{{\scriptsize $\pm$1.51}} /~~1.81\textcolor{gray}{{\scriptsize $\pm$0.32}}} & 18.19\textcolor{gray}{{\scriptsize $\pm$3.47}} /~~4.47\textcolor{gray}{{\scriptsize $\pm$0.92}} & 31.81\textcolor{gray}{{\scriptsize $\pm$3.63}} /~~8.90\textcolor{gray}{{\scriptsize $\pm$1.40}} & 35.38\textcolor{gray}{{\scriptsize $\pm$8.43}} /10.17\textcolor{gray}{{\scriptsize $\pm$2.30}} & 35.69\textcolor{gray}{{\scriptsize $\pm$1.65}} /11.02\textcolor{gray}{{\scriptsize $\pm$1.35}} \\
Random & & & 16.33\textcolor{gray}{{\scriptsize $\pm$1.34}} /~~3.66\textcolor{gray}{{\scriptsize $\pm$0.57}} & 31.67\textcolor{gray}{{\scriptsize $\pm$3.23}} /~~7.51\textcolor{gray}{{\scriptsize $\pm$1.20}} & 32.98\textcolor{gray}{{\scriptsize $\pm$2.74}} /~~8.90\textcolor{gray}{{\scriptsize $\pm$1.23}} & 38.75\textcolor{gray}{{\scriptsize $\pm$2.49}} /10.76\textcolor{gray}{{\scriptsize $\pm$0.97}} \\\hline
PTL & \multirow{2}{*}{HERIDAL} & \multirow{2}{*}{14.64\textcolor{gray}{{\scriptsize $\pm$6.04}} /~~4.59\textcolor{gray}{{\scriptsize $\pm$1.47}}} & 25.14\textcolor{gray}{{\scriptsize $\pm$9.32}} /~~8.23\textcolor{gray}{{\scriptsize $\pm$2.54}} & 35.00\textcolor{gray}{{\scriptsize $\pm$6.30}} /12.15\textcolor{gray}{{\scriptsize $\pm$1.58}} & 37.31\textcolor{gray}{{\scriptsize $\pm$4.15}} /13.38\textcolor{gray}{{\scriptsize $\pm$0.73}} & 38.54\textcolor{gray}{{\scriptsize $\pm$7.75}} /13.37\textcolor{gray}{{\scriptsize $\pm$2.36}} \\
Random & & & 23.77\textcolor{gray}{{\scriptsize $\pm$7.78}} /~~7.43\textcolor{gray}{{\scriptsize $\pm$2.13}} & 34.28\textcolor{gray}{{\scriptsize $\pm$7.17}} /11.83\textcolor{gray}{{\scriptsize $\pm$2.37}} & 35.89\textcolor{gray}{{\scriptsize $\pm$4.74}} /12.69\textcolor{gray}{{\scriptsize $\pm$1.33}} & 40.59\textcolor{gray}{{\scriptsize $\pm$5.17}} /14.18\textcolor{gray}{{\scriptsize $\pm$2.50}} \\\hline
PTL & \multirow{2}{*}{SARD} & \multirow{2}{*}{18.27\textcolor{gray}{{\scriptsize $\pm$1.25}} /~~5.42\textcolor{gray}{{\scriptsize $\pm$0.50}}} & 31.16\textcolor{gray}{{\scriptsize $\pm$5.12}} /~~9.58\textcolor{gray}{{\scriptsize $\pm$1.86}} & 42.93\textcolor{gray}{{\scriptsize $\pm$2.58}} /13.98\textcolor{gray}{{\scriptsize $\pm$1.20}} & 46.04\textcolor{gray}{{\scriptsize $\pm$2.30}} /15.58\textcolor{gray}{{\scriptsize $\pm$0.73}} & 48.15\textcolor{gray}{{\scriptsize $\pm$1.18}} /17.05\textcolor{gray}{{\scriptsize $\pm$0.87}} \\
Random & & & 30.55\textcolor{gray}{{\scriptsize $\pm$2.77}} /~~9.37\textcolor{gray}{{\scriptsize $\pm$0.68}} & 40.69\textcolor{gray}{{\scriptsize $\pm$0.34}} /13.13\textcolor{gray}{{\scriptsize $\pm$0.12}} & 44.10\textcolor{gray}{{\scriptsize $\pm$3.84}} /14.74\textcolor{gray}{{\scriptsize $\pm$1.62}} & 45.62\textcolor{gray}{{\scriptsize $\pm$5.05}} /15.13\textcolor{gray}{{\scriptsize $\pm$2.15}} \\
\end{tabular}
}
\end{subtable}
\vspace{0.3cm}
\\
\begin{subtable}[b]{\linewidth}
\caption{Vis-200}
\vspace{-0.3cm}
\centering
\resizebox{\linewidth}{!}{%
\setlength{\tabcolsep}{5.0pt}
\renewcommand{\arraystretch}{1.2}
\begin{tabular}{l|c|rrrrr}
& & \multicolumn{5}{c}{\# of synthetic image} \\
\multicolumn{1}{c|}{method} & test set & \multicolumn{1}{c}{0} & \multicolumn{1}{c}{100} & \multicolumn{1}{c}{500} & \multicolumn{1}{c}{1000} & \multicolumn{1}{c}{2000} \\\hline
PTL & \multirow{2}{*}{VisDrone} & \multirow{2}{*}{10.55\textcolor{gray}{{\scriptsize $\pm$1.41}} /~~3.18\textcolor{gray}{{\scriptsize $\pm$0.55}}} & 11.65\textcolor{gray}{{\scriptsize $\pm$0.87}} /~~3.55\textcolor{gray}{{\scriptsize $\pm$0.35}} & 12.74\textcolor{gray}{{\scriptsize $\pm$0.90}} /~~3.99\textcolor{gray}{{\scriptsize $\pm$0.35}} & 12.96\textcolor{gray}{{\scriptsize $\pm$0.68}} /~~4.16\textcolor{gray}{{\scriptsize $\pm$0.43}} & 12.78\textcolor{gray}{{\scriptsize $\pm$0.48}} /~~4.10\textcolor{gray}{{\scriptsize $\pm$0.28}} \\
Random & & & 11.35\textcolor{gray}{{\scriptsize $\pm$1.23}} /~~3.42\textcolor{gray}{{\scriptsize $\pm$0.52}} & 12.07\textcolor{gray}{{\scriptsize $\pm$1.11}} /~~3.72\textcolor{gray}{{\scriptsize $\pm$0.46}} & 12.63\textcolor{gray}{{\scriptsize $\pm$0.59}} /~~3.95\textcolor{gray}{{\scriptsize $\pm$0.36}} & 12.59\textcolor{gray}{{\scriptsize $\pm$1.17}} /~~3.97\textcolor{gray}{{\scriptsize $\pm$0.49}} \\\hline
PTL & \multirow{2}{*}{Okutama} & \multirow{2}{*}{38.58\textcolor{gray}{{\scriptsize $\pm$4.81}} /10.25\textcolor{gray}{{\scriptsize $\pm$1.66}}} & 40.23\textcolor{gray}{{\scriptsize $\pm$2.20}} /10.71\textcolor{gray}{{\scriptsize $\pm$1.12}} & 45.56\textcolor{gray}{{\scriptsize $\pm$0.44}} /12.44\textcolor{gray}{{\scriptsize $\pm$0.51}} & 47.79\textcolor{gray}{{\scriptsize $\pm$2.47}} /12.99\textcolor{gray}{{\scriptsize $\pm$0.39}} & 46.62\textcolor{gray}{{\scriptsize $\pm$0.93}} /12.37\textcolor{gray}{{\scriptsize $\pm$0.21}} \\
Random & & & 39.99\textcolor{gray}{{\scriptsize $\pm$2.82}} /10.43\textcolor{gray}{{\scriptsize $\pm$0.48}} & 39.95\textcolor{gray}{{\scriptsize $\pm$2.33}} /10.50\textcolor{gray}{{\scriptsize $\pm$0.71}} & 41.88\textcolor{gray}{{\scriptsize $\pm$1.66}} /11.02\textcolor{gray}{{\scriptsize $\pm$0.74}} & 41.20\textcolor{gray}{{\scriptsize $\pm$0.92}} /10.55\textcolor{gray}{{\scriptsize $\pm$0.03}} \\\hline
PTL & \multirow{2}{*}{ICG} & \multirow{2}{*}{6.50\textcolor{gray}{{\scriptsize $\pm$3.17}} /~~1.39\textcolor{gray}{{\scriptsize $\pm$0.63}}} & 9.50\textcolor{gray}{{\scriptsize $\pm$1.65}} /~~2.22\textcolor{gray}{{\scriptsize $\pm$0.44}} & 20.56\textcolor{gray}{{\scriptsize $\pm$2.32}} /~~5.01\textcolor{gray}{{\scriptsize $\pm$0.40}} & 25.67\textcolor{gray}{{\scriptsize $\pm$4.82}} /~~6.44\textcolor{gray}{{\scriptsize $\pm$1.65}} & 30.48\textcolor{gray}{{\scriptsize $\pm$0.34}} /~~8.21\textcolor{gray}{{\scriptsize $\pm$0.43}} \\
Random & & & 9.27\textcolor{gray}{{\scriptsize $\pm$2.16}} /~~2.49\textcolor{gray}{{\scriptsize $\pm$0.16}} & 18.29\textcolor{gray}{{\scriptsize $\pm$5.19}} /~~4.45\textcolor{gray}{{\scriptsize $\pm$1.31}} & 23.29\textcolor{gray}{{\scriptsize $\pm$5.46}} /~~5.98\textcolor{gray}{{\scriptsize $\pm$1.31}} & 27.29\textcolor{gray}{{\scriptsize $\pm$4.52}} /~~7.41\textcolor{gray}{{\scriptsize $\pm$1.46}} \\\hline
PTL & \multirow{2}{*}{HERIDAL} & \multirow{2}{*}{14.76\textcolor{gray}{{\scriptsize $\pm$9.76}} /~~4.68\textcolor{gray}{{\scriptsize $\pm$2.72}}} & 19.87\textcolor{gray}{{\scriptsize $\pm$3.79}} /~~6.34\textcolor{gray}{{\scriptsize $\pm$0.69}} & 30.51\textcolor{gray}{{\scriptsize $\pm$6.31}} /10.34\textcolor{gray}{{\scriptsize $\pm$1.39}} & 34.76\textcolor{gray}{{\scriptsize $\pm$8.89}} /12.21\textcolor{gray}{{\scriptsize $\pm$2.76}} & 37.60\textcolor{gray}{{\scriptsize $\pm$2.12}} /13.74\textcolor{gray}{{\scriptsize $\pm$1.46}} \\
Random & & & 17.66\textcolor{gray}{{\scriptsize $\pm$7.26}} /~~5.72\textcolor{gray}{{\scriptsize $\pm$2.29}} & 26.26\textcolor{gray}{{\scriptsize $\pm$9.95}} /~~8.55\textcolor{gray}{{\scriptsize $\pm$3.30}} & 29.62\textcolor{gray}{{\scriptsize $\pm$6.13}} /10.09\textcolor{gray}{{\scriptsize $\pm$1.44}} & 33.74\textcolor{gray}{{\scriptsize $\pm$8.06}} /11.69\textcolor{gray}{{\scriptsize $\pm$2.56}} \\\hline
PTL & \multirow{2}{*}{SARD} & \multirow{2}{*}{21.87\textcolor{gray}{{\scriptsize $\pm$7.48}} /~~6.98\textcolor{gray}{{\scriptsize $\pm$1.97}}} & 30.17\textcolor{gray}{{\scriptsize $\pm$4.28}} /~~9.67\textcolor{gray}{{\scriptsize $\pm$1.03}} & 40.01\textcolor{gray}{{\scriptsize $\pm$2.13}} /13.35\textcolor{gray}{{\scriptsize $\pm$0.57}} & 46.91\textcolor{gray}{{\scriptsize $\pm$5.03}} /16.13\textcolor{gray}{{\scriptsize $\pm$1.69}} & 49.60\textcolor{gray}{{\scriptsize $\pm$1.35}} /17.73\textcolor{gray}{{\scriptsize $\pm$0.49}} \\
Random & & & 27.51\textcolor{gray}{{\scriptsize $\pm$5.72}} /~~8.74\textcolor{gray}{{\scriptsize $\pm$1.69}} & 38.96\textcolor{gray}{{\scriptsize $\pm$5.45}} /12.43\textcolor{gray}{{\scriptsize $\pm$2.31}} & 44.59\textcolor{gray}{{\scriptsize $\pm$1.56}} /14.80\textcolor{gray}{{\scriptsize $\pm$0.82}} & 43.25\textcolor{gray}{{\scriptsize $\pm$5.80}} /14.85\textcolor{gray}{{\scriptsize $\pm$2.10}} \\
\end{tabular}
}
\end{subtable}
\end{table}

\begin{figure*}[t]
\begin{subfigure}{\linewidth}
\centering
\includegraphics[trim=10mm 0mm 10mm 0mm,clip,width=.19\linewidth]{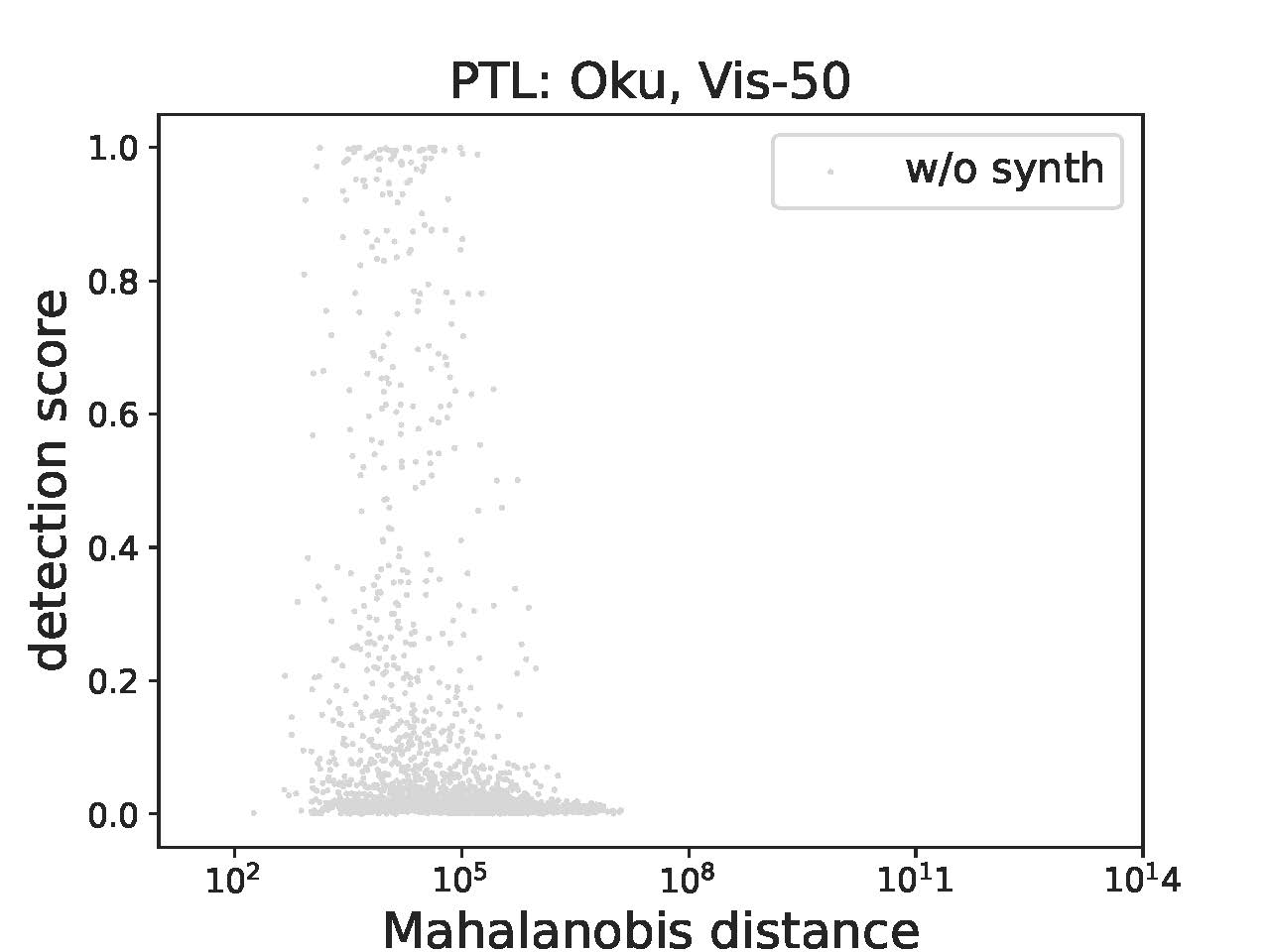}
\includegraphics[trim=10mm 0mm 10mm 0mm,clip,width=.19\linewidth]{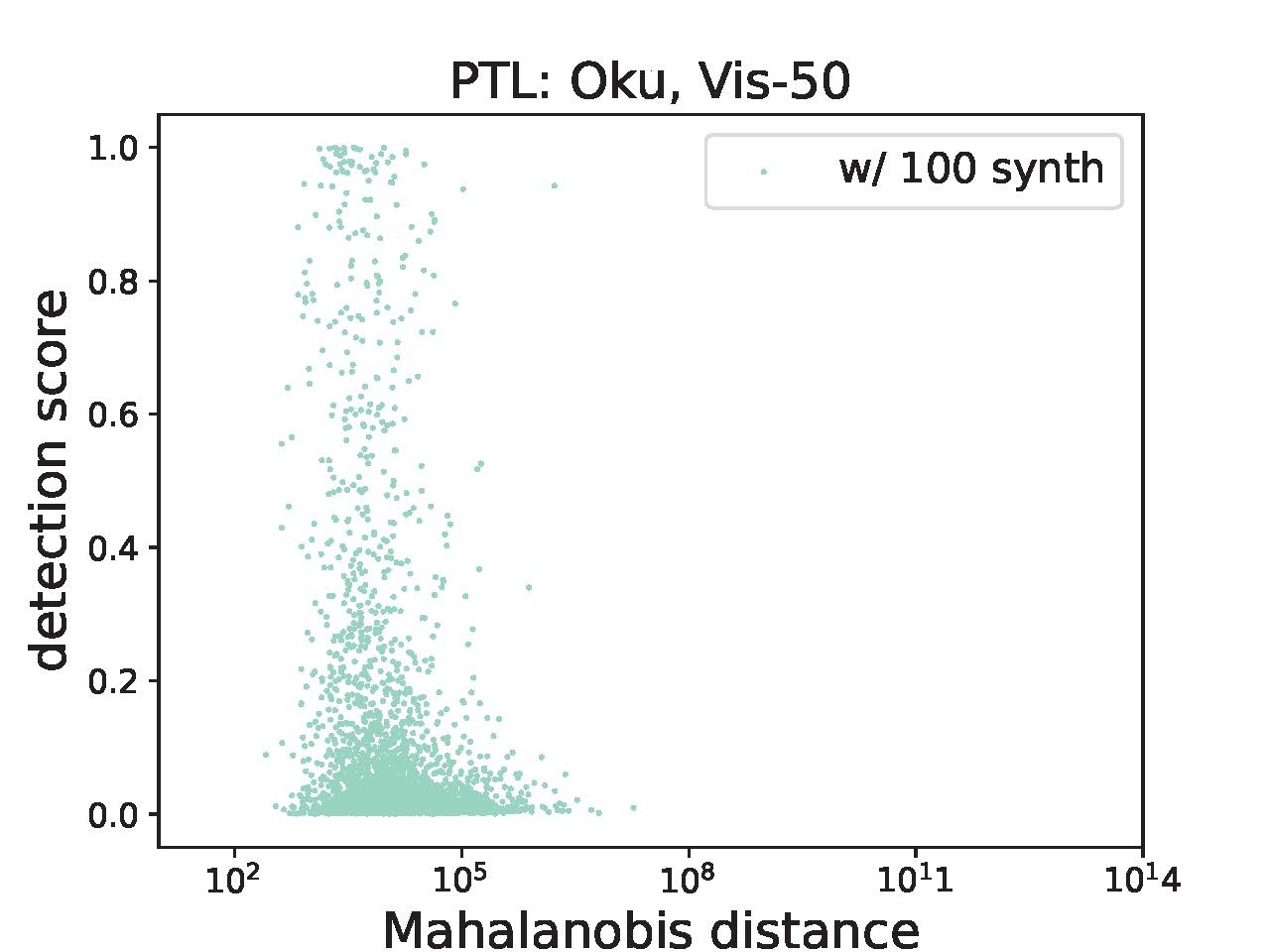}
\includegraphics[trim=10mm 0mm 10mm 0mm,clip,width=.19\linewidth]{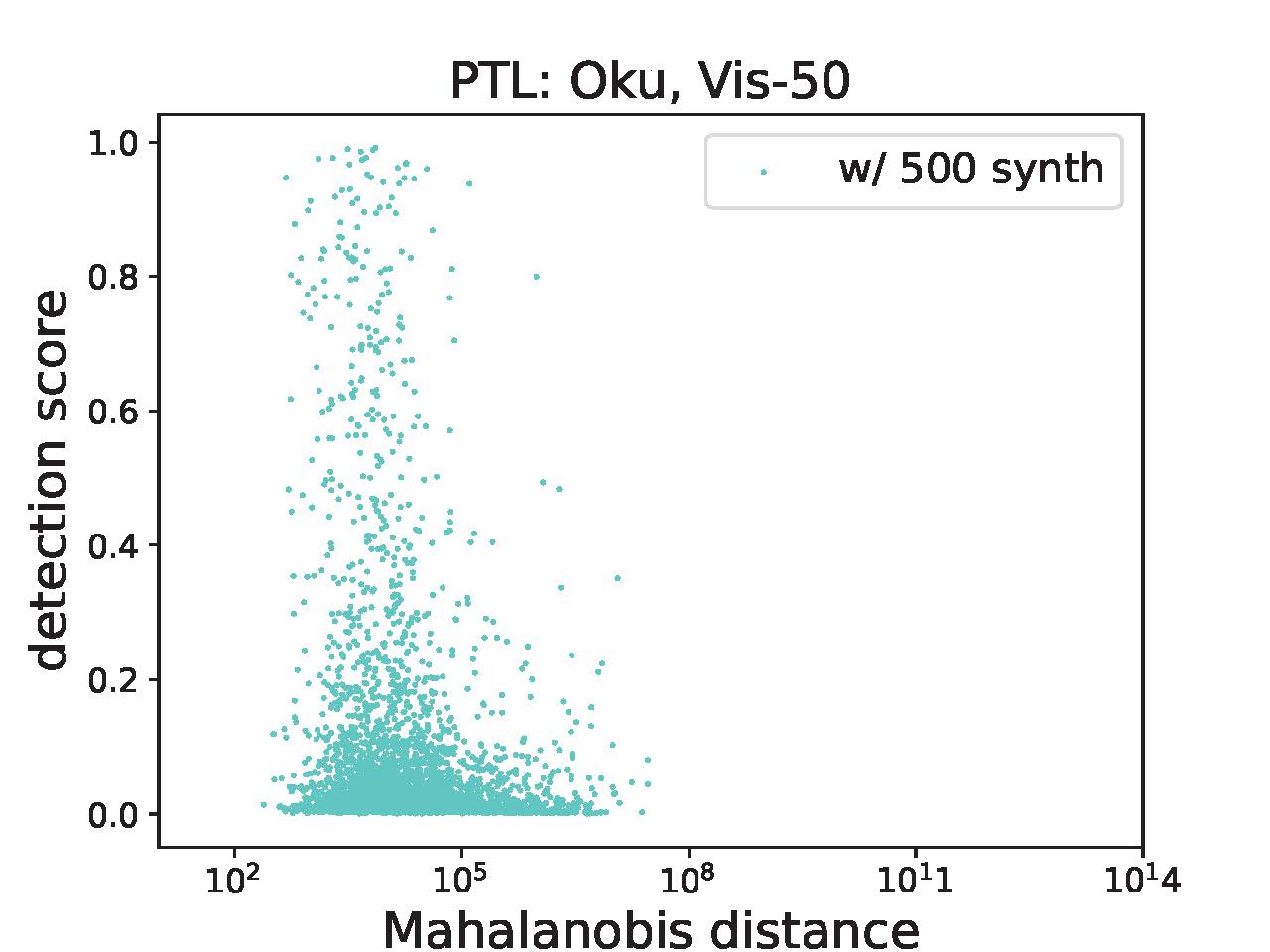}
\includegraphics[trim=10mm 0mm 10mm 0mm,clip,width=.19\linewidth]{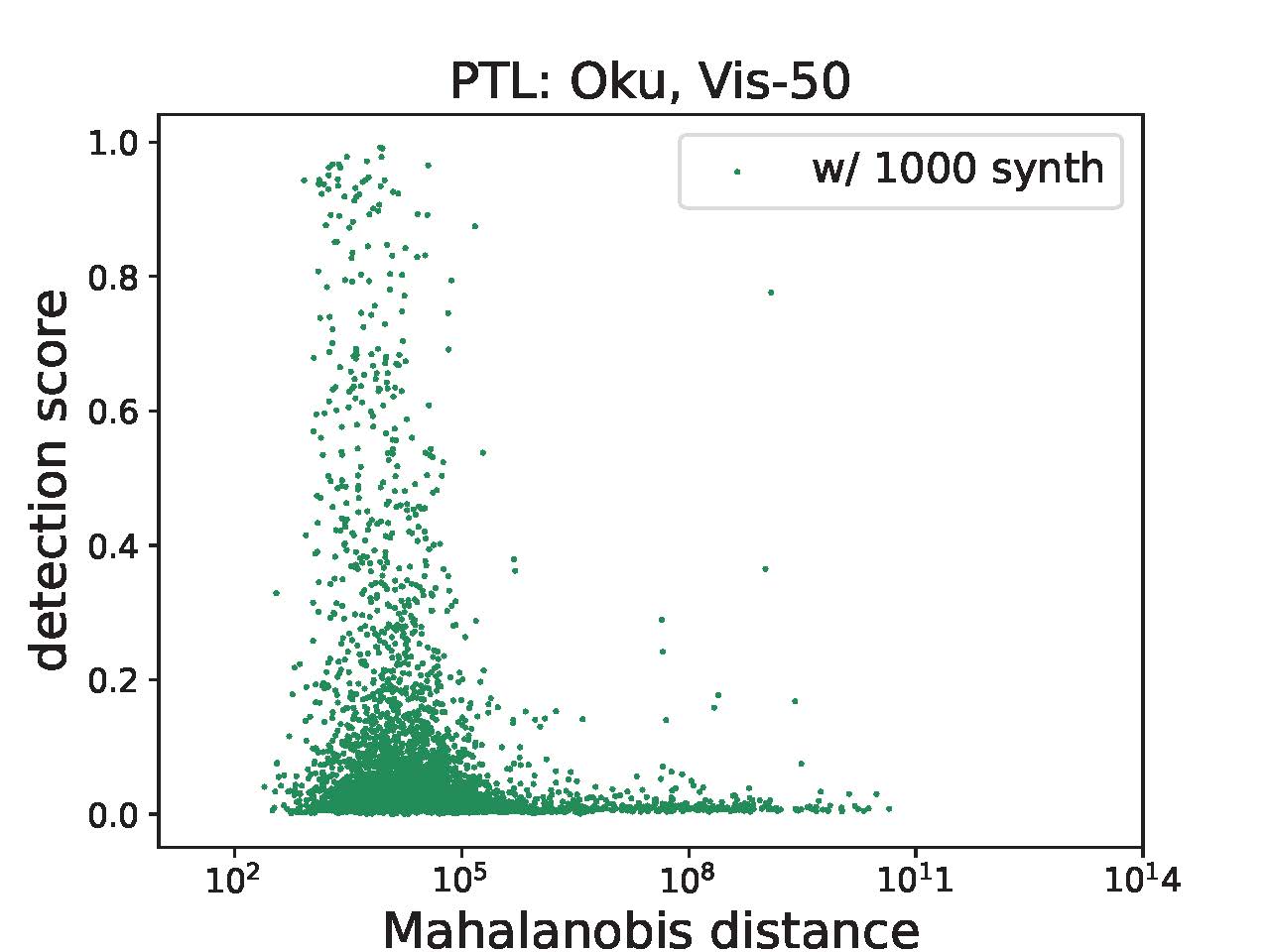}
\includegraphics[trim=10mm 0mm 10mm 0mm,clip,width=.19\linewidth]{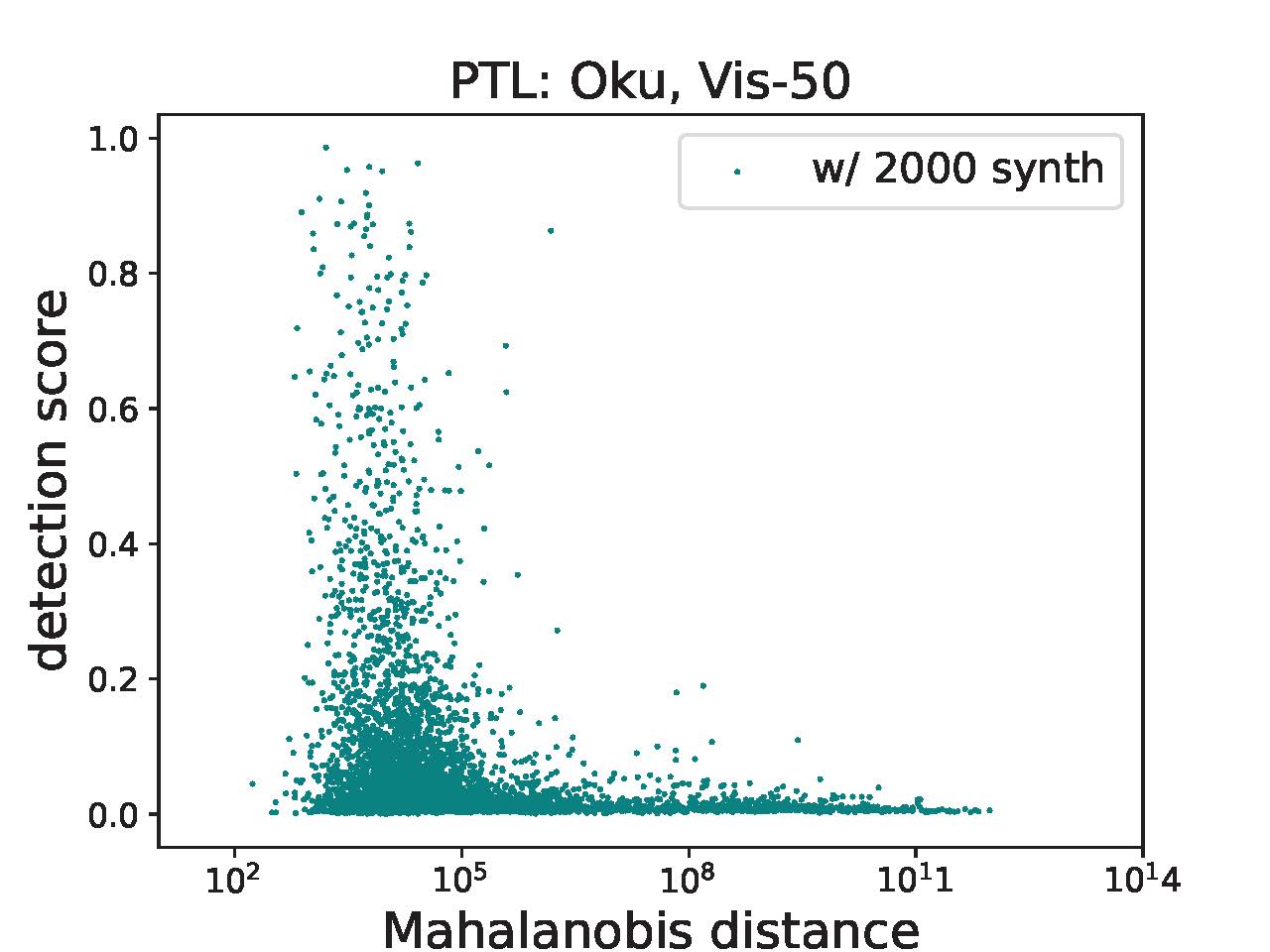}
\\
\includegraphics[trim=10mm 0mm 10mm 0mm,clip,width=.19\linewidth]{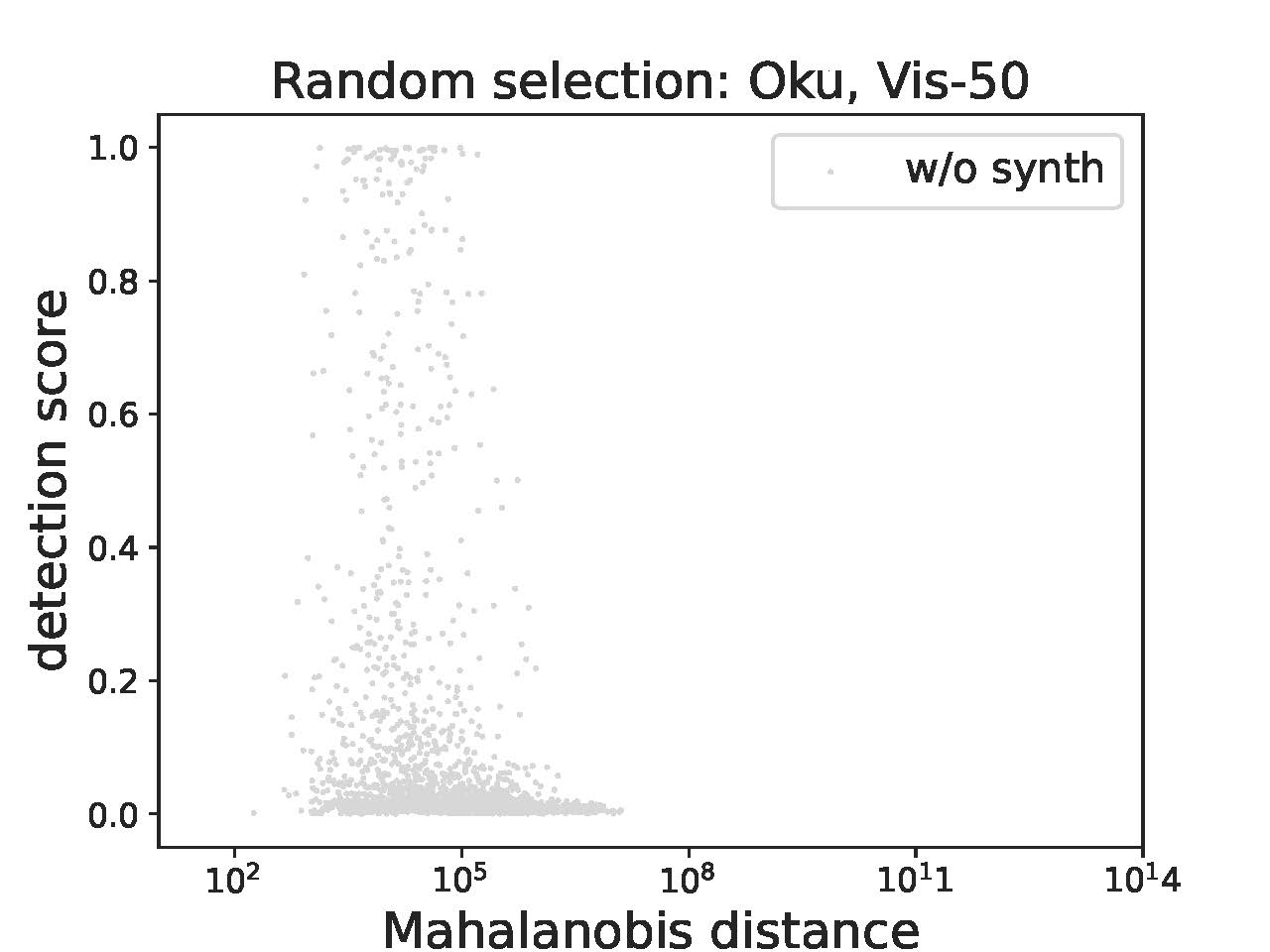}
\includegraphics[trim=10mm 0mm 10mm 0mm,clip,width=.19\linewidth]{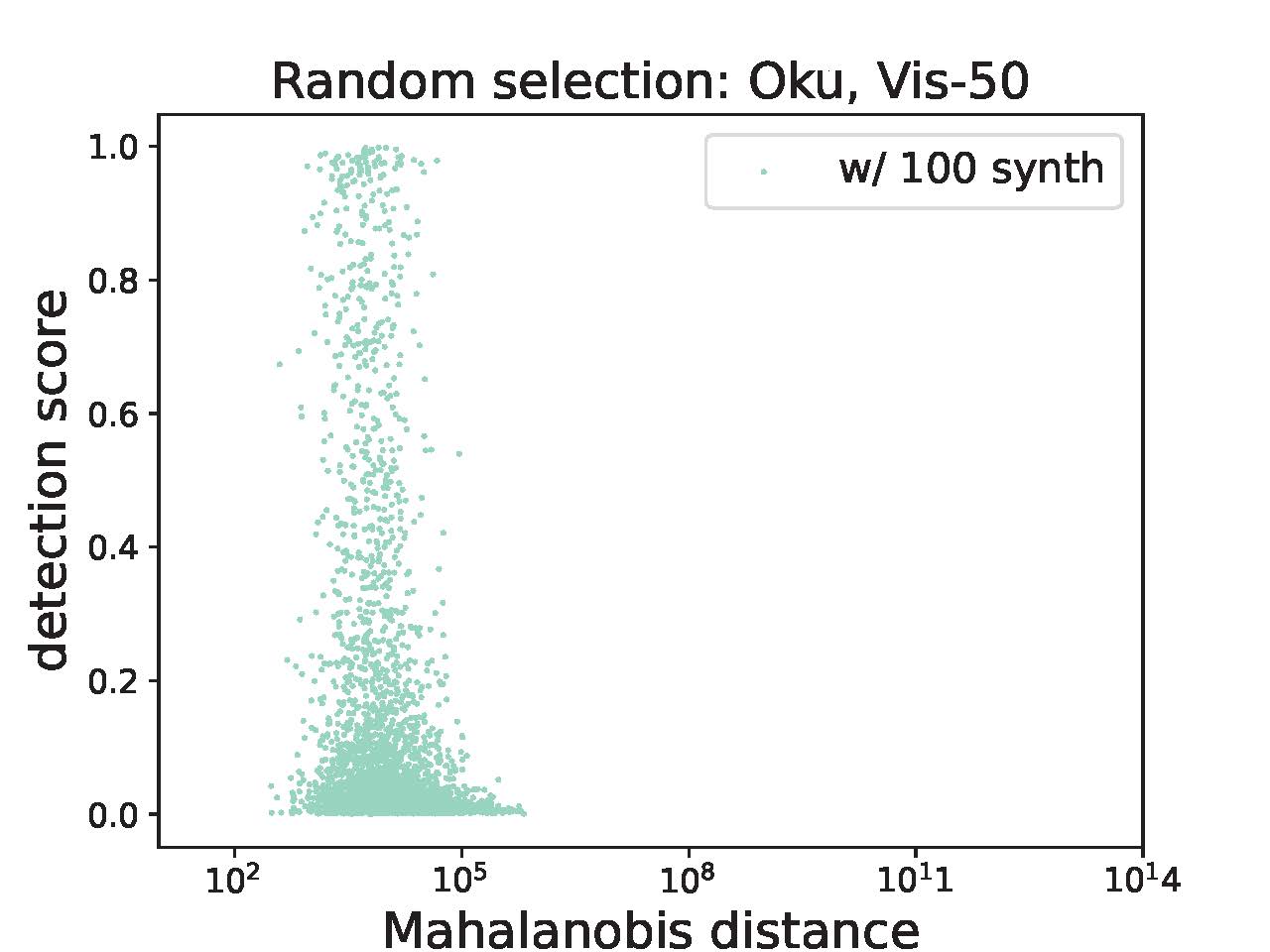}
\includegraphics[trim=10mm 0mm 10mm 0mm,clip,width=.19\linewidth]{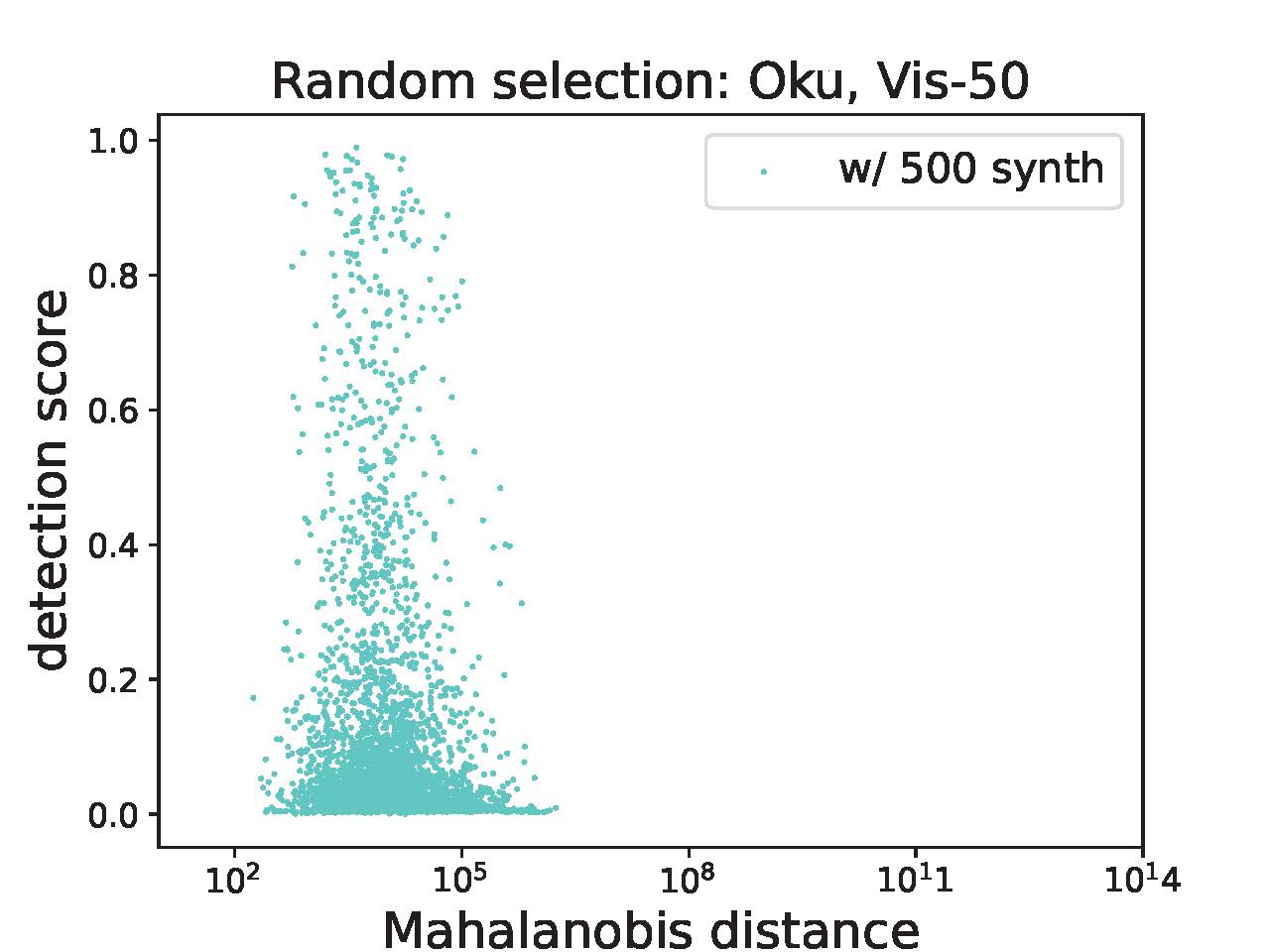}
\includegraphics[trim=10mm 0mm 10mm 0mm,clip,width=.19\linewidth]{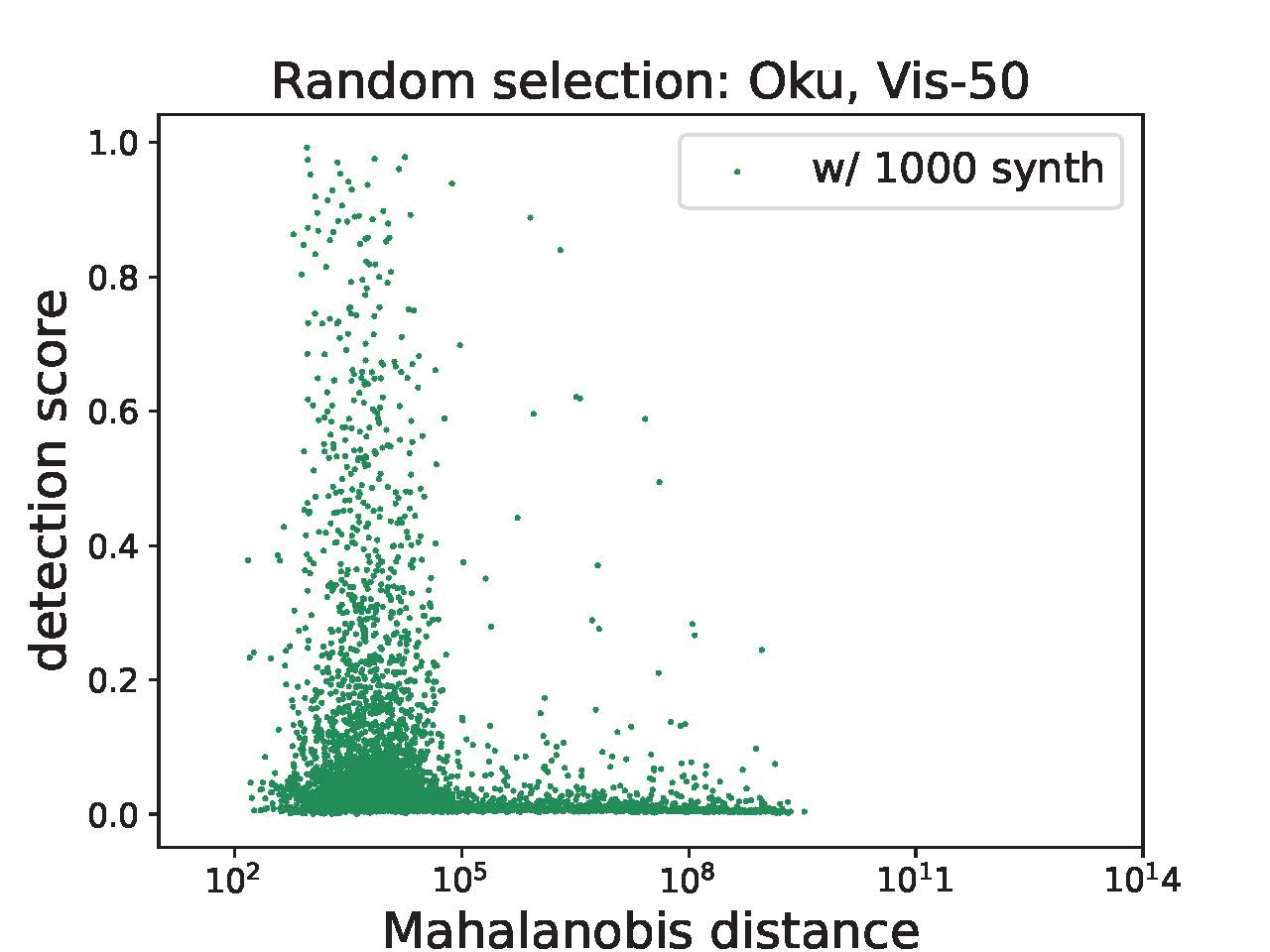}
\includegraphics[trim=10mm 0mm 10mm 0mm,clip,width=.19\linewidth]{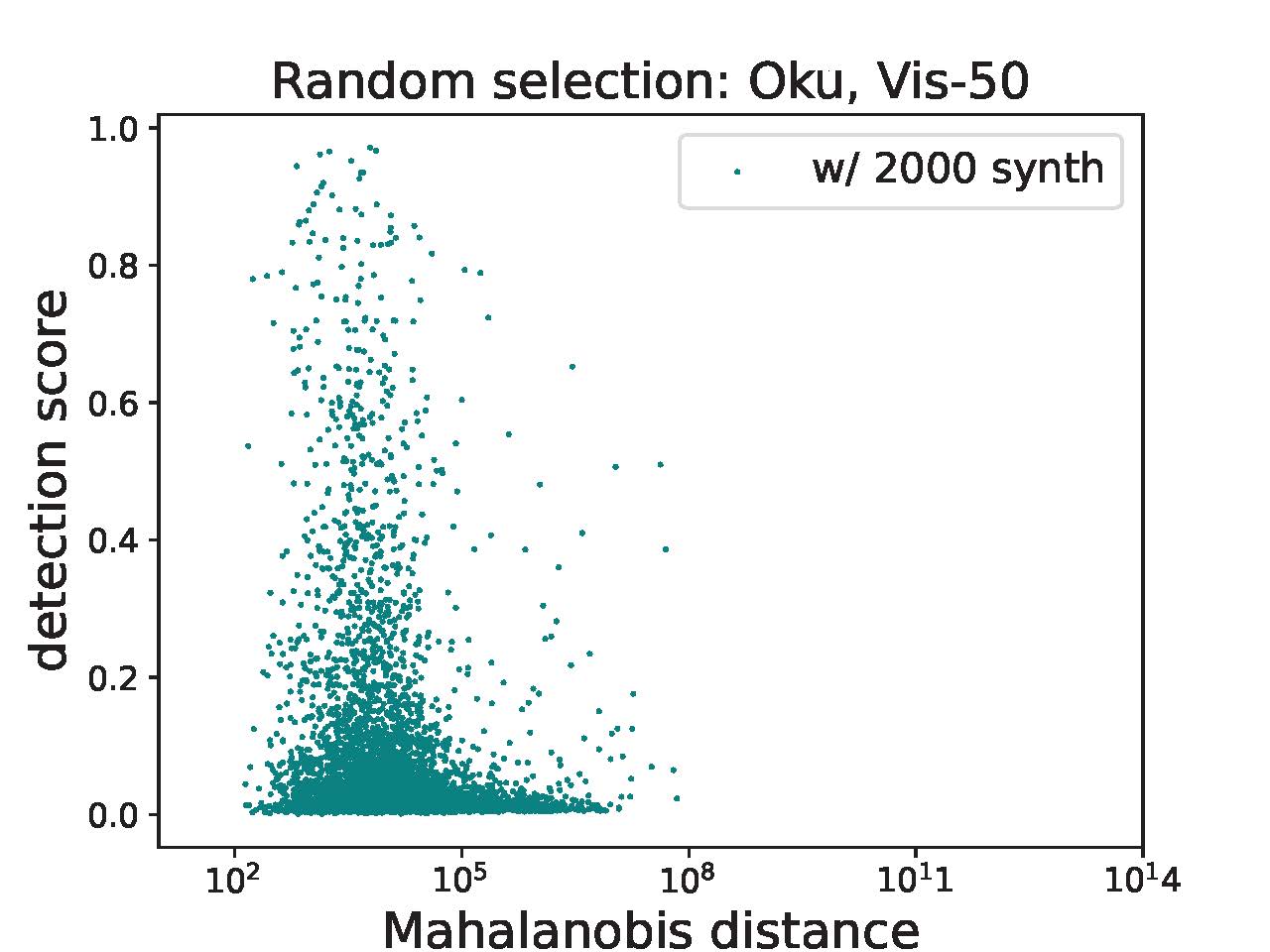}
\caption{Train: Vis-50, Test: Okutama-Action}
\label{fig:scatter_oku_vis50}
\end{subfigure}
\\\\
\begin{subfigure}{\linewidth}
\centering
\includegraphics[trim=10mm 0mm 10mm 0mm,clip,width=.19\linewidth]{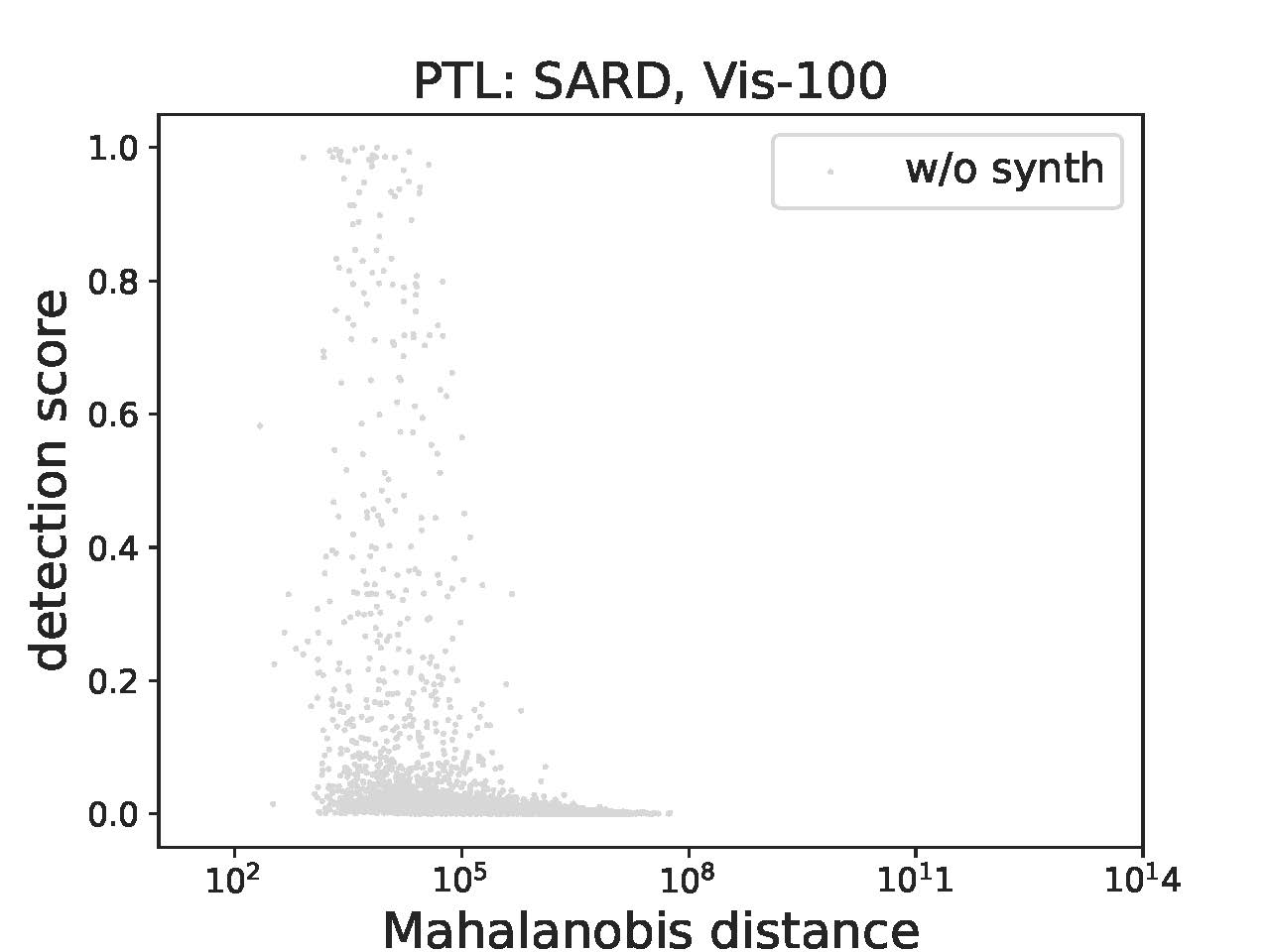}
\includegraphics[trim=10mm 0mm 10mm 0mm,clip,width=.19\linewidth]{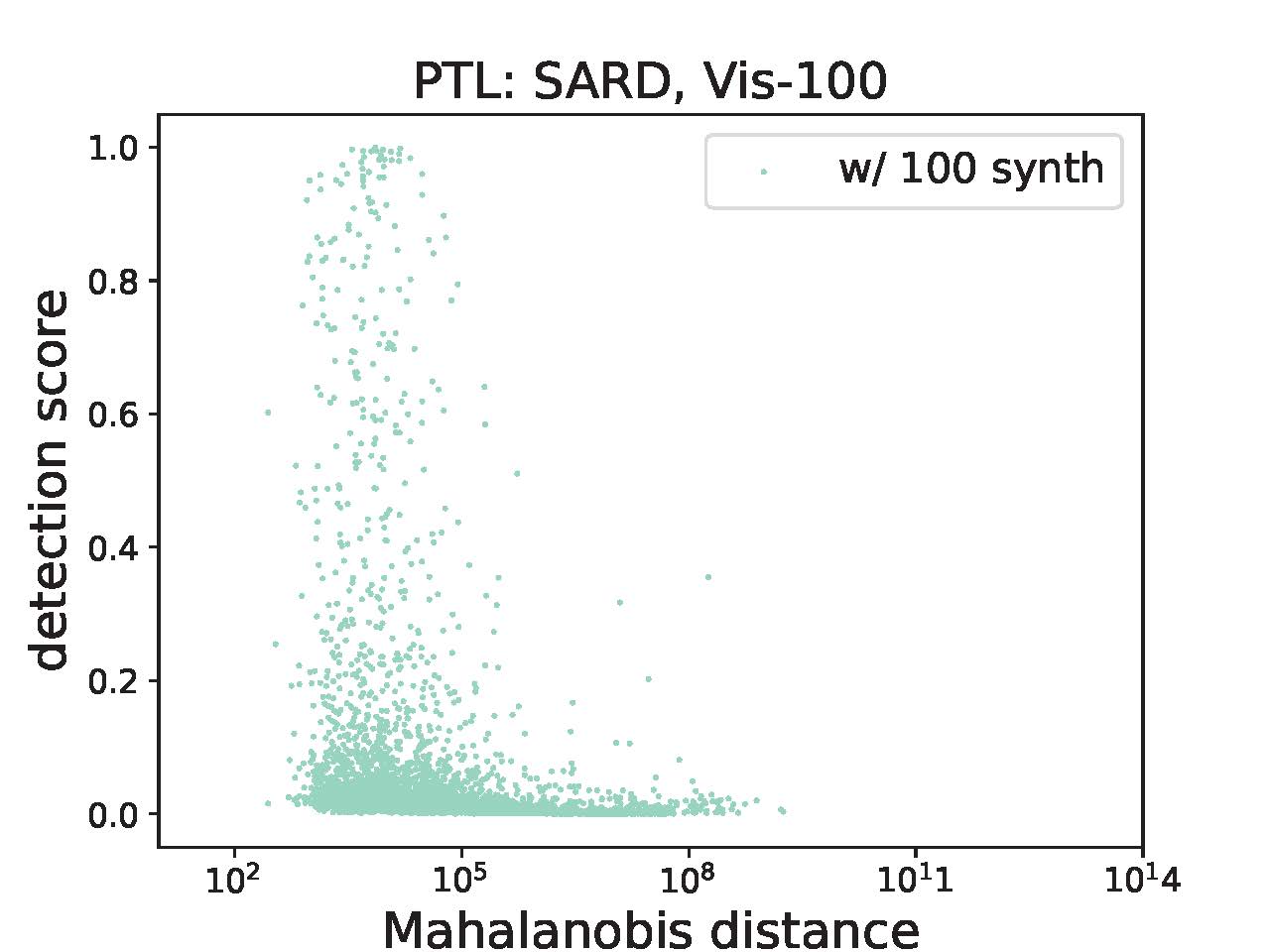}
\includegraphics[trim=10mm 0mm 10mm 0mm,clip,width=.19\linewidth]{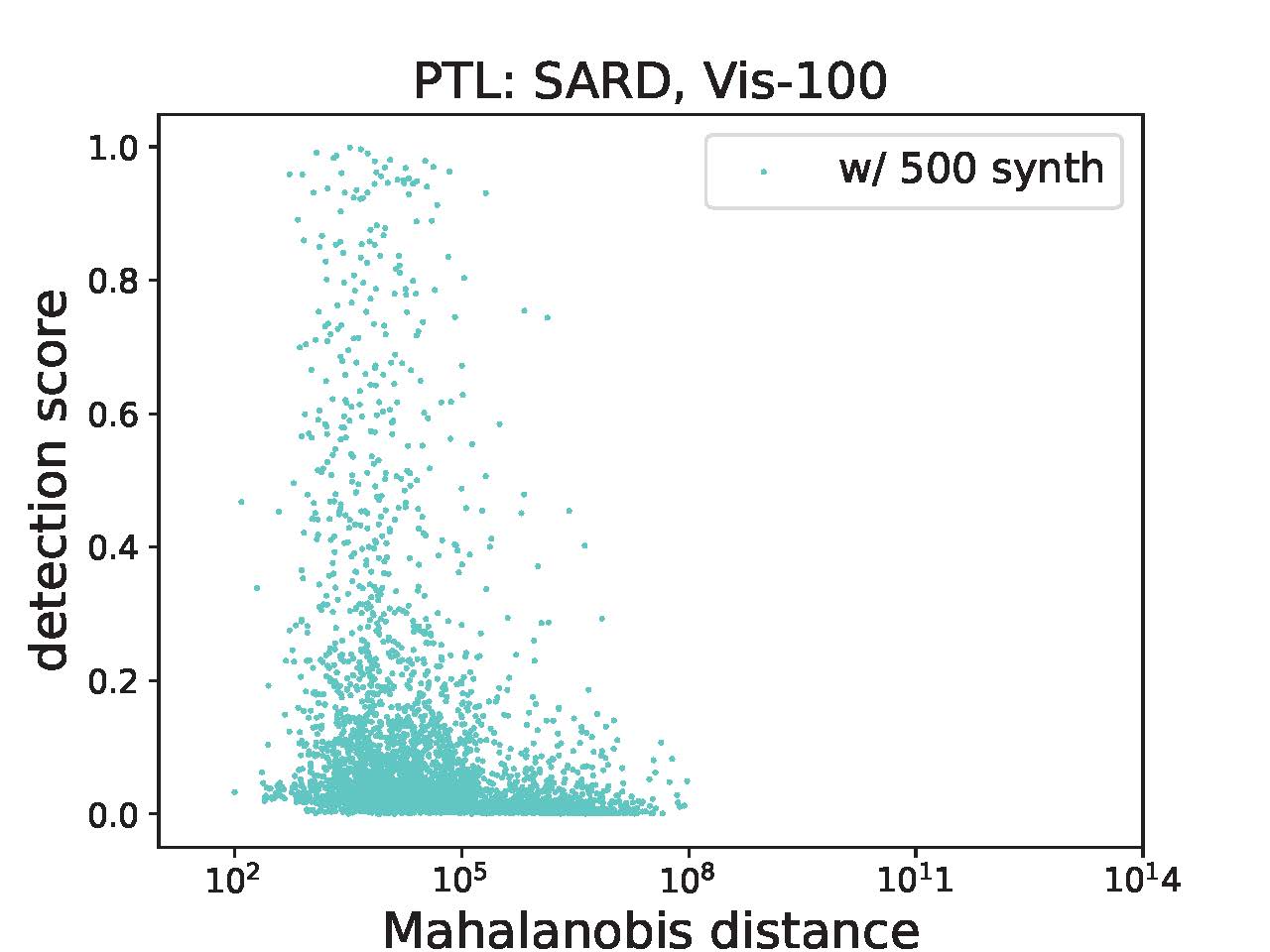}
\includegraphics[trim=10mm 0mm 10mm 0mm,clip,width=.19\linewidth]{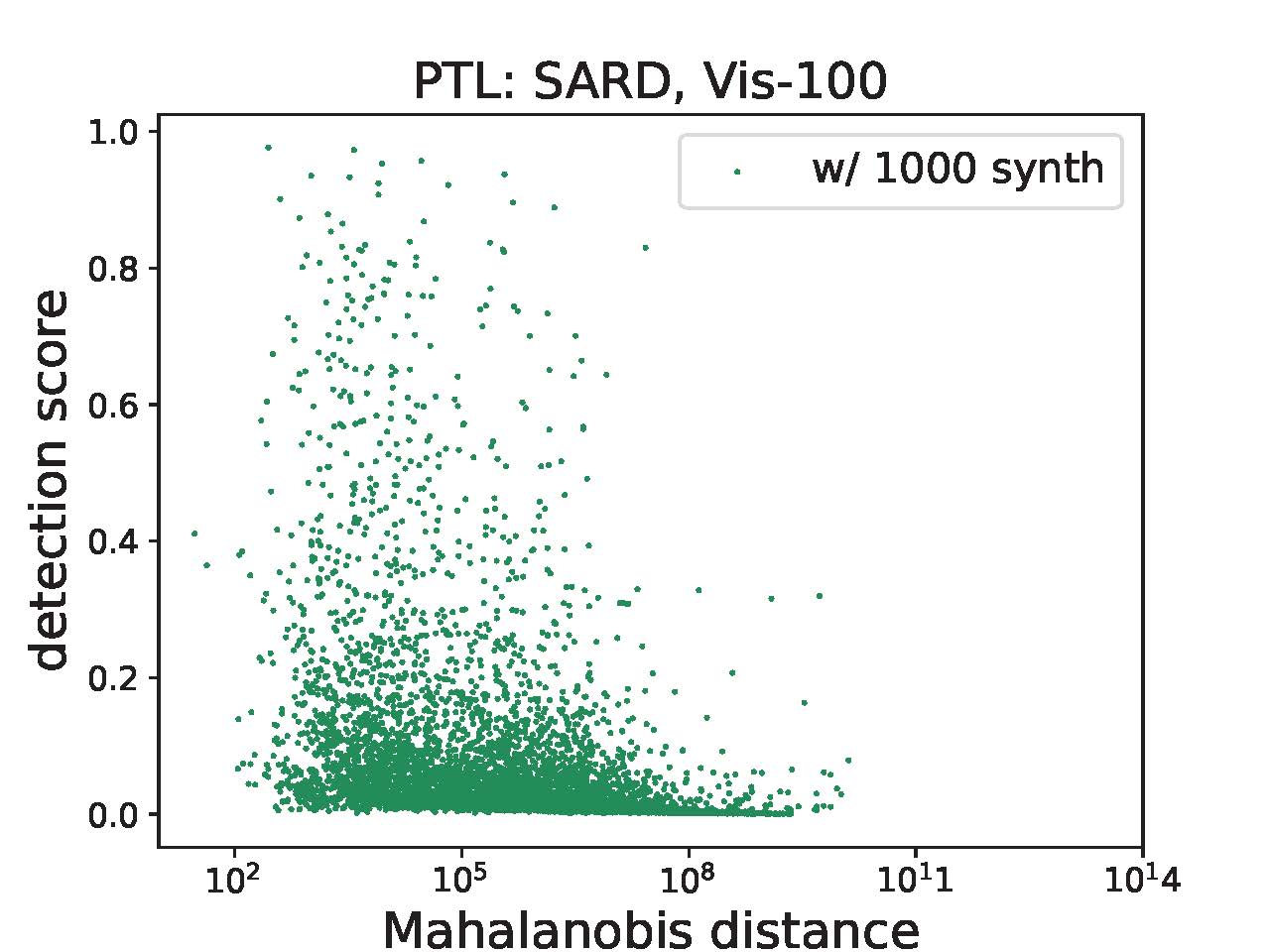}
\includegraphics[trim=10mm 0mm 10mm 0mm,clip,width=.19\linewidth]{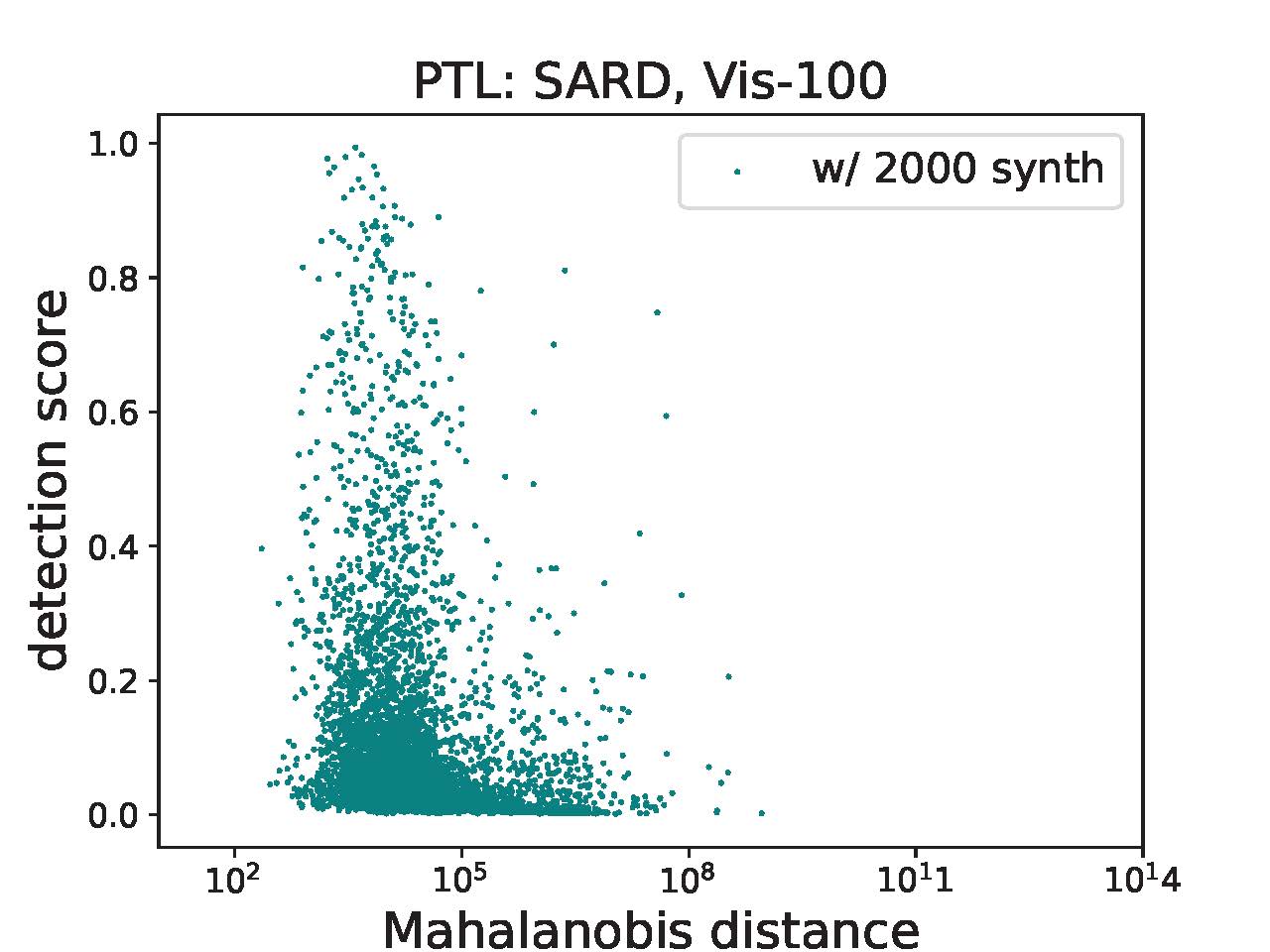}
\\
\includegraphics[trim=10mm 0mm 10mm 0mm,clip,width=.19\linewidth]{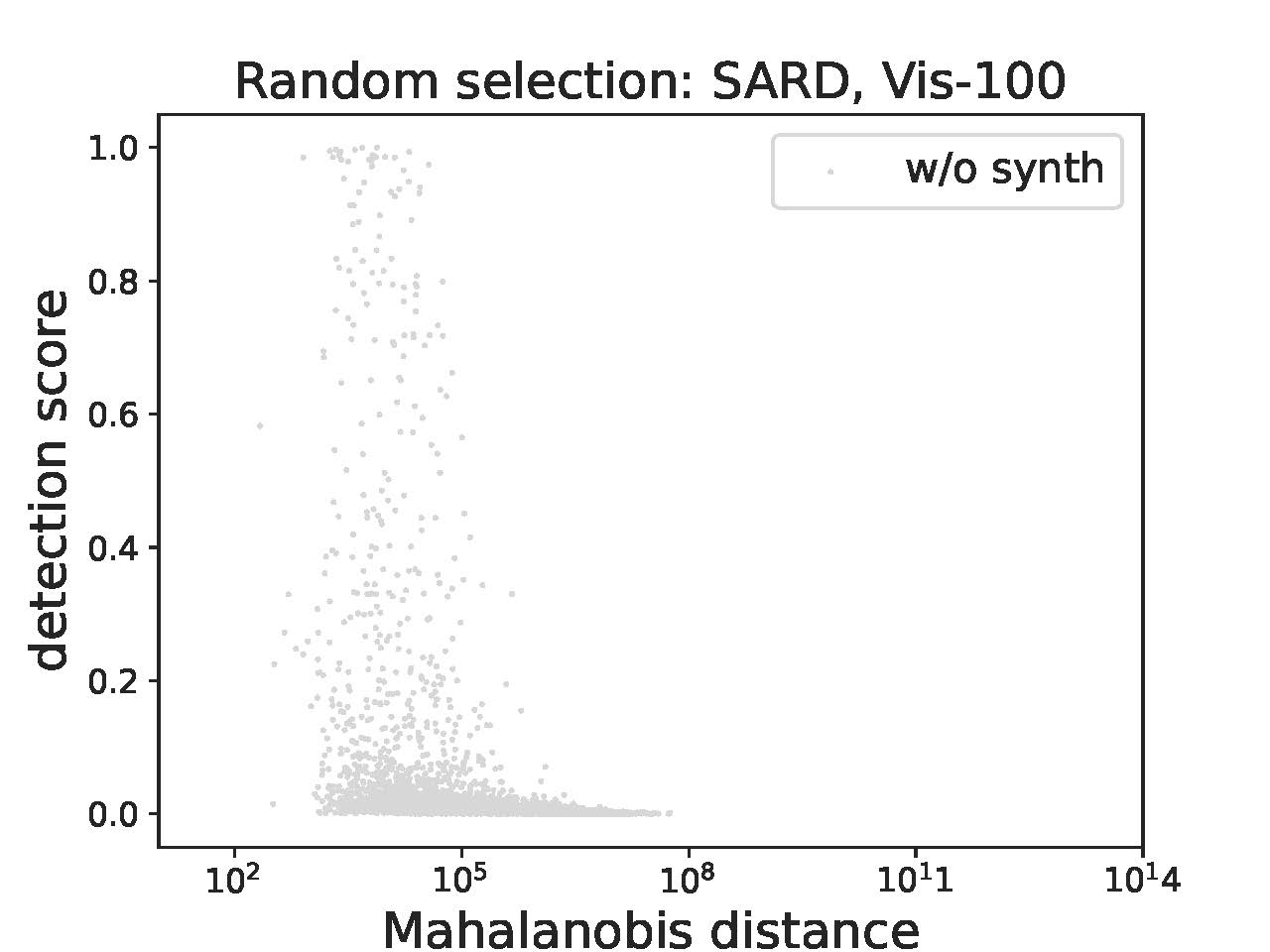}
\includegraphics[trim=10mm 0mm 10mm 0mm,clip,width=.19\linewidth]{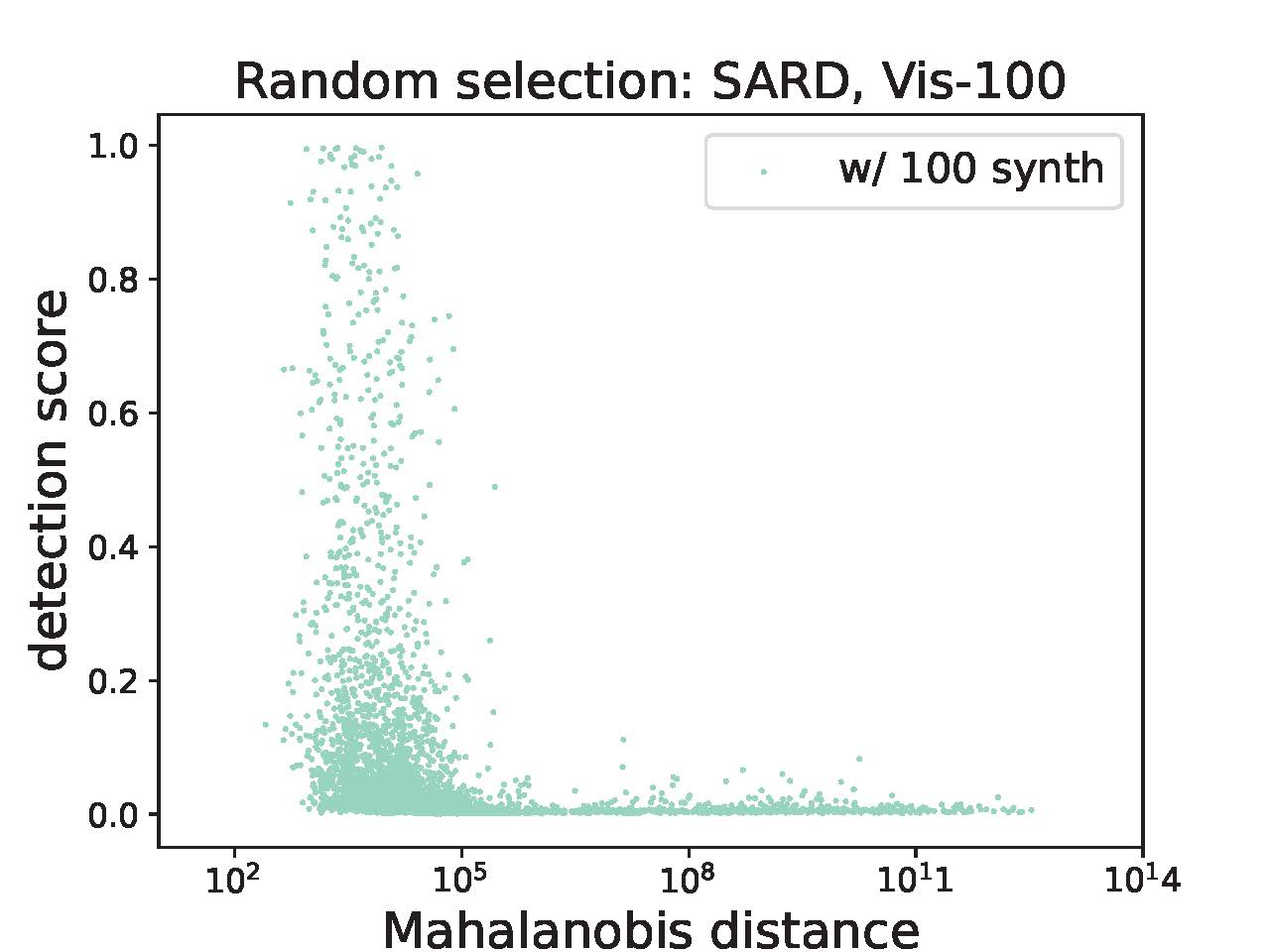}
\includegraphics[trim=10mm 0mm 10mm 0mm,clip,width=.19\linewidth]{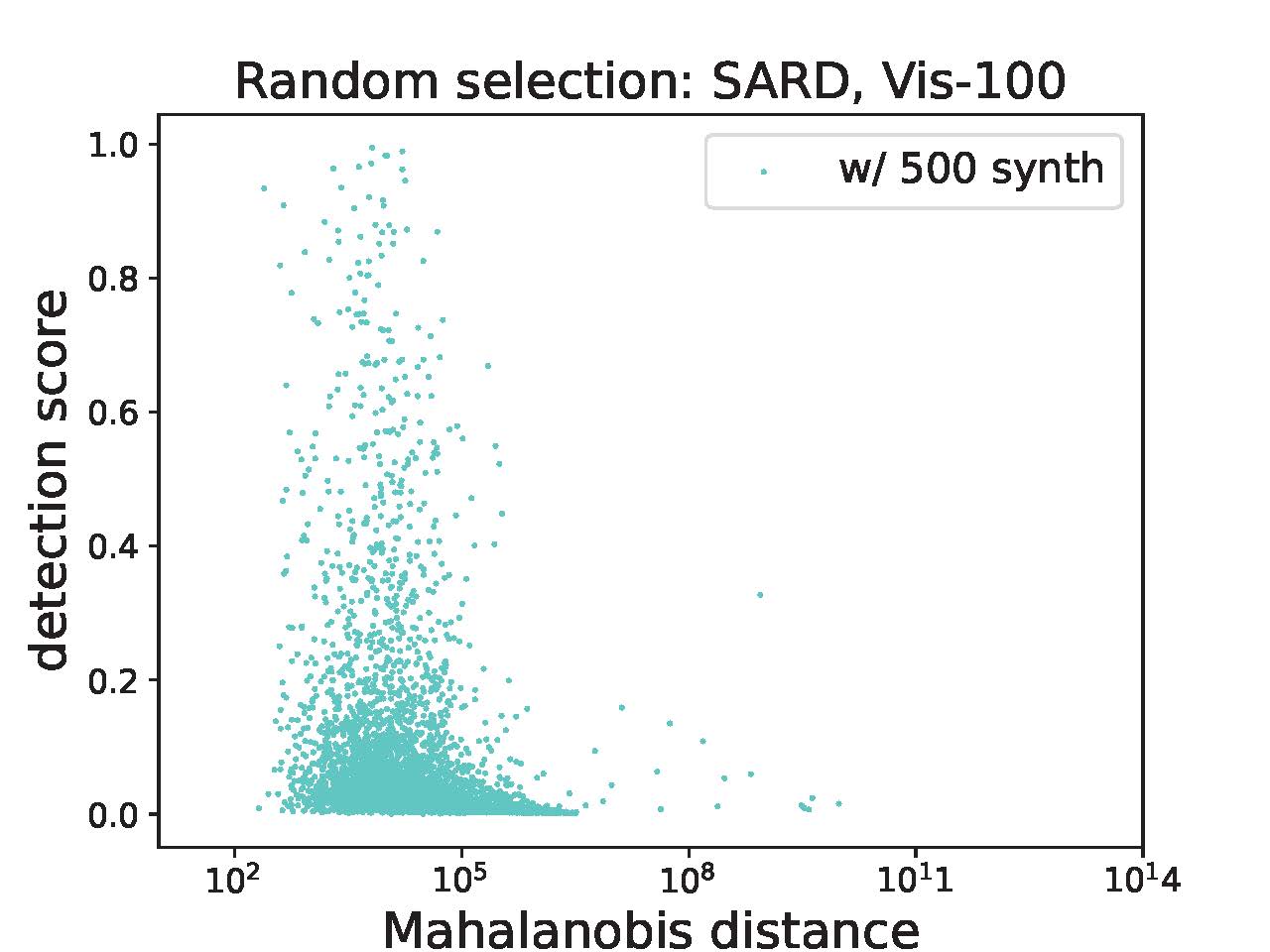}
\includegraphics[trim=10mm 0mm 10mm 0mm,clip,width=.19\linewidth]{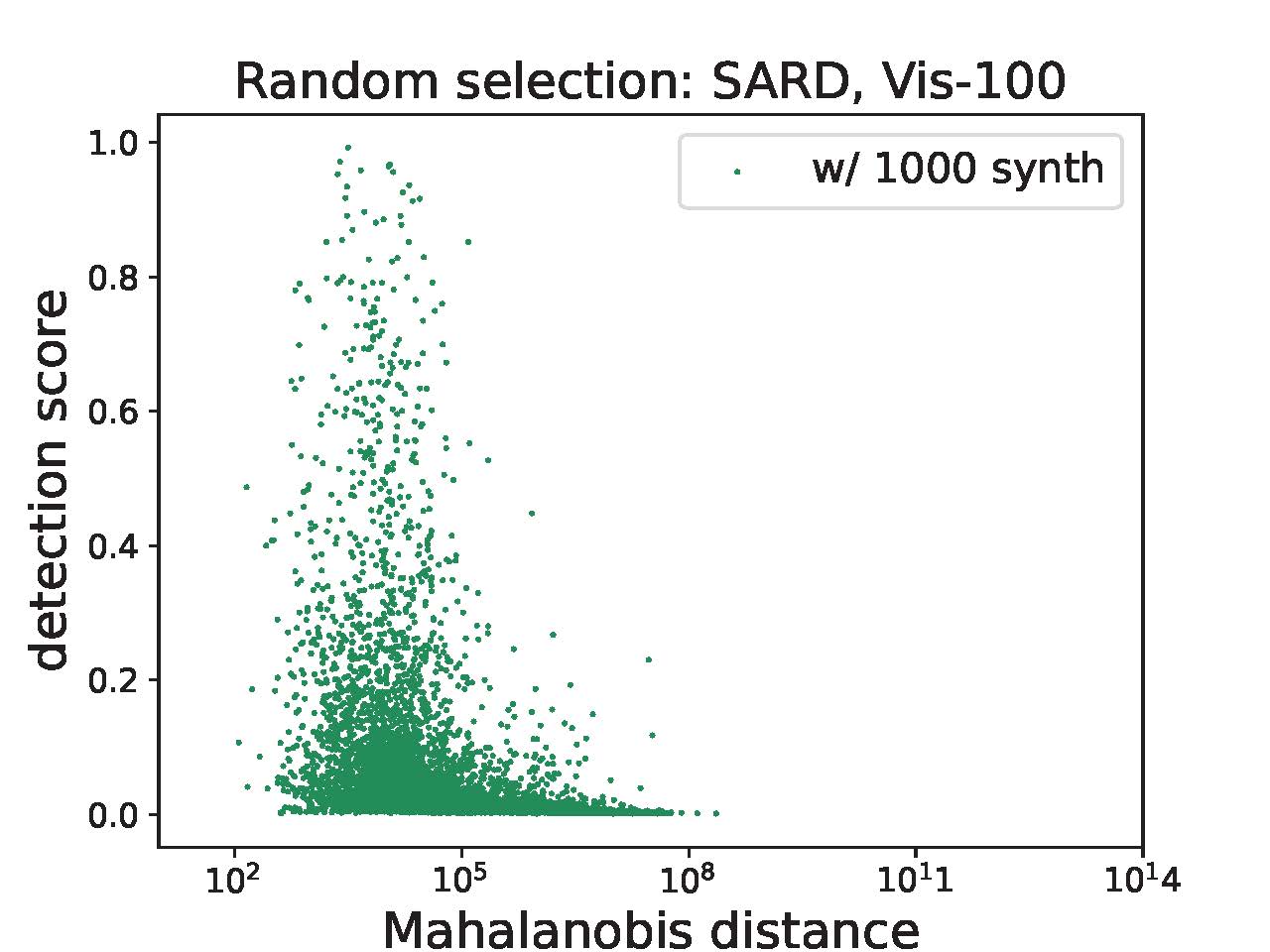}
\includegraphics[trim=10mm 0mm 10mm 0mm,clip,width=.19\linewidth]{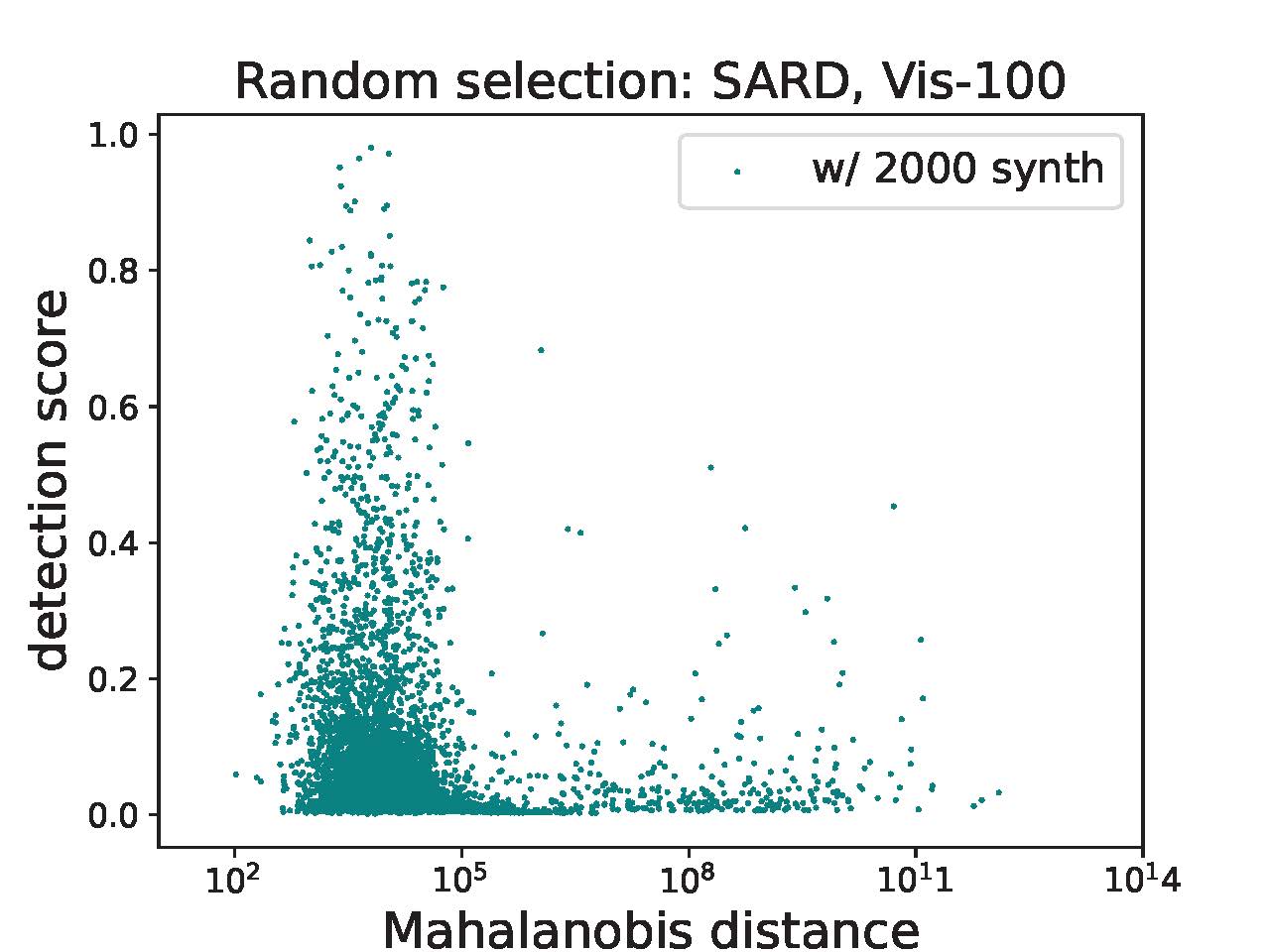}
\caption{Train: Vis-100, Test: SARD}
\label{fig:scatter_sard_vis100}
\end{subfigure}
\\\\
\begin{subfigure}{\linewidth}
\centering
\includegraphics[trim=10mm 0mm 10mm 0mm,clip,width=.19\linewidth]{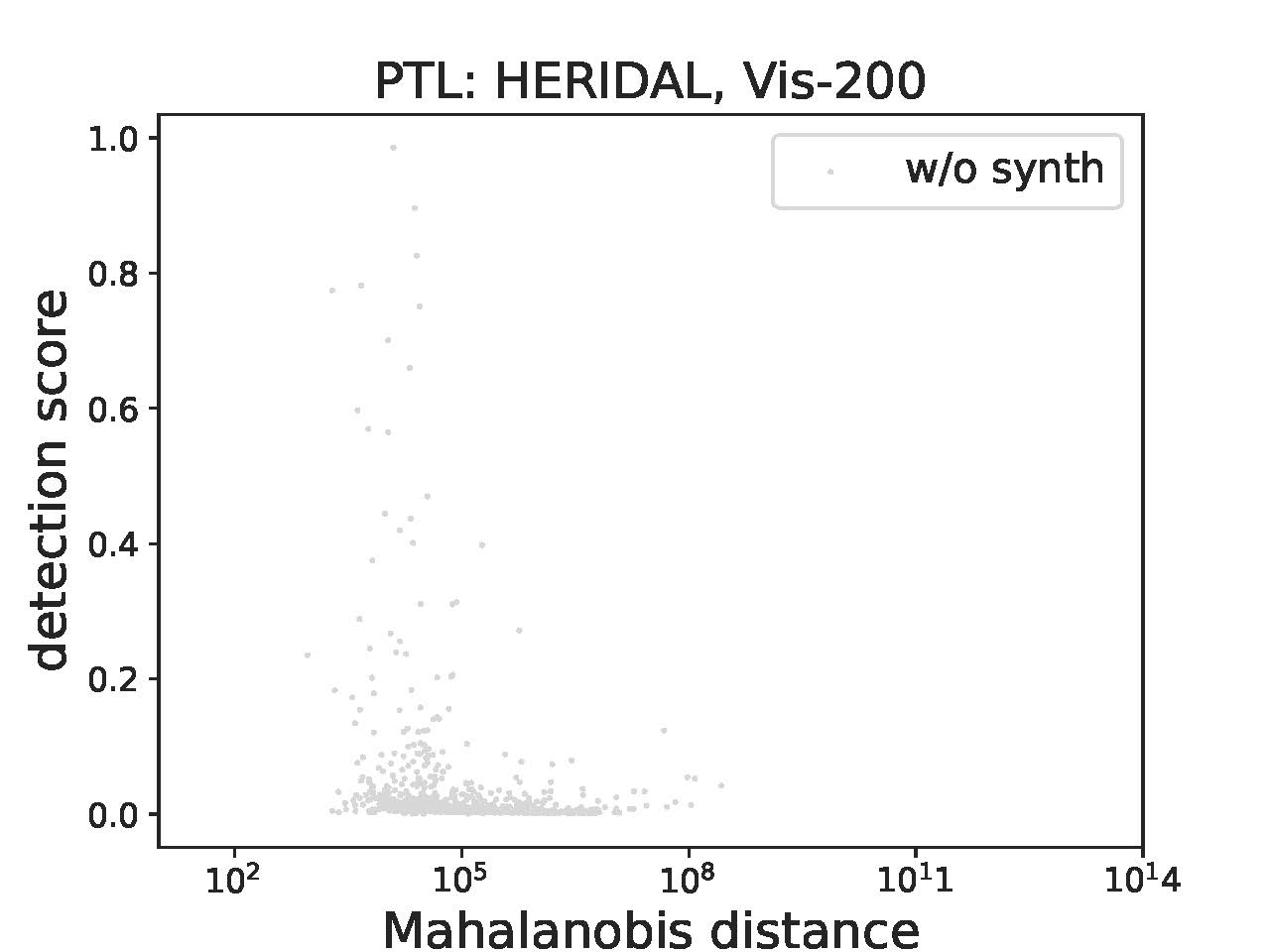}
\includegraphics[trim=10mm 0mm 10mm 0mm,clip,width=.19\linewidth]{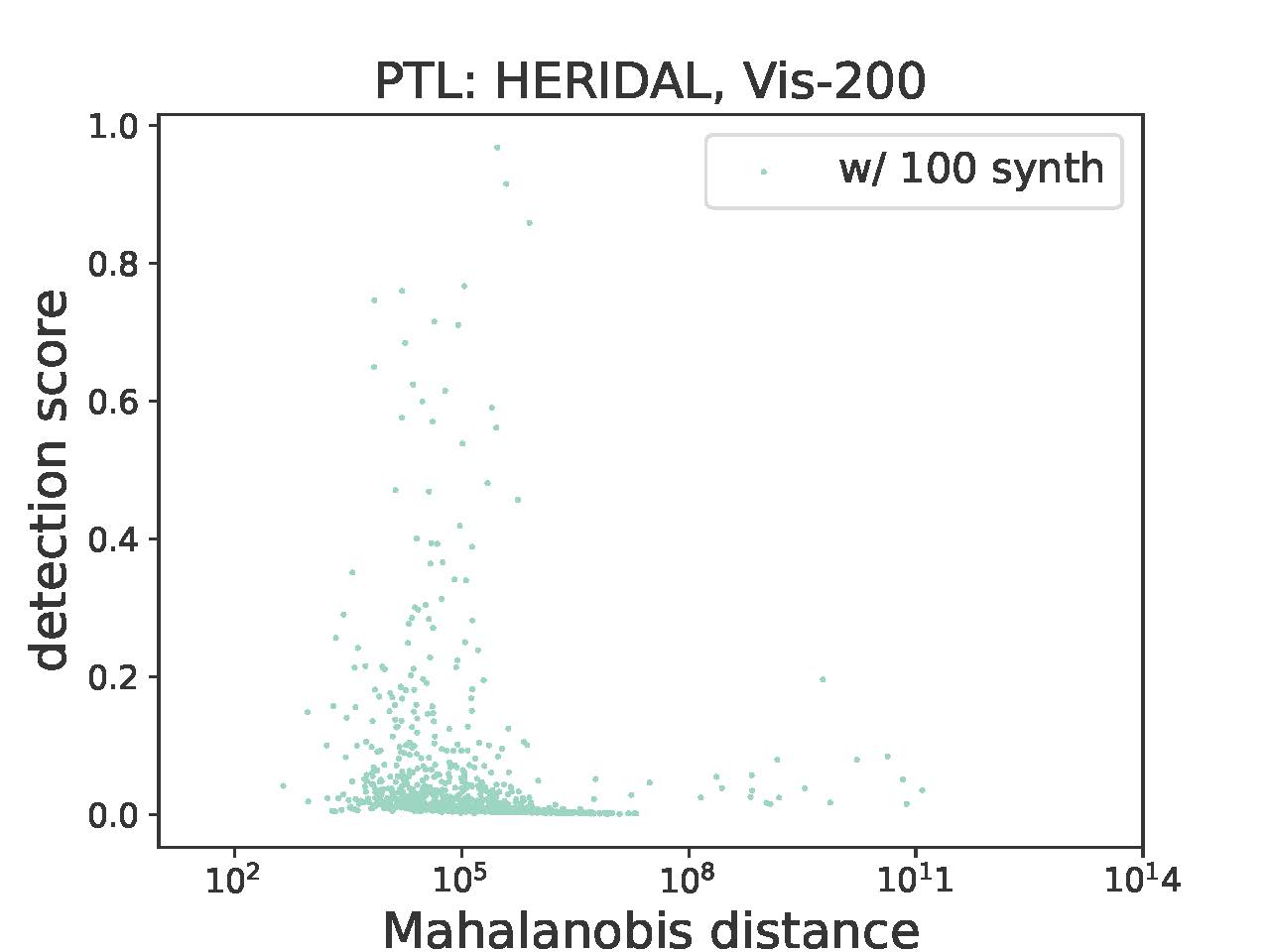}
\includegraphics[trim=10mm 0mm 10mm 0mm,clip,width=.19\linewidth]{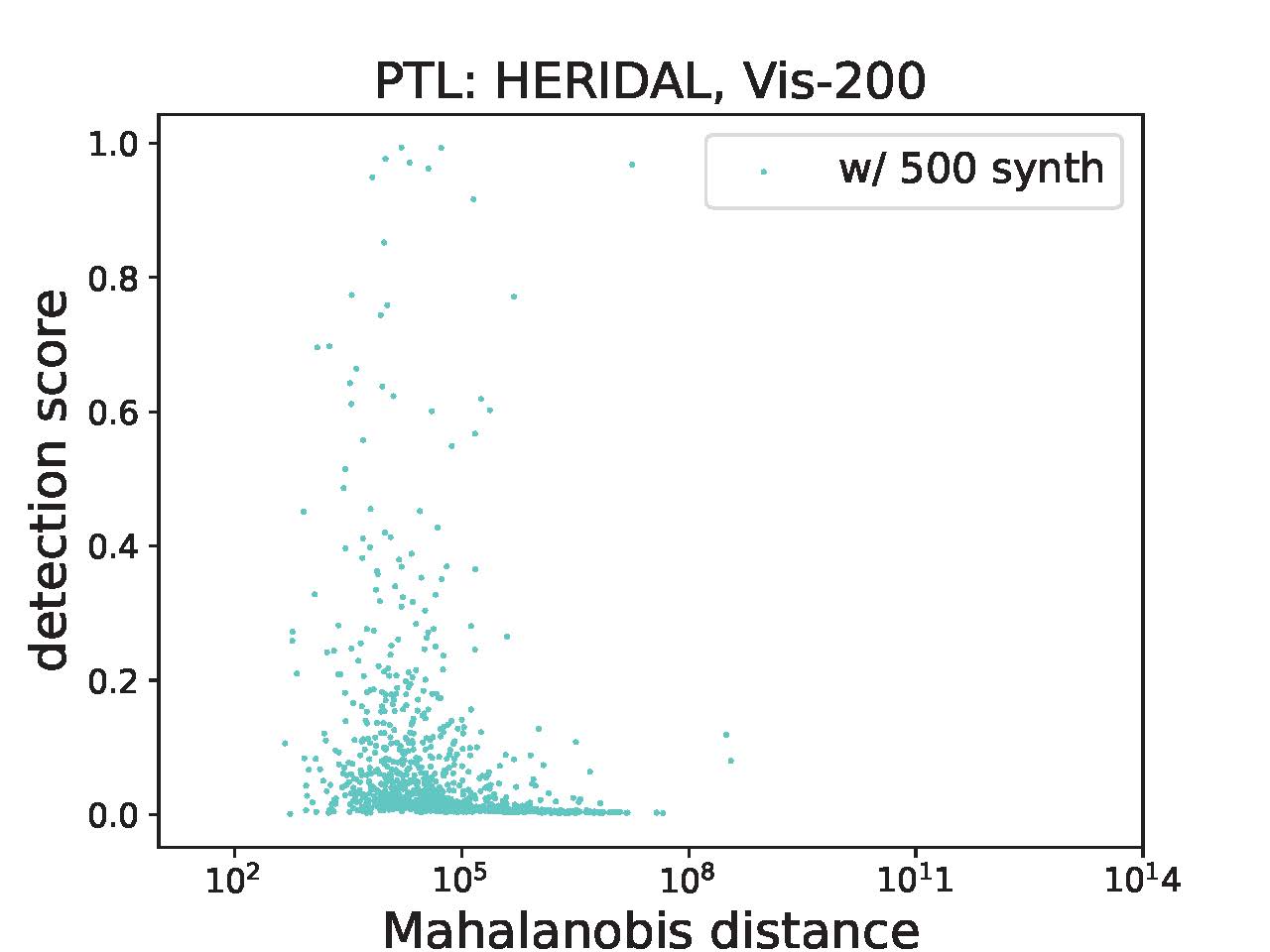}
\includegraphics[trim=10mm 0mm 10mm 0mm,clip,width=.19\linewidth]{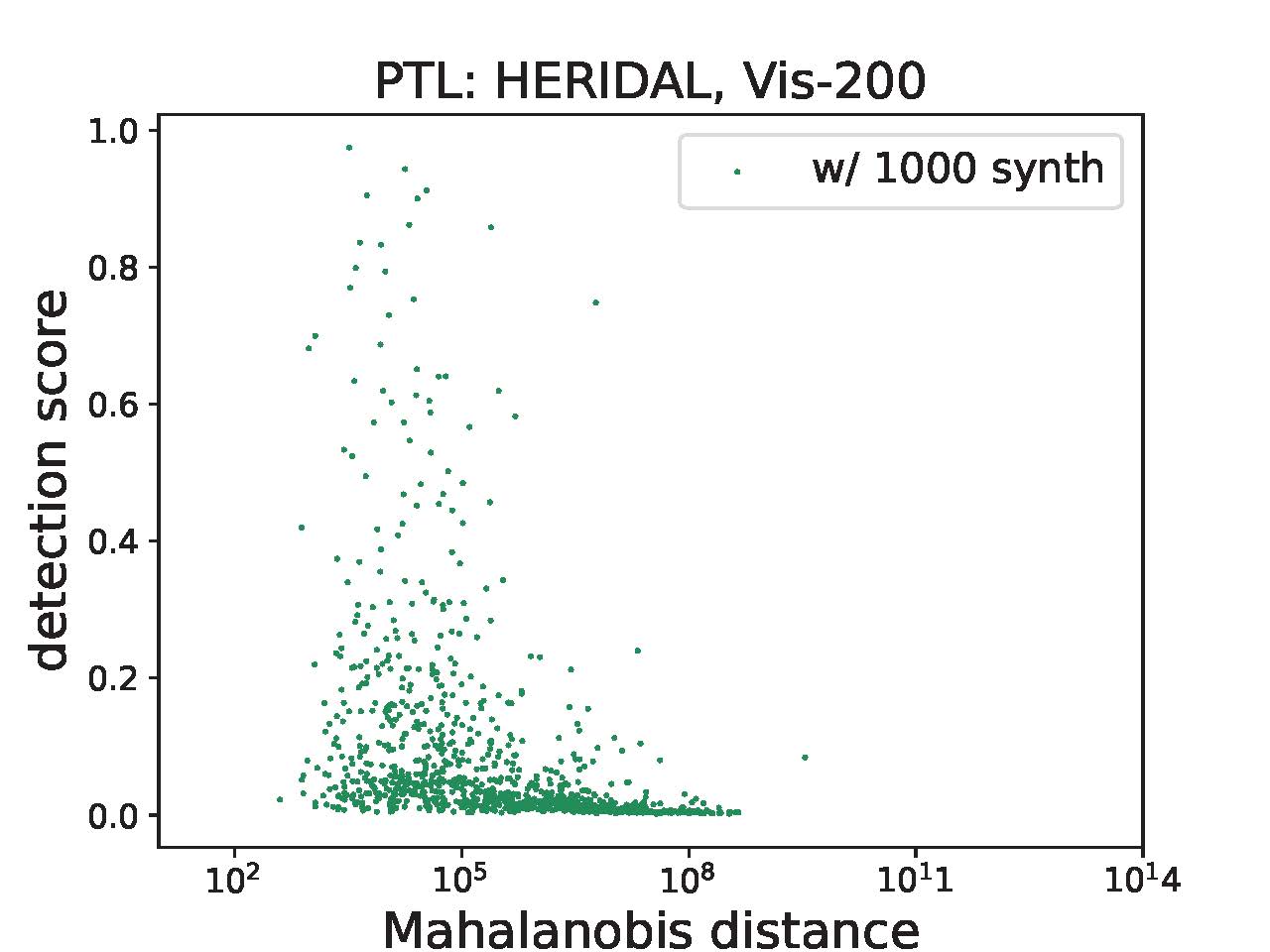}
\includegraphics[trim=10mm 0mm 10mm 0mm,clip,width=.19\linewidth]{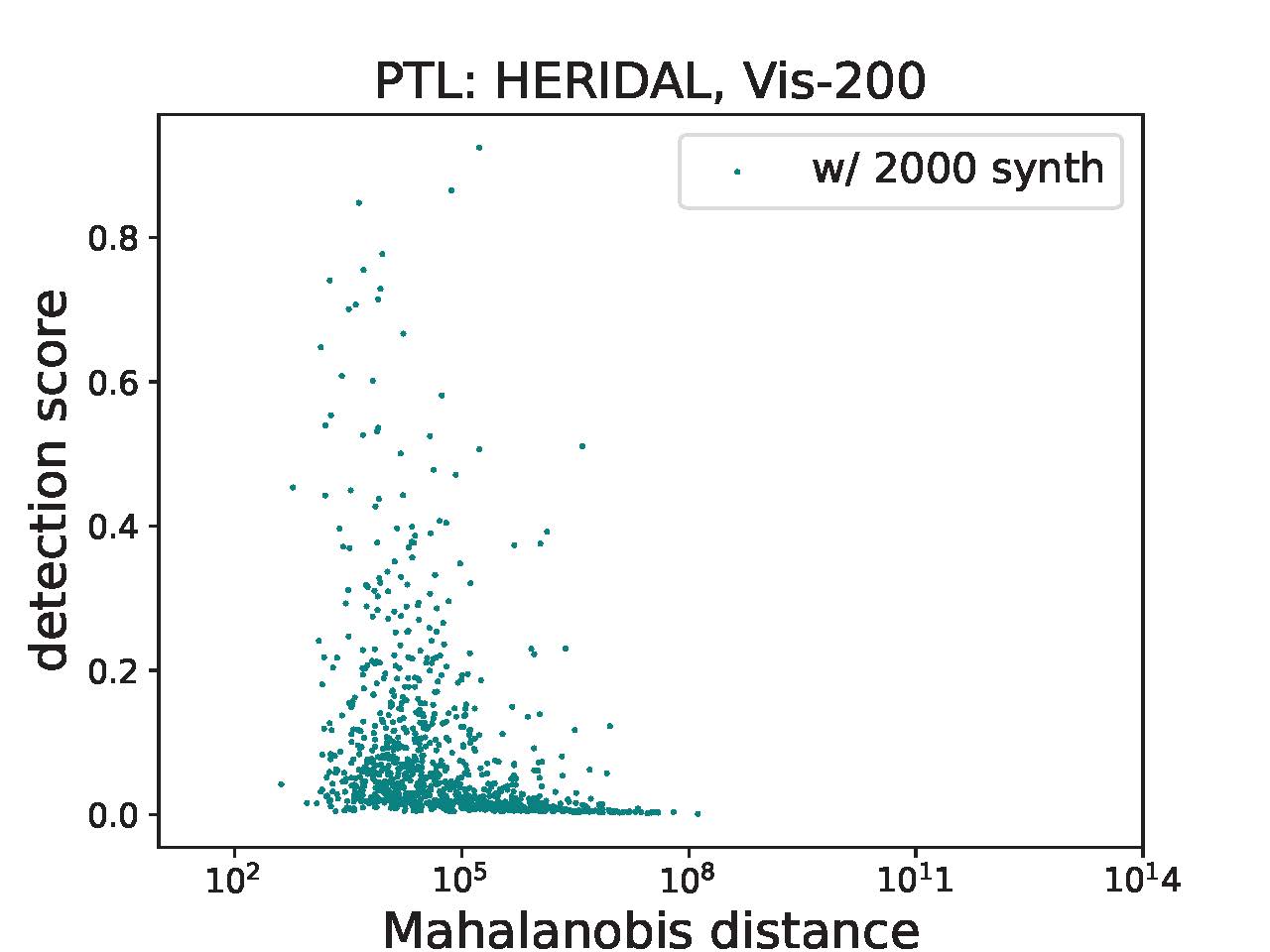}
\\
\includegraphics[trim=10mm 0mm 10mm 0mm,clip,width=.19\linewidth]{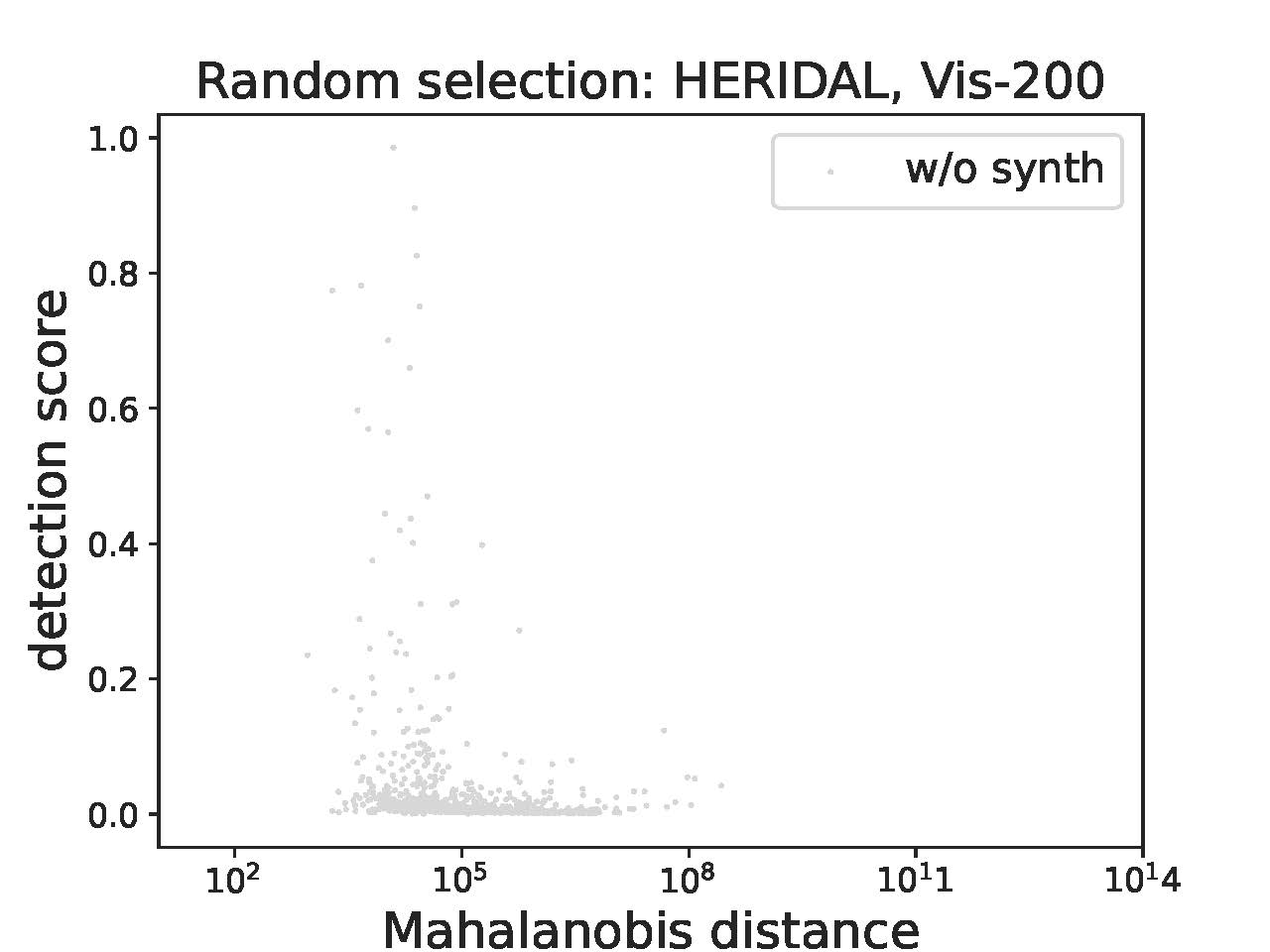}
\includegraphics[trim=10mm 0mm 10mm 0mm,clip,width=.19\linewidth]{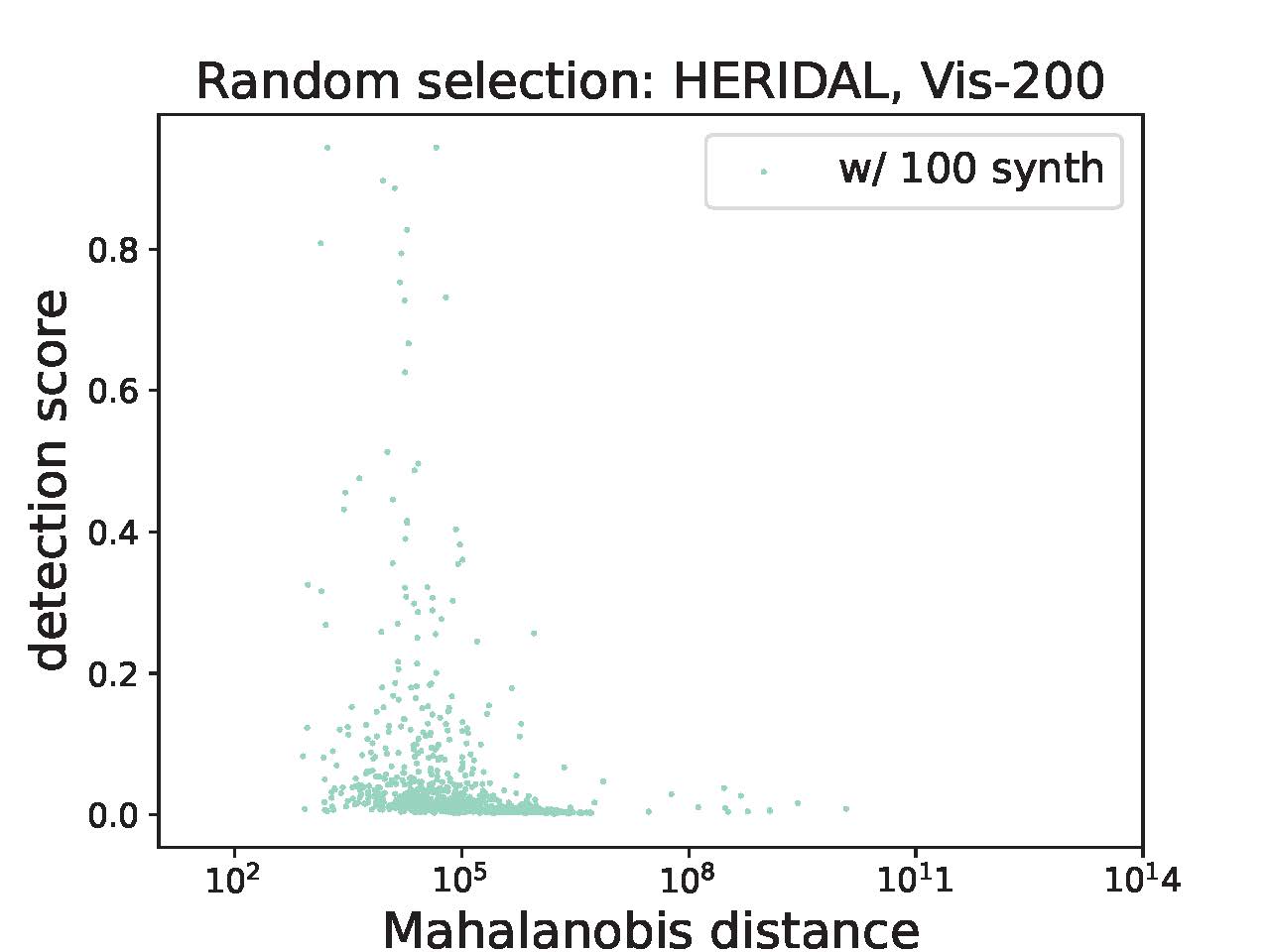}
\includegraphics[trim=10mm 0mm 10mm 0mm,clip,width=.19\linewidth]{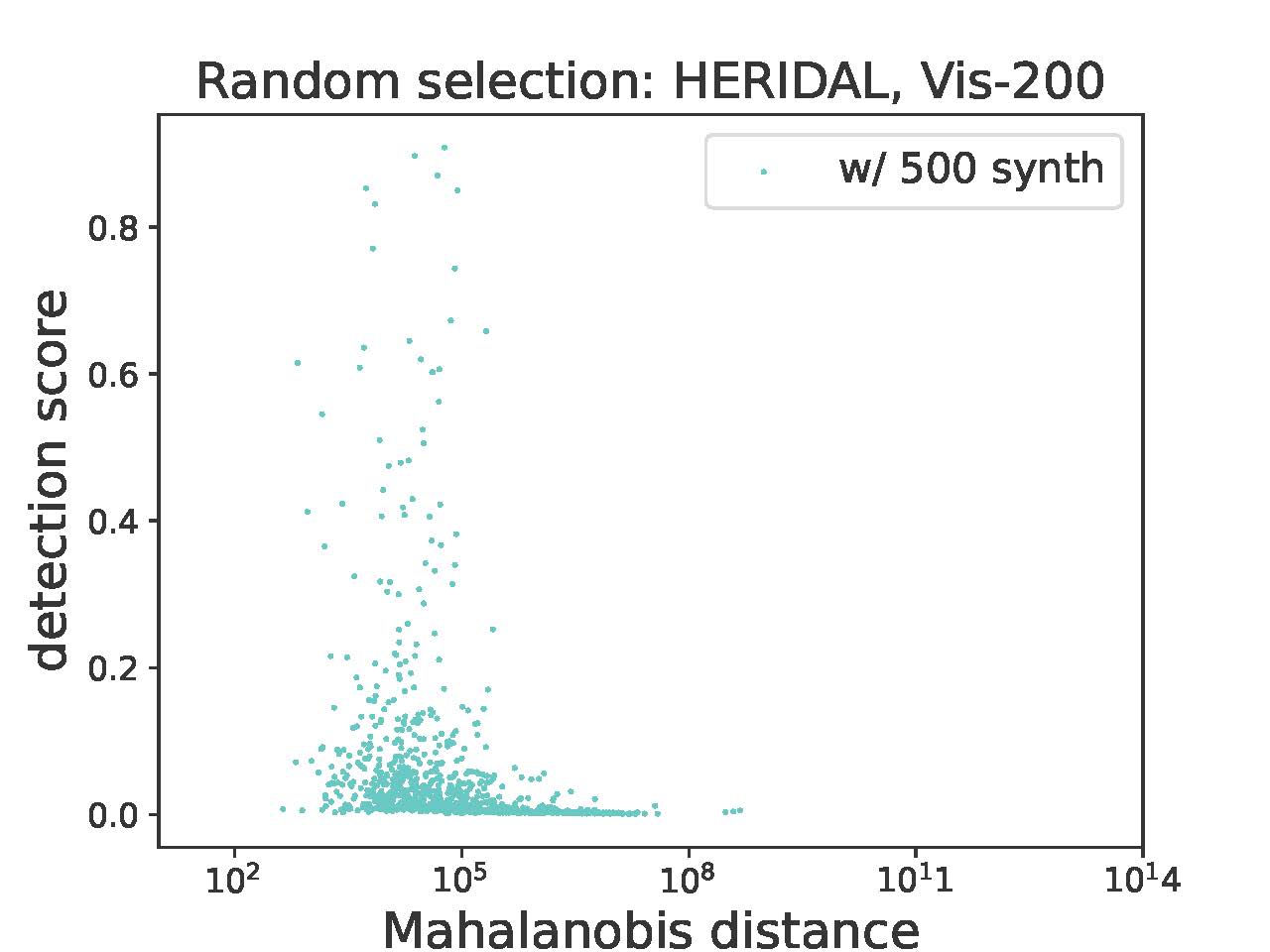}
\includegraphics[trim=10mm 0mm 10mm 0mm,clip,width=.19\linewidth]{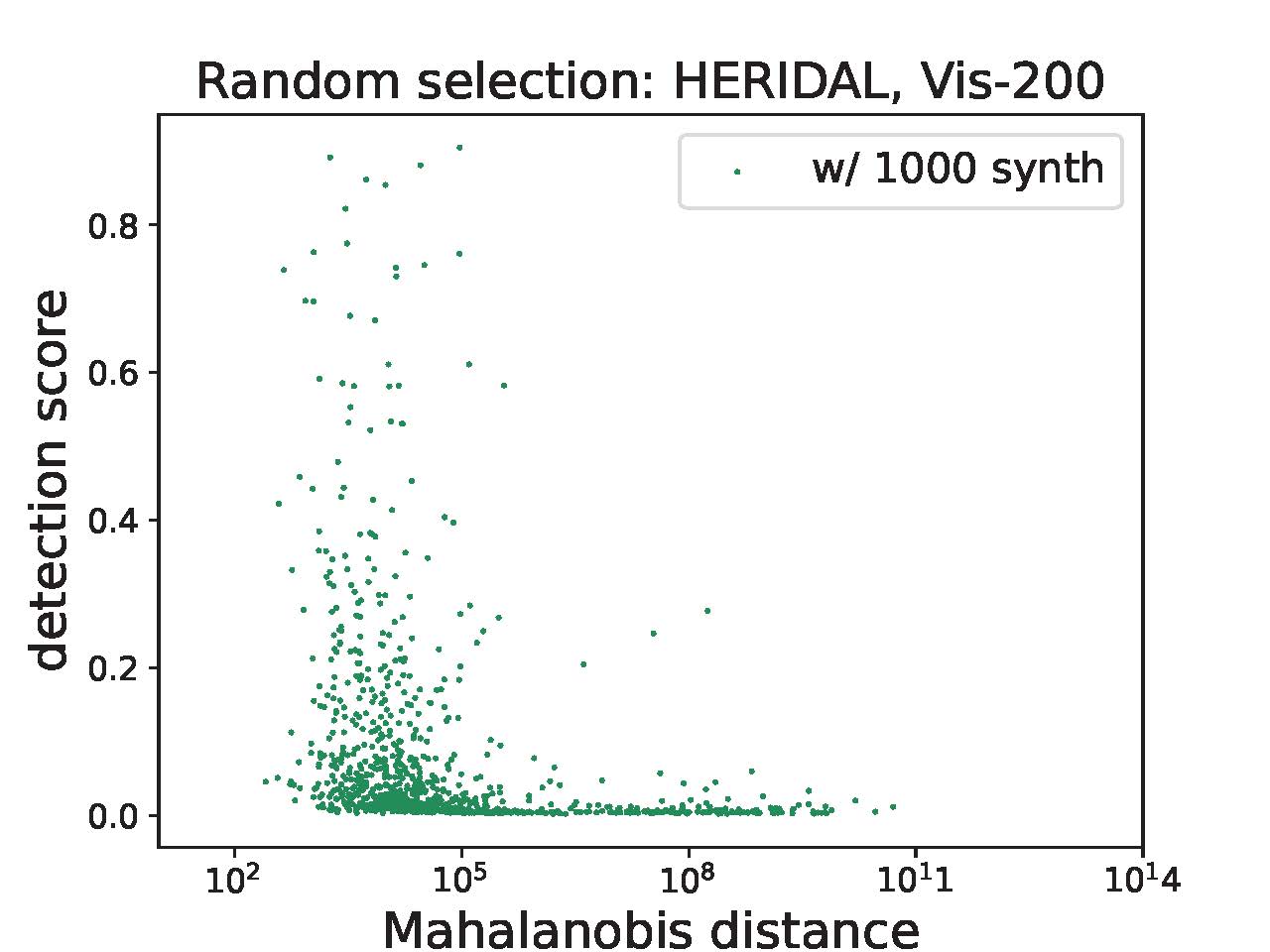}
\includegraphics[trim=10mm 0mm 10mm 0mm,clip,width=.19\linewidth]{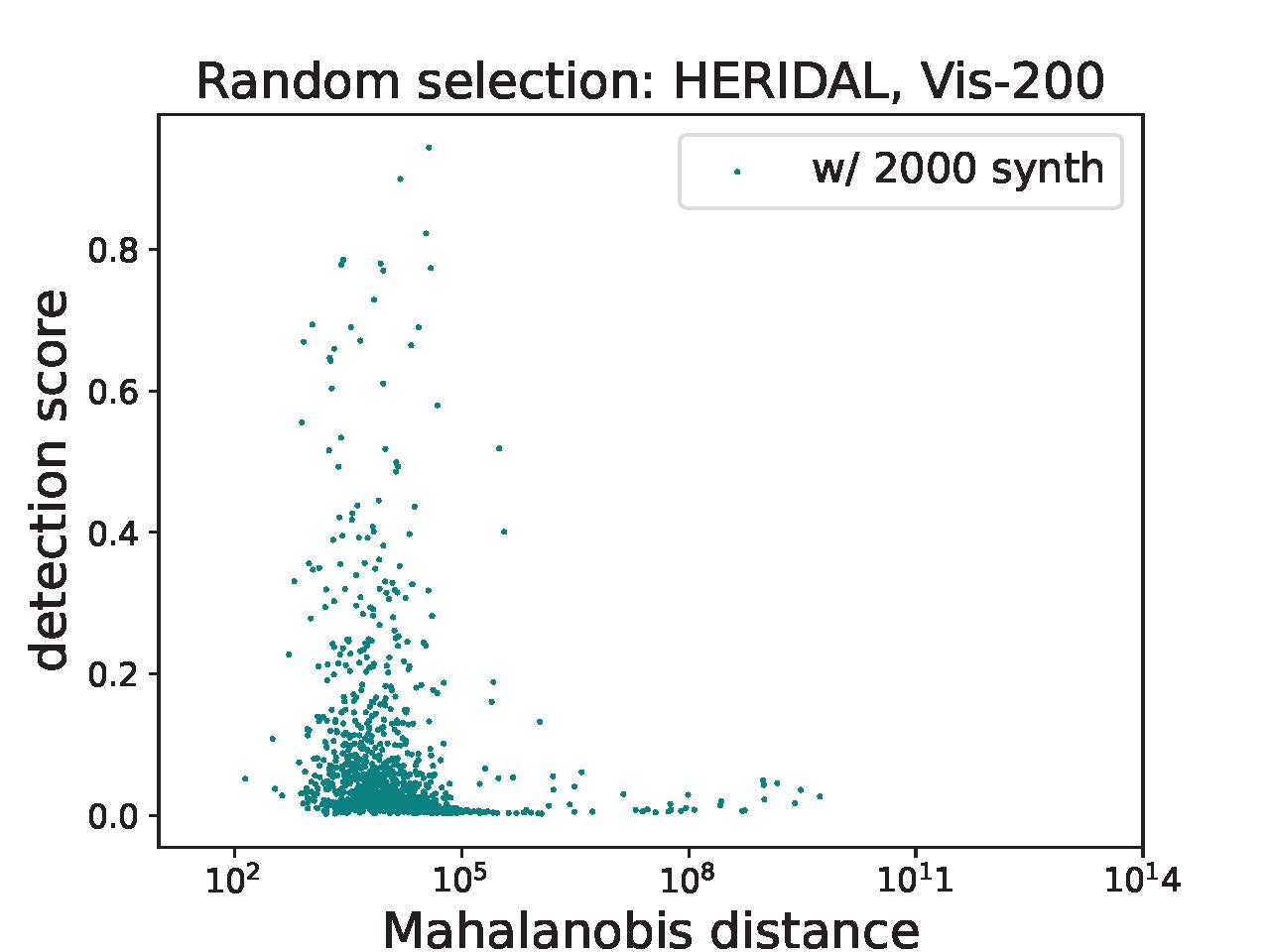}
\caption{Train: Vis-200, Test: HERIDAL}
\label{fig:scatter_heridal_vis200}
\end{subfigure}
\caption{{\bf Scatter plot of detection scores and Mahalanobis distances} with various numbers of synthetic images. For each case, plots in the first row and the second row represent the scatter results for PTL and random selection, respectively. Each of the five plots in each row shows the results without using synthetic images, or with using 100, 500, 1000, or 2000 synthetic images, in order.}
\label{fig:scatter_scalability_synthimg}
\end{figure*}

In Figure~\ref{fig:scatter_scalability_synthimg}, we also show the scatter plots of detection scores and Mahalanobis distances for different numbers of synthetic images used in training. Among different experimental settings, we present the scatter plots for three cases: i) testing on the Okutama-Action dataset in Vis-50, ii) testing on the SARD dataset in Vis-100, and iii) testing on the HERIDAL dataset in Vis-200. For reference, the first case is the same as the scatter plots in Fig.~3(b) and Fig.~5 of the main manuscript. To better focus on the distribution of each scatter plot, scatter plots are shown separately for each size of synthetic data. In the main manuscript, these scatterplots are shown together in one figure to emphasize the differences between the plots. The observations of change in the scatter plots for the other two cases are similar to those in the first case, which has been analyzed in the main manuscript.

%
%
\bibliographystyle{Styles/splncs04}
\bibliography{main}

\end{document}